\documentclass[12pt,a4paper]{article}

\usepackage[british]{babel}

\usepackage[a4paper,top=2cm,bottom=2cm,left=2.5cm,right=2.5cm,marginparwidth=1.75cm]{geometry}

\usepackage[T1]{fontenc}
\usepackage[utf8]{inputenc} 
\usepackage{lmodern}

\usepackage[most]{tcolorbox}
\newtcolorbox{promptbox}[1]{
  breakable,
  colback=white,
  colframe=black,
  title={#1},
  fonttitle=\bfseries,
  left=1mm,right=1mm,top=1mm,bottom=1mm,
  boxsep=1mm,
  enhanced,
  fontupper=\ttfamily\small
}
\usepackage[
backend=biber,
style=numeric,
sorting=none
]{biblatex}
\bibliography{main}
\RequirePackage{snapshot}

\usepackage{amsmath}
\usepackage{graphicx}
\usepackage[colorlinks=true, allcolors=blue]{hyperref}
\usepackage{hyperref}
\usepackage[title]{appendix}
\usepackage{mathrsfs}
\usepackage{amsfonts}
\usepackage{booktabs, tabularx, array} 
\usepackage{caption} 
\usepackage{threeparttable} 
\usepackage{algorithm}
\usepackage{algorithmicx}
\usepackage{algpseudocode}
\usepackage{listings}
\usepackage{enumitem}
\usepackage{chngcntr}
\usepackage{booktabs}
\usepackage{lipsum}
\usepackage{subcaption}
\usepackage{authblk}
\usepackage[T1]{fontenc}  
\usepackage{csquotes}  
\usepackage{diagbox}
\usepackage{comment}
\usepackage{url}
\usepackage{cleveref}
\newcolumntype{Y}{>{\raggedright\arraybackslash}X}

\usepackage[justification=raggedright, singlelinecheck=false]{caption}
\usepackage{helvet}  %

\usepackage{setspace}
\usepackage{titlesec}
\titleformat{\section} %
  {\normalfont\Large\bfseries}{\thesection.}{1em}{}
  
\usepackage{lineno} %

\rightlinenumbers %

\usepackage{float}   %
\usepackage{caption}

\usepackage[most]{tcolorbox}
\usepackage{fvextra} %
\tcbset{colback=gray!3, colframe=black!40, boxrule=0.4pt, arc=1.2mm}
\usepackage{authblk}

\newenvironment{prompttext}{%
  \Verbatim[
    fontsize=\small,
    formatcom=\ttfamily,
    breaklines=true,     %
    breaksymbolleft={},  %
    obeytabs=true,
    showspaces=false,
    showtabs=false
  ]}{\endVerbatim}
\date{}  %

\begin{document}

\title{Creativity Benchmark: A benchmark for marketing creativity for large language models}

\author{Ninad Bhat\thanks{Contact: \texttt{ninad@springboards.ai}}}
\author{Kieran Browne}
\author{Pip Bingemann}
\affil{Springboards.ai}
\date{} %

\maketitle

\begin{abstract}
We introduce \textit{Creativity Benchmark}, an evaluation framework for large language models (LLMs) in marketing creativity. The benchmark covers 100 brands (12 categories) and three prompt types (Insights, Ideas, Wild Ideas). Human pairwise preferences from 678 practising creatives over 11{,}012 anonymised comparisons, analysed with Bradley-Terry models, show tightly clustered performance with no model dominating across brands or prompt types: the top-bottom spread is $\Delta\theta \approx 0.45$, which implies a head-to-head win probability of $0.61$; the highest-rated model beats the lowest only about $61\%$ of the time. We also analyse model diversity using cosine distances to capture intra- and inter-model variation and sensitivity to prompt reframing. Comparing three LLM-as-judge setups with human rankings reveals weak, inconsistent correlations and judge-specific biases, underscoring that automated judges cannot substitute for human evaluation. Conventional creativity tests also transfer only partially to brand-constrained tasks. Overall, the results highlight the need for expert human evaluation and diversity-aware workflows.
\end{abstract}

\textbf{Keywords}: AI, creativity, benchmark, LLM.

\section{Introduction}

Large language models (LLMs) are increasingly used for creative work across domains from long-form writing to rapid ideation and concept development \cite{chui2023economic}. In marketing and advertising, they are increasingly applied across the workflow from surfacing brand insights, refining briefs to shaping marketing campaigns and  generating creative concepts. Evaluating such outputs is difficult: the open-ended nature of creative tasks admits many plausible answers and lacks a single ground truth, limiting the value of overlap-based automatic metrics and lexical similarity scores. 

One approach is to compare models on public benchmarks that include creativity-oriented assessments (for example, Chatbot Arena and EQ-Bench) \cite{chiang2024chatbotarenaopenplatform, paech2024eqbenchemotionalintelligencebenchmark}. These benchmarks typically rely on human preference judgments aggregated from pairwise comparisons and category labels, or on rubric-based ratings collected via LLM-as-judge protocols. While informative, their relevance to advertising is uncertain. Most prompts resemble general creative writing rather than brand-constrained briefs, so transfer to marketing tasks is unclear. Moreover, it is not established that LLM-as-judge preferences align with domain experts. Outcomes are sensitive to prompt wording, judge choice, and decoding settings. As a result, cross-model comparisons are fragile. Small changes in brand, prompt and judge can flip the ranking and make results difficult to reproduce.

Another option is to adapt standard human creativity tests such as Torrance Tests of Creative Thinking (TTCT) and Divergent Association Task (DAT) for LLM evaluation \cite{Zhao_Zhang_Li_Huang_Guo_Peng_Hao_Wen_Hu_Du_etal_2024,  fukuda-etal-2025-building,}. Prior studies have used both LLM-as-judge and human raters, but report limited agreement: overall Kendall’s tau around 0.50 and task-wise Pearson correlations ranging from about -0.21 to 1.00 depending on the task and prompt set. As with general-purpose benchmarks, these domain-agnostic tests may not transfer cleanly to advertising and marketing.

To address these gaps, we present Creativity Benchmark, a domain-grounded evaluation for marketing tasks. Marketing and advertising professionals with relevant domain knowledge rate model outputs, and we compare these ratings with LLM-as-judge scores to assess alignment. We also evaluate a TTCT/DAT-style scoring adaptation and test how well it correlates domain-specific judgments.

\section{Creativity: Definition and Criteria}

Creativity has multiple definitions, but for the purposes of this study we define creativity as generating outputs that are both \emph{novel} and \emph{appropriate} to the prompt and context. In this work, both components are necessary, high novelty without fit is noise, and high appropriateness without diversity is cliché. Creativity is assessed at both response-level and model-level.

\subsubsection*{Response-level criteria}
Participants judged creativity directly, with no explicit rubric. In interpreting results, we follow the common view that creativity combines novelty and appropriateness, though raters were not instructed in these terms.

\subsubsection*{Model-level criteria}
A model can produce a single striking response yet still be a weak creative partner if it cannot generate variety on demand. We therefore treat \emph{diversity} as a model-level requirement, encompassing both inter-prompt and intra-prompt variation:

\begin{itemize}
  \item \textbf{Inter-prompt diversity:} Vary ideas when the brand, audience, channel, or brief changes; avoid recycling the same frames across different contexts.
  \item \textbf{Intra-prompt diversity:} For the \emph{same} brand/audience/channel brief, produce meaningfully different options across samples rather than paraphrases or near-duplicates.
  \item \textbf{Anti-collapse:} Avoid narrowing to a small set of high-frequency templates even under repeated sampling.
\end{itemize}

\section{Structure of this study}
This study is divided into four parts. Parts A and B constitute the core benchmark, where we evaluate the creative performance of several LLMs in an advertising and marketing context. Parts C and D examine whether proxy evaluation methods (LLM-as-judge and conventional creativity tests) correlate with human ratings in this domain. To ensure the benchmark reflects real practice, the methodology was co-developed with practising creatives and with guidance from professional industry bodies across multiple regions.  Below is a brief overview of each component, full details are provided in the respective sections.

\subsection{Part A: Model Evaluation via Human Rating}
We ask industry practitioners in advertising, marketing, and strategy to evaluate model outputs in head-to-head pairings and select the preferred response. Model rankings are derived from the resulting pairwise comparisons using Bradley-Terry scores.

\subsection{Part B: Model Diversity}
To assess within-model variation, we generate multiple responses per prompt. Diversity is quantified using cosine similarity between embedding representations of each response (lower similarity indicates higher diversity).

\subsection{Part C: LLM-as-judge}
We apply the same head-to-head evaluation procedure as in Part A, but replace human raters with multiple LLM-as-judges. Bradley-Terry scores are again used to rank models. We then compare these rankings to the human-generated rankings to assess alignment.

\subsection{Part D: Conventional creativity tests}
We adapt tasks from standard creativity tests such as the Torrance Tests of Creative Thinking (TTCT) and the Divergent Association Task (DAT). Model performance is scored using common creativity metrics (fluency, originality, flexibility, and elaboration) and compared to the domain-specific human ratings reported in Part A.

\section{Benchmark Setup}
\subsection{Models evaluated}
Model selection was conducted with stakeholders to reflect LLMs in active use and strong baselines. Stakeholder organisations included: the American Association of Advertising Agencies (4A’s), Advertising Council Australia, the Account Planning Group (APG), D\&AD, the Institute of Practitioners in Advertising (IPA), the International Advertising Association (IAA), and The One Club for Creativity. The pool included proprietary APIs and open-source models, spanning both instruction-tuned general-purpose systems and reasoning-oriented variants. To examine progress over time, we included historical baselines (e.g., GPT-3.5 Turbo) alongside current releases. The evaluated models and API identifiers are listed in \Cref{tab:models}.

All generations were produced with a fixed sampling temperature ($T$) of $1.0$, and unless stated otherwise, all other decoding parameters followed each provider’s defaults (e.g., top-$p$, frequency/presence penalties). We chose $T\!=\!1.0$ because it is the default in most vendor APIs and provides a common, minimally tuned operating point.

\begin{table}[htbp]
\centering
\begin{tabular}{l l l}
\toprule
\textbf{Model Name} & \textbf{Model ID} & \textbf{Provider} \\
\midrule
Qwen3 & qwen/qwen3-32B & Alibaba \\
Claude 3.5 Sonnet & claude-3-5-sonnet-20241022 & Anthropic \\
Claude 3.7 Sonnet & claude-3-7-sonnet-20250219 & Anthropic \\
DeepSeek Chat & deepseek-chat & DeepSeek \\
DeepSeek Reasoner & deepseek-reasoner & DeepSeek \\
Gemini 1.5 Pro & gemini-1.5-pro & Google \\
Gemini 2.0 Flash & gemini-2.0-flash-001 & Google \\
Gemini 2.5 Pro & gemini-2.5-pro-preview-05-06 & Google \\
Grok 3 Beta & grok-3-beta & Grok \\
LLaMA 4 Scout & meta/llama-4-scout-17b-16e-instruct-maas & Meta \\
LLaMA 3.1 405B & meta/llama-3.1-405b-instruct-maas & Meta \\
Mistral Large 2411 & mistral-large-2411 & Mistral \\
GPT-3.5 Turbo & gpt-3.5-turbo-1106 & OpenAI \\
GPT-4o & gpt-4o-2024-11-20 & OpenAI \\
GPT-4.5 Preview & gpt-4.5-preview-2025-02-27 & OpenAI \\
OpenAI O3 & o3-2025-04-16 & OpenAI \\
\bottomrule
\end{tabular}
\caption{Models evaluated in this study.}
\label{tab:models}

\end{table}

\subsection{Prompt Selection} \label{sec:prompt_selection}
Prompt design was conducted in consultation with industry stakeholders to ensure real-world applicability, focused on a typical advertising agency process. User prompts were grouped into three types: \textit{Insights} (a concise, under 10-word insight that surfaced a surprising observation about people, culture, category, or product), \textit{Ideas} (a platformable campaign concept under 50 words grounded in a strategic or cultural truth), and \textit{Wild Ideas} (an unconventional yet on-brand campaign concept under 50 words intended to provoke discussion and spark headlines). The exact wording of the system and user prompts is provided in \Cref{app:prompt}.

\subsection{Brand Selection}
To ensure broad applicability, we curated 100 brands across 12 market categories, with a minimum of eight per category. The set spans 17 countries and balances global leaders with emerging challengers. The full list can be found in \Cref{app:brands}.

\section{Part A: Model Ranking via Human Rating}

\subsection{Introduction}
Large language models (LLMs) can generate brand-relevant insights and ideas, but their relative creative performance in the marketing and advertising field remains largely unknown. Crowd-sourced evaluations offer scale, yet often lack the domain knowledge required to judge resonance, originality, and brand fit. Similarly, LLM-as-judge protocols avoid human effort but can encode systematic preferences and drift from expert preferences. In this part, we therefore rely on expert evaluation: practising marketing and advertising professionals compare anonymised model responses and select the one they would choose in a real review setting.

We asked \textbf{678} advertising, marketing, and strategy practitioners to compare \textbf{11{,}012} anonymised pairs of model outputs for \textbf{100} brands across three prompt types (\emph{Insights}, \emph{Ideas}, \emph{Wild Ideas}). Model identities were blinded, response order was randomised, and brand-prompt assignments were balanced across participants. From these pairwise choices, we fitted a Bradley-Terry model to estimate latent strengths interpretable as head-to-head win probabilities.

\subsection{Methods}

\subsubsection{Participant Selection}
Participation was voluntary. We collaborated with seven industry organisations, including the American Association of Advertising Agencies (4A’s), Advertising Council Australia, the Account Planning Group (APG), D\&AD, the Institute of Practitioners in Advertising (IPA), the International Advertising Association (IAA), and The One Club for Creativity, to invite practising creatives, strategists, and marketers via member newsletters and forums. To encourage engagement, participants were shown a provisional personal ranking of models after 15 evaluations. In total, 678 participants contributed votes. Participation was anonymous, and trials were blinded. Model names were shown only after participants submitted their selections. We collected basic demographics (age, gender, country) to enable subgroup analyses. Additional details on participant composition are provided in \Cref{app:demographics}.

\subsubsection{Voting Options}
Each trial presented a brand and one prompt type (Insight, Platform Idea, or Wild Idea) with two anonymised responses labeled X and Y. Participants selected exactly one of:

\begin{itemize}
  \item \textbf{Response X}
  \item \textbf{Response Y}
  \item \textbf{They are too similar} (draw)
  \item \textbf{Not sure} (skip)
\end{itemize}
The interface is shown in \Cref{fig:interface}. The order of X and Y was randomised in every trial. Brand and prompt assignments were balanced across participants to reduce selection bias. A progress indicator was shown to participants throughout the task. In total, we recorded \textbf{11{,}012} pairwise votes.

\subsubsection{Ranking Calculation}
We estimated a latent skill parameter $\theta_m$ for each model $m$ using the Bradley-Terry framework \cite{bradley1952rank} on pairwise comparisons where participants selected the better response to the same brand brief. The win probability for model $i$ against model $j$ is
\begin{equation}
P(i \succ j)\;=\;\frac{e^{\theta_i}}{e^{\theta_i}+e^{\theta_j}}\;=\;\sigma(\theta_i-\theta_j),
\end{equation}
where $\sigma(x)=1/(1+e^{-x})$ is the logistic function.  
We fitted one pooled Bradley-Terry model across all prompt types and three separate models for \emph{Insights}, \emph{Ideas}, and \emph{Wild Ideas} to obtain both overall and family-specific rankings.

\subsection{Results}

\begin{figure}[H]
    \centering
    \includegraphics[width=0.85\textwidth]{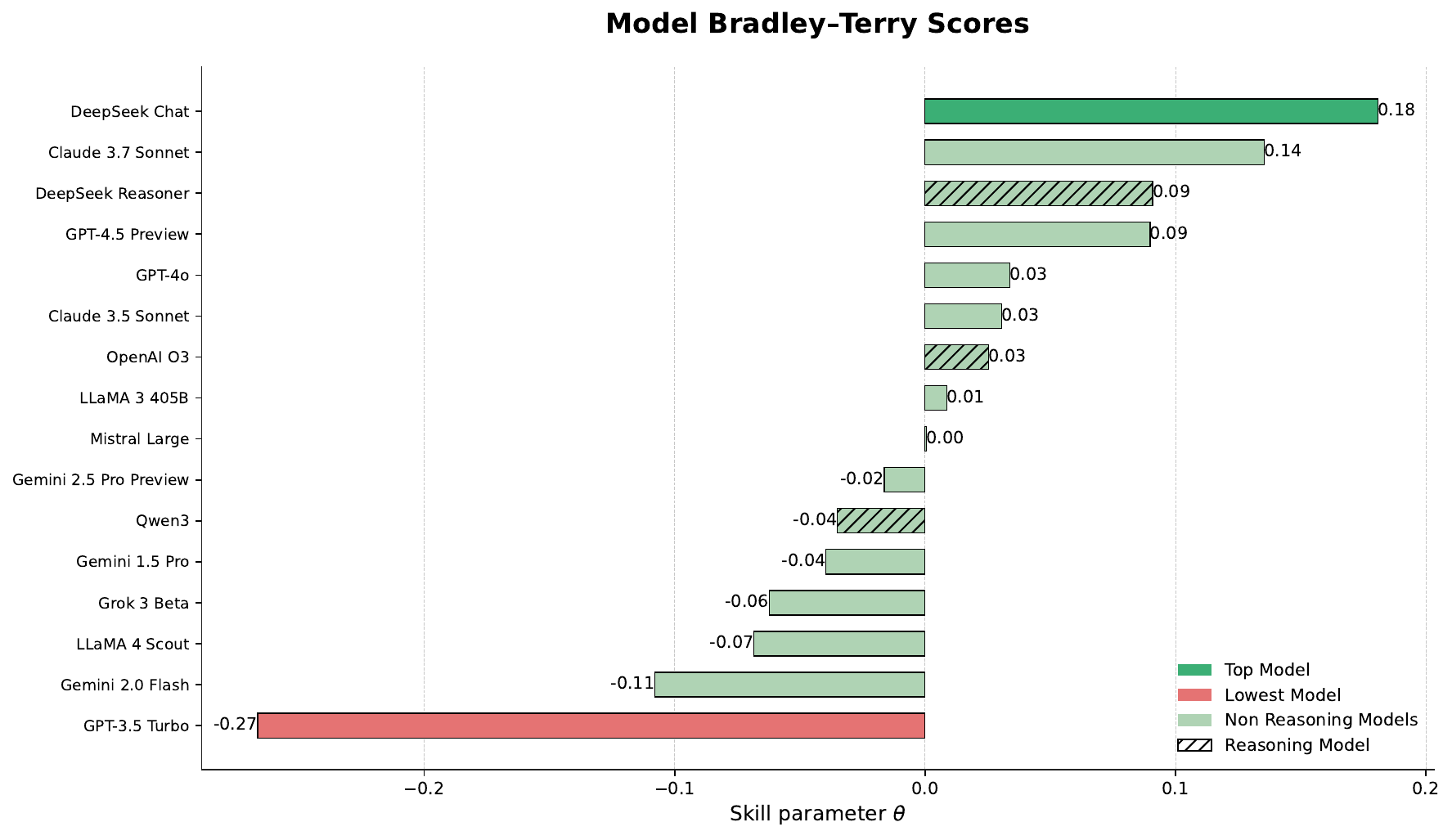}
    \caption{Overall Model Bradley-Terry scores (higher $\theta$ indicates better performance).}    \label{fig:overall_bt_scores}
\end{figure}

\Cref{fig:overall_bt_scores} reports Bradley-Terry model strengths ($\theta$) estimated from pairwise human preferences, aggregated across all prompt types. DeepSeek Chat ranks first ($\theta = 0.18$), followed by Claude 3.7 Sonnet and DeepSeek Reasoner. The spread is modest, with scores spanning $0.18$ to $-0.27$. For the corresponding head-to-head win rates computed directly from human votes, see \Cref{fig:h2h_winrates} in the appendix. Under a Bradley-Terry interpretation, the top model would be preferred in about $61\%$ of head-to-head comparisons against the lowest-ranked model, indicating limited practical distance and showing that lower-ranked LLMs often produce competitive outputs. Reasoning-oriented models such as DeepSeek Reasoner, OpenAI O3, and Qwen3 cluster near the middle. GPT-3.5 Turbo trails the group, reflecting gains achieved by more recent models.

\subsubsection{Prompt Type Breakdown}

\begin{figure}[H]
  \centering
  \begin{minipage}[t]{0.32\textwidth}
    \centering
    \includegraphics[width=\linewidth]{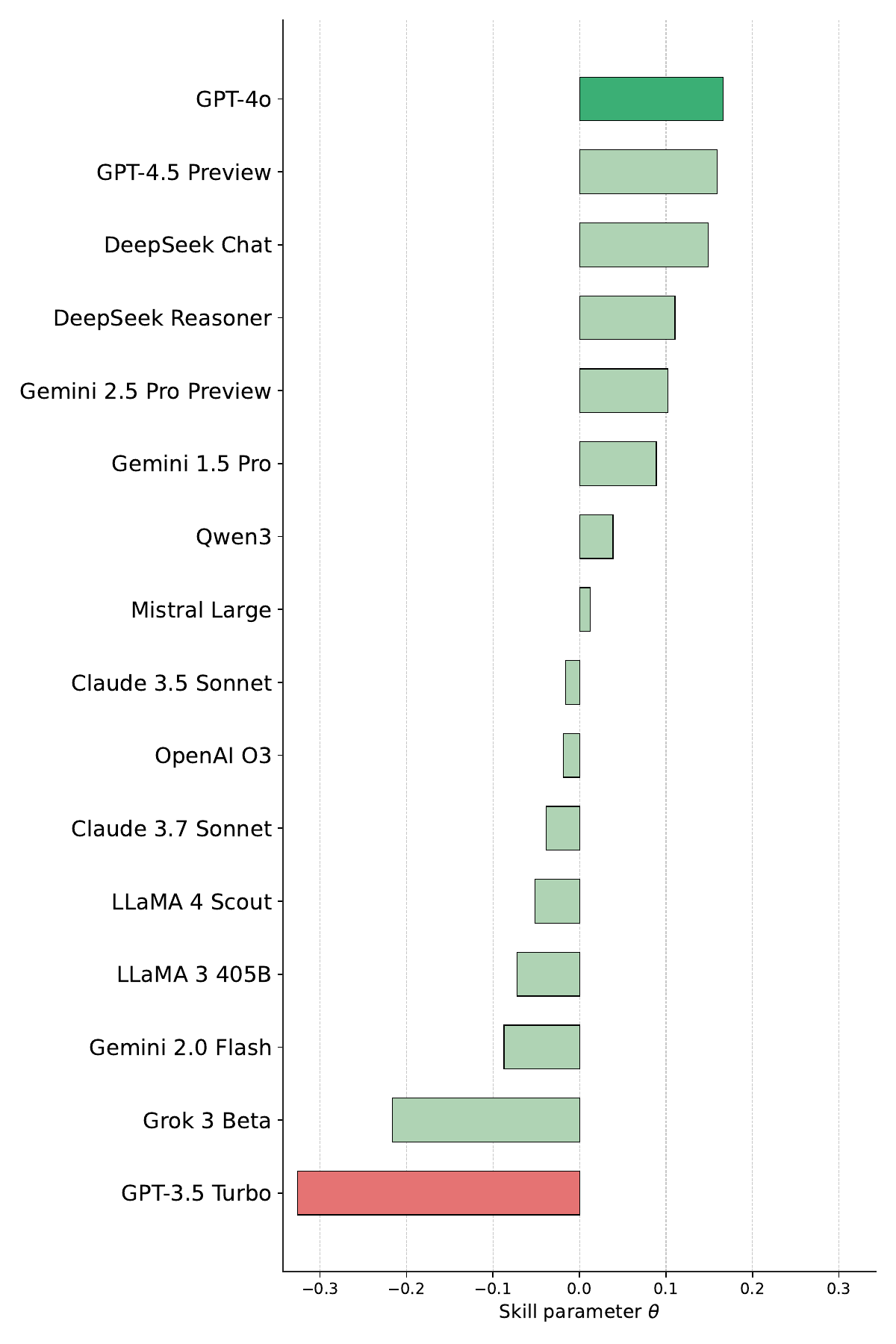}
    \caption*{(a) Insights}
  \end{minipage}\hfill
  \begin{minipage}[t]{0.32\textwidth}
    \centering
    \includegraphics[width=\linewidth]{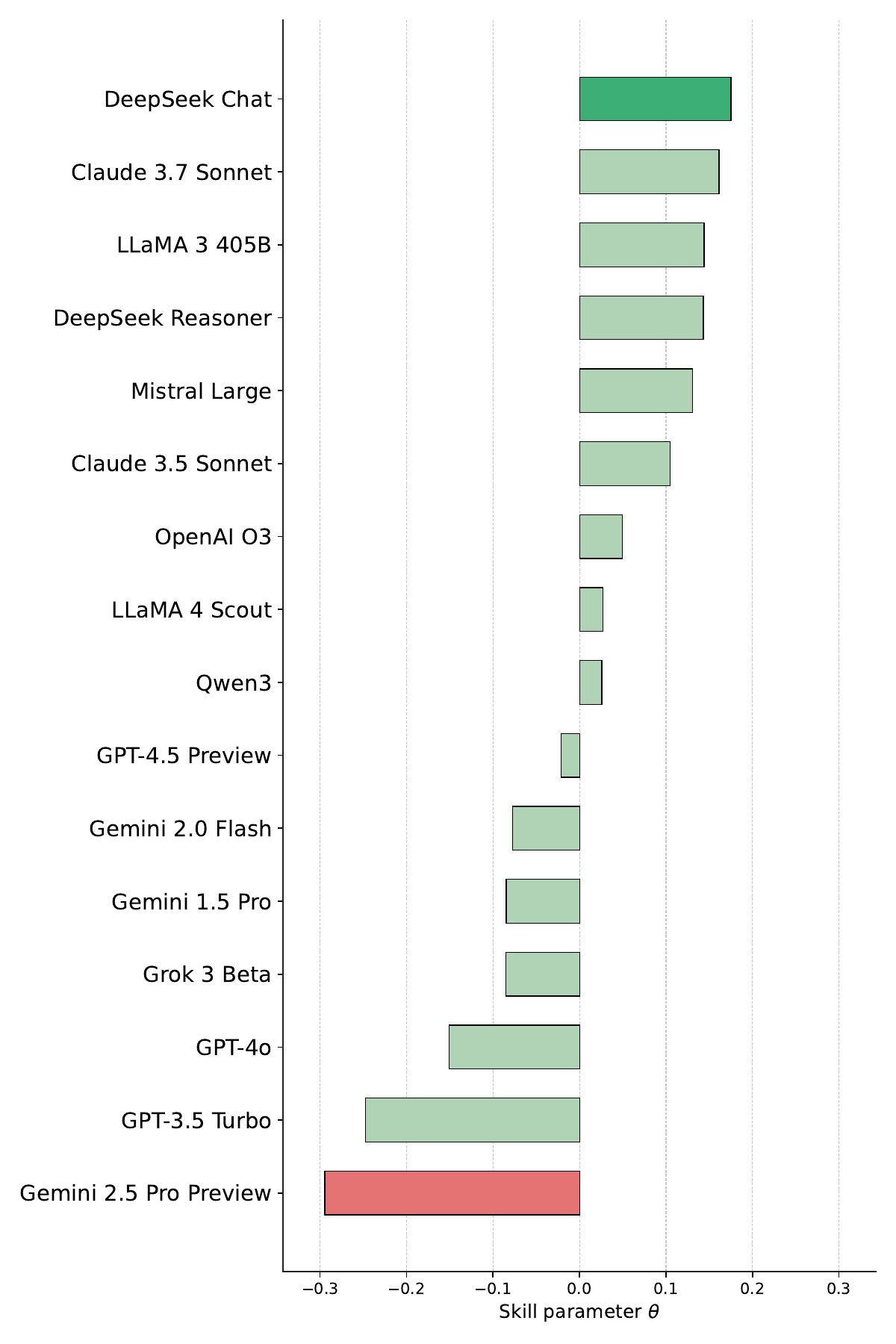}
    \caption*{(b) Ideas}
  \end{minipage}\hfill
  \begin{minipage}[t]{0.32\textwidth}
    \centering
    \includegraphics[width=\linewidth]{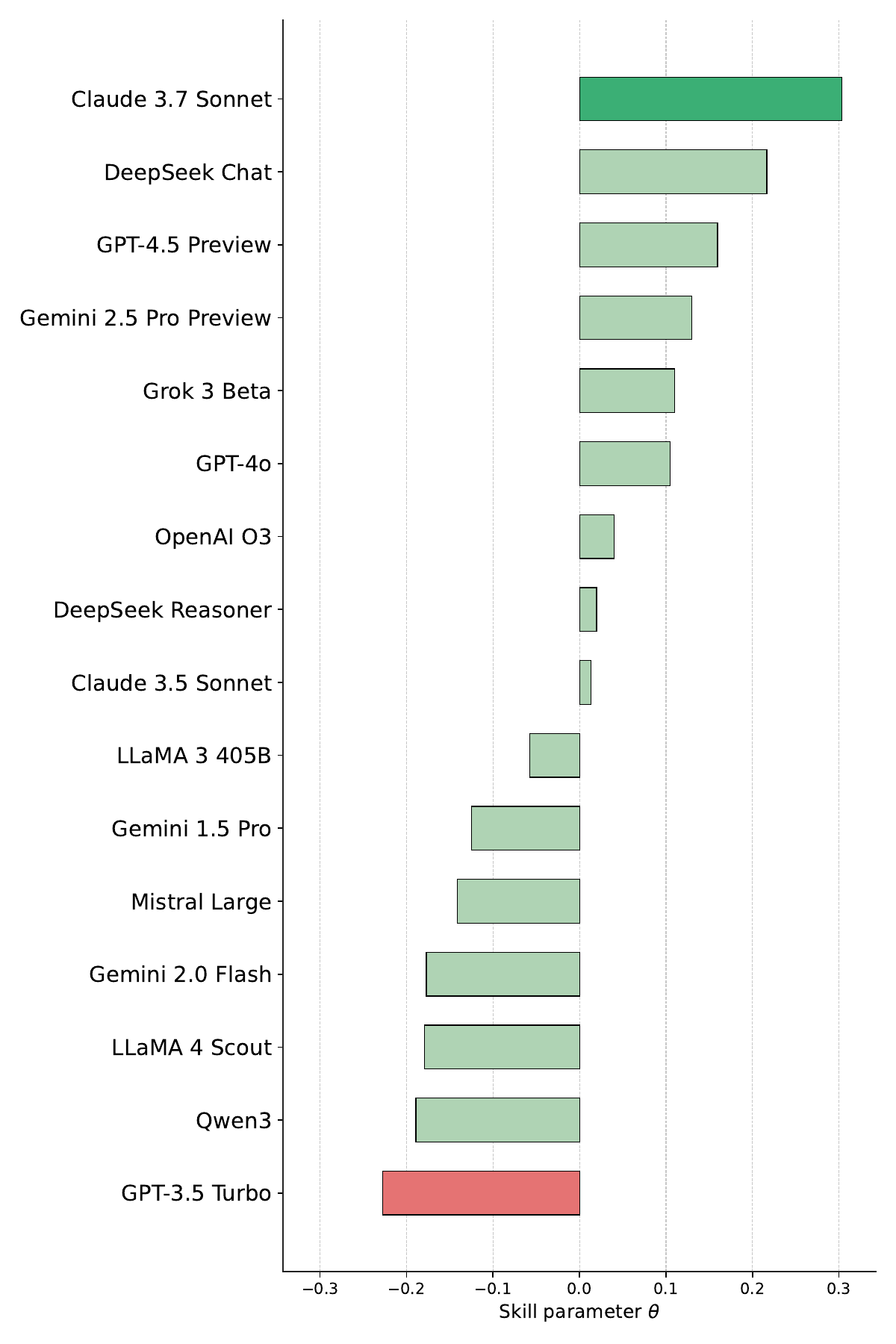}
    \caption*{(c) Wild Ideas}
  \end{minipage}

  \caption{Model ranking across three prompt types.}
  \label{fig:three_plots_minipage}
\end{figure}

\Cref{fig:three_plots_minipage} reports Bradley-Terry strengths by prompt type. No model dominates across categories. Performance is prompt-dependent and dispersion is limited: the gap between the highest-scoring and lowest-scoring LLMs rarely exceeds $0.3$ in Bradley-Terry scores, corresponding to small differences in pairwise win probabilities. Ranking differences are therefore present but not large for typical creative tasks.

By category, \textit{Ideas} is led by DeepSeek Chat, Claude 3.7 Sonnet, and LLaMA 3.1 405B, while Gemini 2.5 Pro Preview and GPT-3.5 Turbo rank lower. In \textit{Insights}, GPT-4o, GPT-4.5 Preview, and DeepSeek Chat achieve the highest scores, with Claude 3.7 Sonnet comparatively weaker. For \textit{Wild Ideas}, Claude 3.7 Sonnet attains the top score, followed by DeepSeek Chat and GPT-4.5 Preview; Qwen3 and Gemini 2.0 Flash appear lower. Overall, DeepSeek Chat and Claude 3.7 Sonnet emerge as strong generalists, remaining near the top across prompt types. Because pairwise win-rate margins are small (typically 0.45-0.55), subgroup orderings may be unstable and should be interpreted with caution.

\subsubsection{Models preference across demographic dimensions.}

\begin{figure}[H]
  \centering
  \begin{minipage}[t]{0.32\textwidth}
    \centering
    \includegraphics[width=\linewidth]{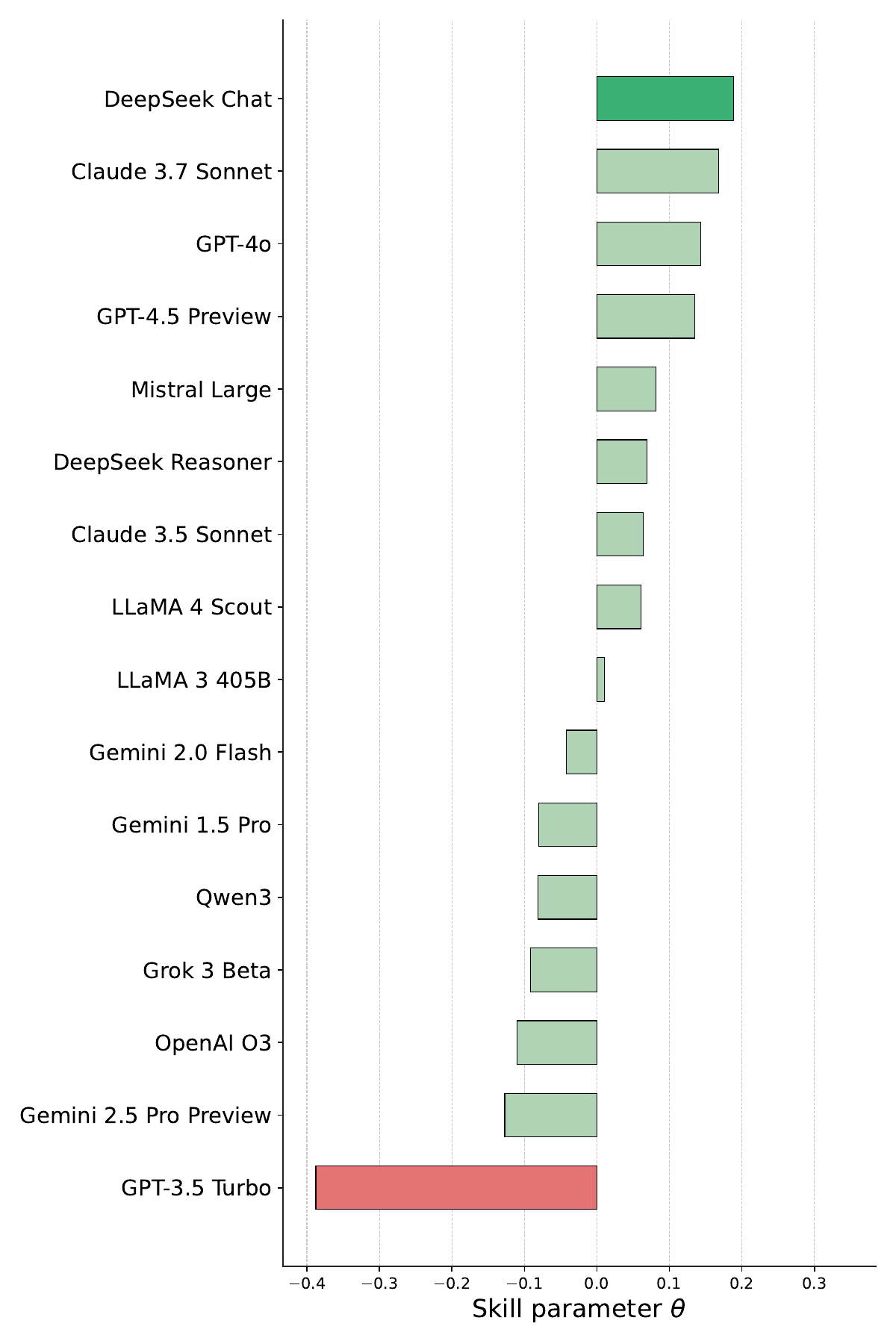}
    \caption*{(a) Australia}
  \end{minipage}\hfill
  \begin{minipage}[t]{0.32\textwidth}
    \centering
    \includegraphics[width=\linewidth]{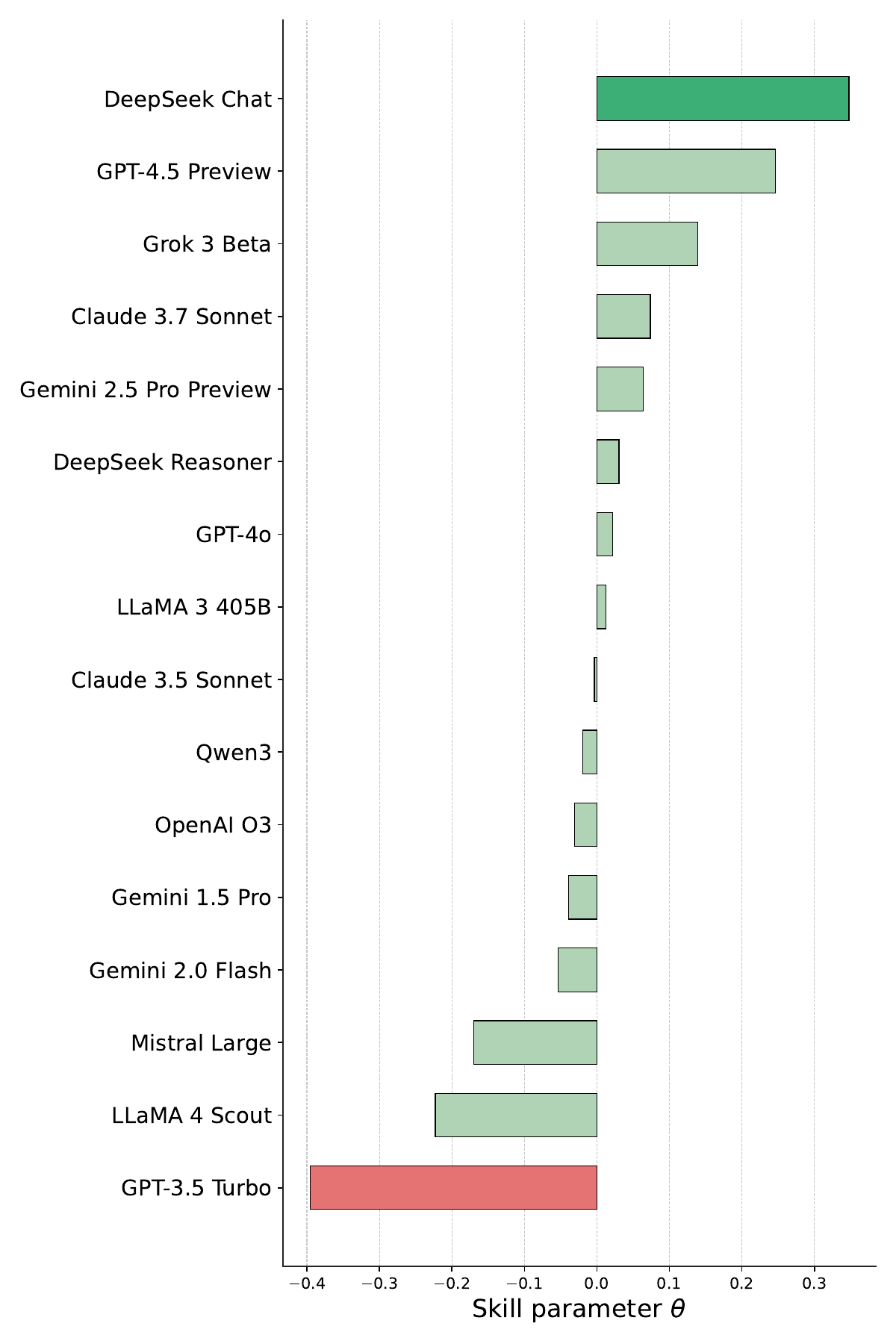}
    \caption*{(b) United Kingdom}
  \end{minipage}\hfill
  \begin{minipage}[t]{0.32\textwidth}
    \centering
    \includegraphics[width=\linewidth]{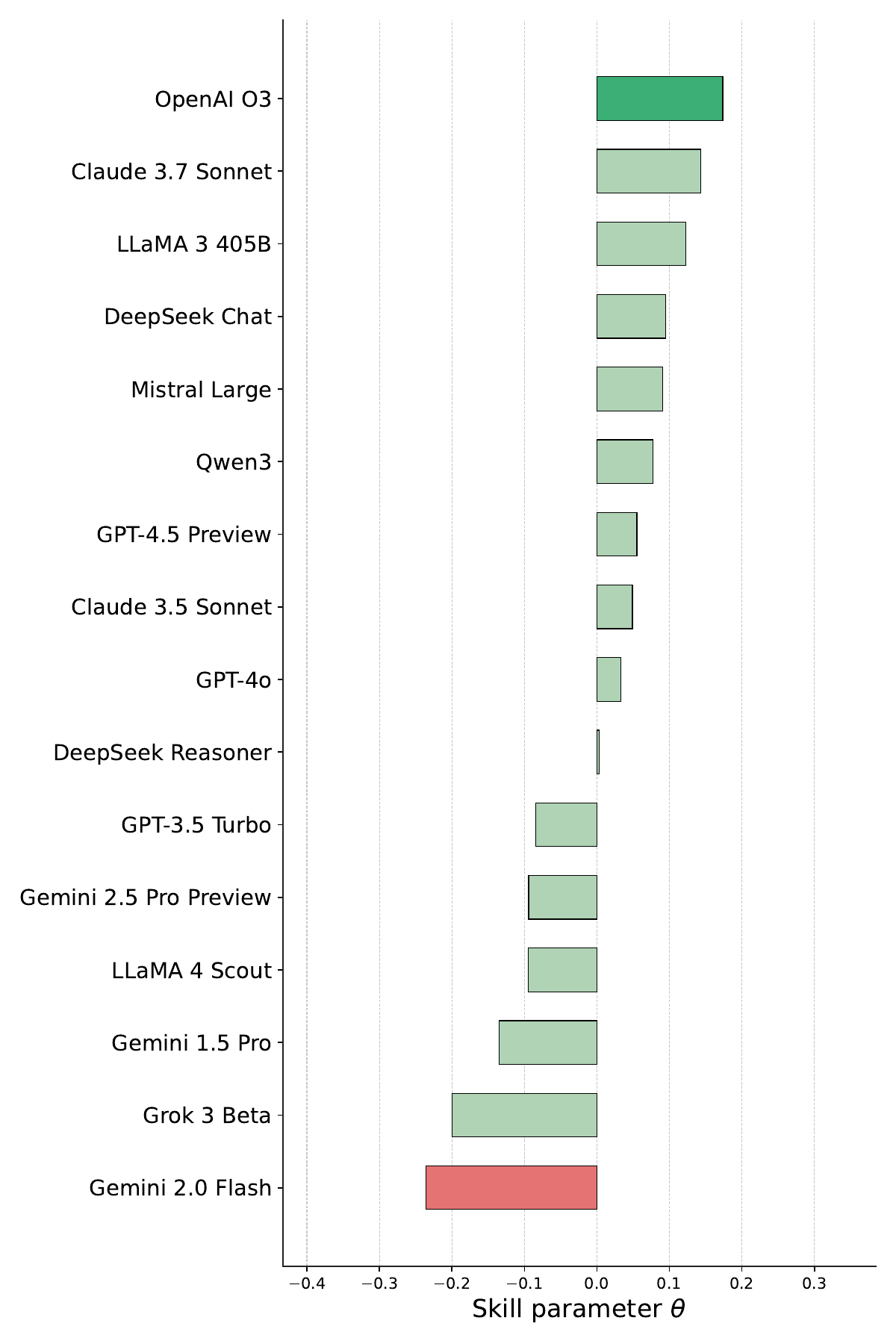}
    \caption*{(c) United States}
  \end{minipage}

  \caption{Model ranking across country.}
  \label{fig:bt_by_country}
\end{figure}

\Cref{fig:bt_by_country} shows modest geographic variation in model preferences.  The analysis focuses on Australia, the United Kingdom, and the United States because these groups constitute the majority of participants. Other countries had smaller samples, yielding less stable estimates, and are not shown. Respondents from Australia and the United Kingdom tend to prefer DeepSeek Chat, whereas respondents from the United States favour OpenAI O3, which ranks lower in the other two groups. Despite these shifts, dispersion within each country is limited. For most models, roughly two thirds of Bradley-Terry scores lie between $-0.10$ and $0.15$, indicating that creative performance remains tightly clustered even when national preferences differ. Alternative breakdowns by age and gender, which can be seen in \Cref{app:model_pref_demo}, show no significant reordering of models.

\subsection{Implications}

\paragraph{Model choice.}
Creative performance is tightly clustered. Across brands and prompt types, the gap between the highest and lowest $\theta$ values is usually below $0.3$, which corresponds to small differences in pairwise win rates. This suggests that, in deployment, it is wise to prioritise other differentiators such as brand-voice control, safety/guardrails, latency, cost, and tool integration, rather than small average rank gaps. Personal and team preferences also matter: practitioners differ in the interfaces, workflows, and stylistic tendencies they find effective, so testing with end users and selecting what they will actually adopt is often the deciding factor.

\section{Model Diversity}

\subsection{Introduction}
Most evaluations of large language models focus on the quality and novelty of a single response. In creative work, effective ideation also depends on diversity, that is, the breadth of distinct on-brief options a model can produce for the same prompt. An LLM that yields one strong answer and many near-duplicates narrows the option set and offers limited support to users.

Part B analyses intra-prompt diversity. For each model, we sample multiple completions per prompt and quantify how distinct the outputs are. This allows us to assess whether models provide a useful spread of ideas rather than recycling a few generic patterns, and to study the relationship between response quality and response diversity.

\subsection{Methods}

\subsubsection{Prompts Used}
We reused the prompt set described in \Cref{sec:prompt_selection} to ensure direct comparability between the quality and diversity results. For each brand–prompt pair, every model generated ten independent responses. The temperature setting was the same as in the quality-assessment phase ($T = 1.0$).

\subsubsection{Embedding}
Every response was converted into a 1024-dimension embedding with BAAI/bge-large-en-v1.5 \cite{bge_embedding}, a general-purpose sentence embedding model. Working in embedding space allowed us to capture semantic differences beyond surface-level wording \cite{cer2018universalsentenceencoder}.

\subsubsection{Diversity metrics}

\paragraph{Intra-model diversity.}
For each model $m$ and prompt $p$, we generated $N=10$ completions and embedded each response with a fixed encoder to obtain vectors $\mathbf{v}_{m,p,i}\in\mathbb{R}^{d}$ for $i=1,\dots,N$. Cosine similarity between two responses was
\[
s_{m,p,i,j}  =  \frac{\mathbf{v}_{m,p,i}\cdot \mathbf{v}_{m,p,j}}{\|\mathbf{v}_{m,p,i}\|\,\|\mathbf{v}_{m,p,j}\|}\,,
\qquad 1\le i<j\le N,
\]
and we defined cosine distance as $d_{m,p,i,j}=1-s_{m,p,i,j}$. Identical texts yielded $d=0$; orthogonal embeddings yielded $d=1$. The per-prompt intra-model diversity was the average pairwise distance
\[
D_{m,p}  =  \frac{2}{N(N-1)}\sum_{1\le i<j\le N} d_{m,p,i,j},
\]
and the model-level score averaged over $P$ prompts
\[
\overline{D}_{m}  =  \frac{1}{P}\sum_{p=1}^{P} D_{m,p}.
\]
Higher $\overline{D}_{m}$ indicated that a model explored a broader set of ideas when the prompt was held fixed.

\paragraph{Inter-model diversity.}
Inter-model diversity measures how distinct two models are on the same prompt. For models $a$ and $b$, we computed all cross pairs
\[
s_{a,b,p,i,j}  =  \frac{\mathbf{v}_{a,p,i}\cdot \mathbf{v}_{b,p,j}}{\|\mathbf{v}_{a,p,i}\|\,\|\mathbf{v}_{b,p,j}\|}, 
\qquad d_{a,b,p,i,j}=1-s_{a,b,p,i,j}, 
\qquad 1\le i,j\le N,
\]
and averaged to obtain a prompt-level distance
\[
D_{a,b,p}  =  \frac{1}{N^{2}}\sum_{i=1}^{N}\sum_{j=1}^{N} d_{a,b,p,i,j}.
\]
Aggregating over prompts gives
\[
\overline{D}_{a,b}  =  \frac{1}{P}\sum_{p=1}^{P} D_{a,b,p}.
\]
The quantity $\overline{D}_{a,b}$ is symmetric, equals $0$ only when the two models produce identical embeddings for every prompt, and increases as the models explore more disjoint regions of the solution space. Larger $\overline{D}_{a,b}$ values indicate greater dissimilarity; in practice, pairing such models can expand \emph{coverage of the idea space} (i.e., a wider range of distinct concepts and perspectives for the same brief) and reduce \emph{redundancy in candidate responses} (i.e., fewer semantically near-duplicate options).

\subsection{Results}
\subsubsection{Intra Model Diversity}
\begin{figure}[H]
  \centering
  \begin{subfigure}[t]{0.32\textwidth}
    \centering
    \includegraphics[width=\linewidth]{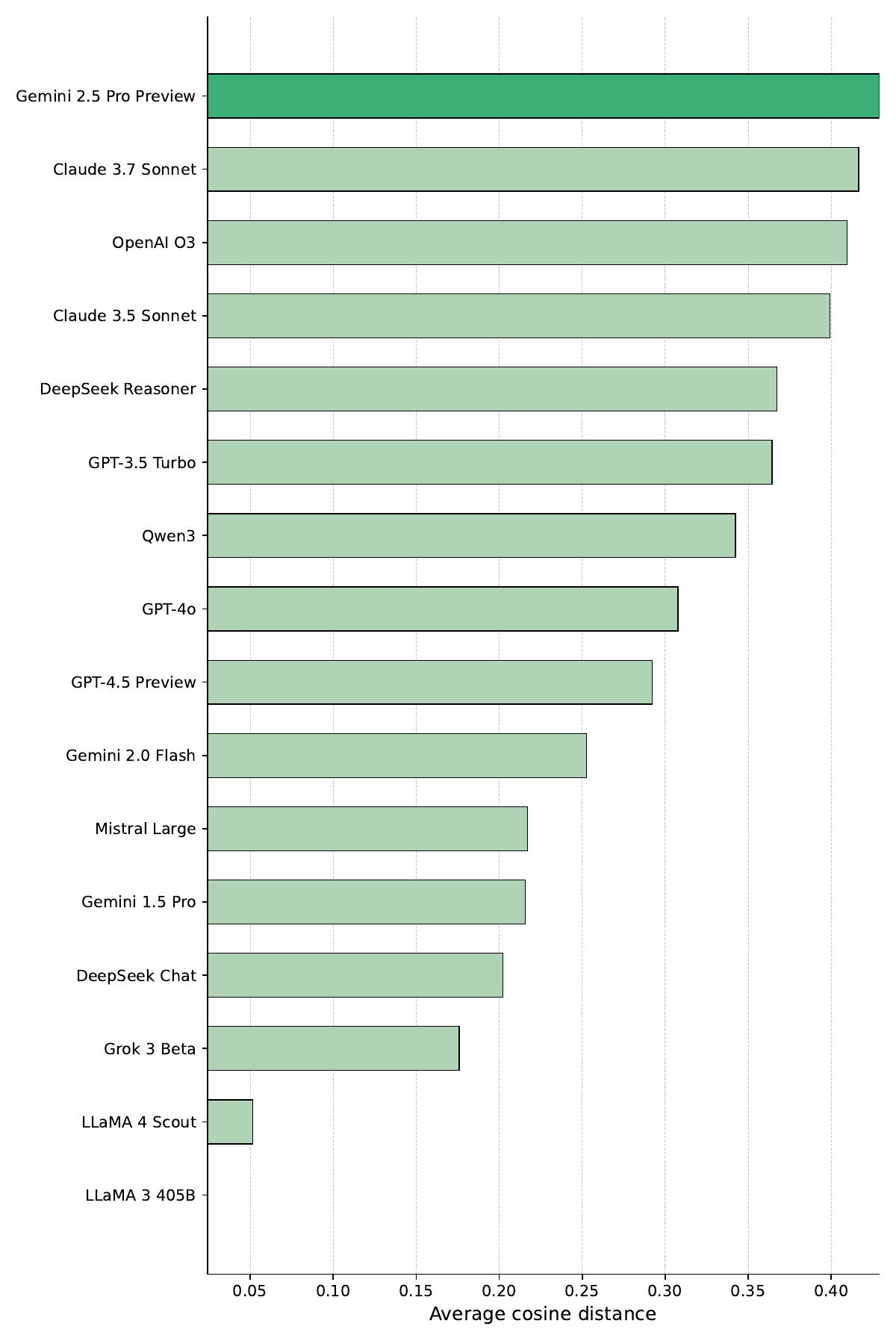}
    \caption{Insights}
  \end{subfigure}\hfill
  \begin{subfigure}[t]{0.32\textwidth}
    \centering
    \includegraphics[width=\linewidth]{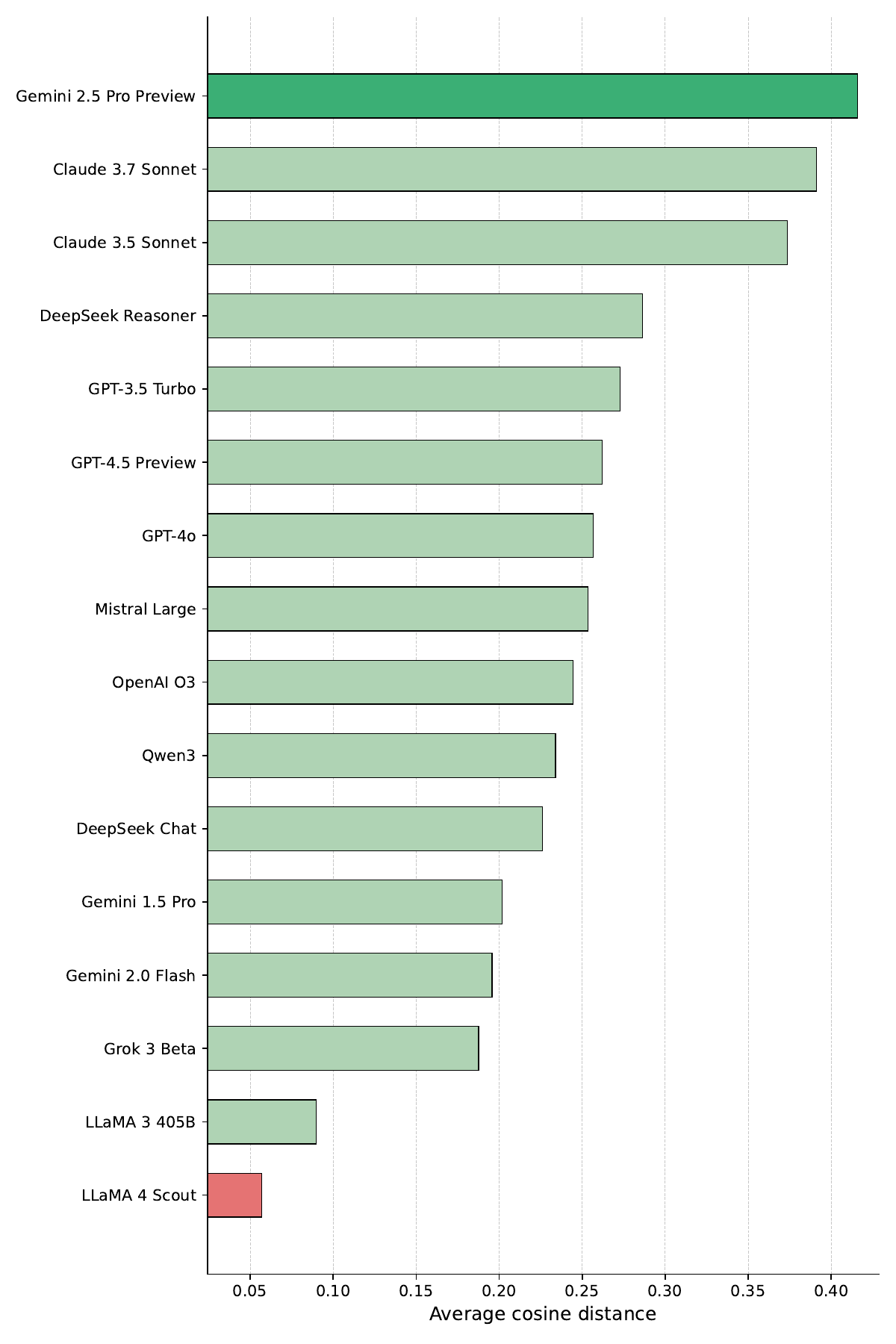}
    \caption{Ideas}
  \end{subfigure}\hfill
  \begin{subfigure}[t]{0.32\textwidth}
    \centering
    \includegraphics[width=\linewidth]{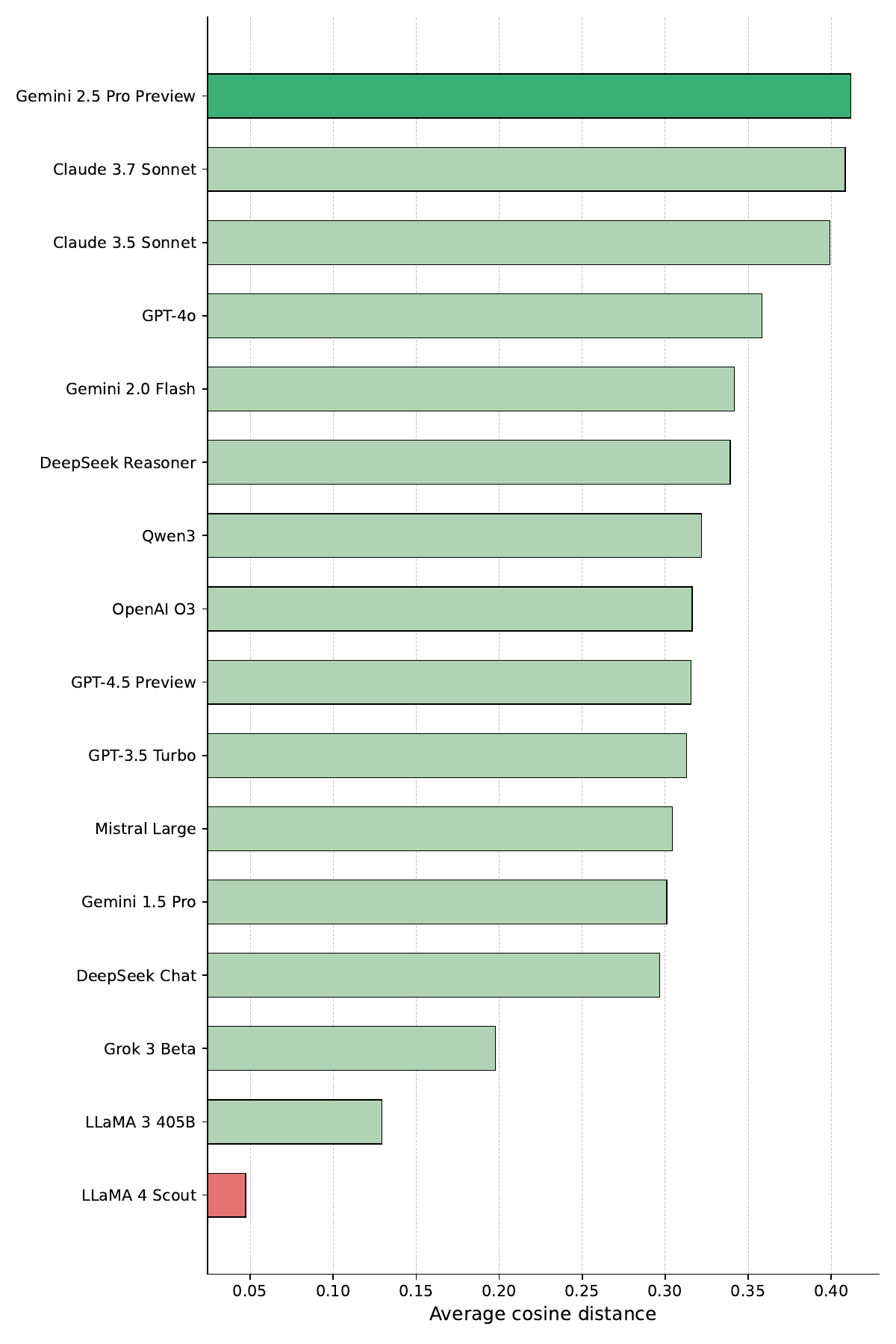}
    \caption{Wild Ideas}
  \end{subfigure}

  \caption{Intra-model diversity: average cosine distance between pairs of responses produced by the same model for the same prompt. Higher values indicate greater diversity.}
  \label{fig:intra_model_diversity}
\end{figure}

Intra-model diversity, which measures how much a model varies its outputs when the prompt is held fixed, is shown in \Cref{fig:intra_model_diversity}. Higher values mean the model explores distinct options rather than repeating near-duplicates. Gemini 2.5 Pro Preview and Claude 3.7 Sonnet show the highest diversity across Ideas, Insights, and Wild Ideas, consistently offering the broadest assortment of alternatives. OpenAI O3 is similarly diverse on Insight prompts but only moderate on Ideas and Wild Ideas, which suggests strength in exploring insights rather than in originating entirely new ideas. At the lower end, Grok 3 Beta and the two LLaMA baselines tend to repeat themselves, limiting their usefulness for divergent ideation. Finally, we find no systematic relationship between explicit reasoning tuning and diversity, reasoning-focused models are neither uniformly high nor uniformly low on this metric.

\subsubsection{Inter Model Diversity}
\begin{figure}[H]
  \centering

  \begin{subfigure}[t]{0.48\textwidth}
    \centering
    \includegraphics[width=\linewidth]{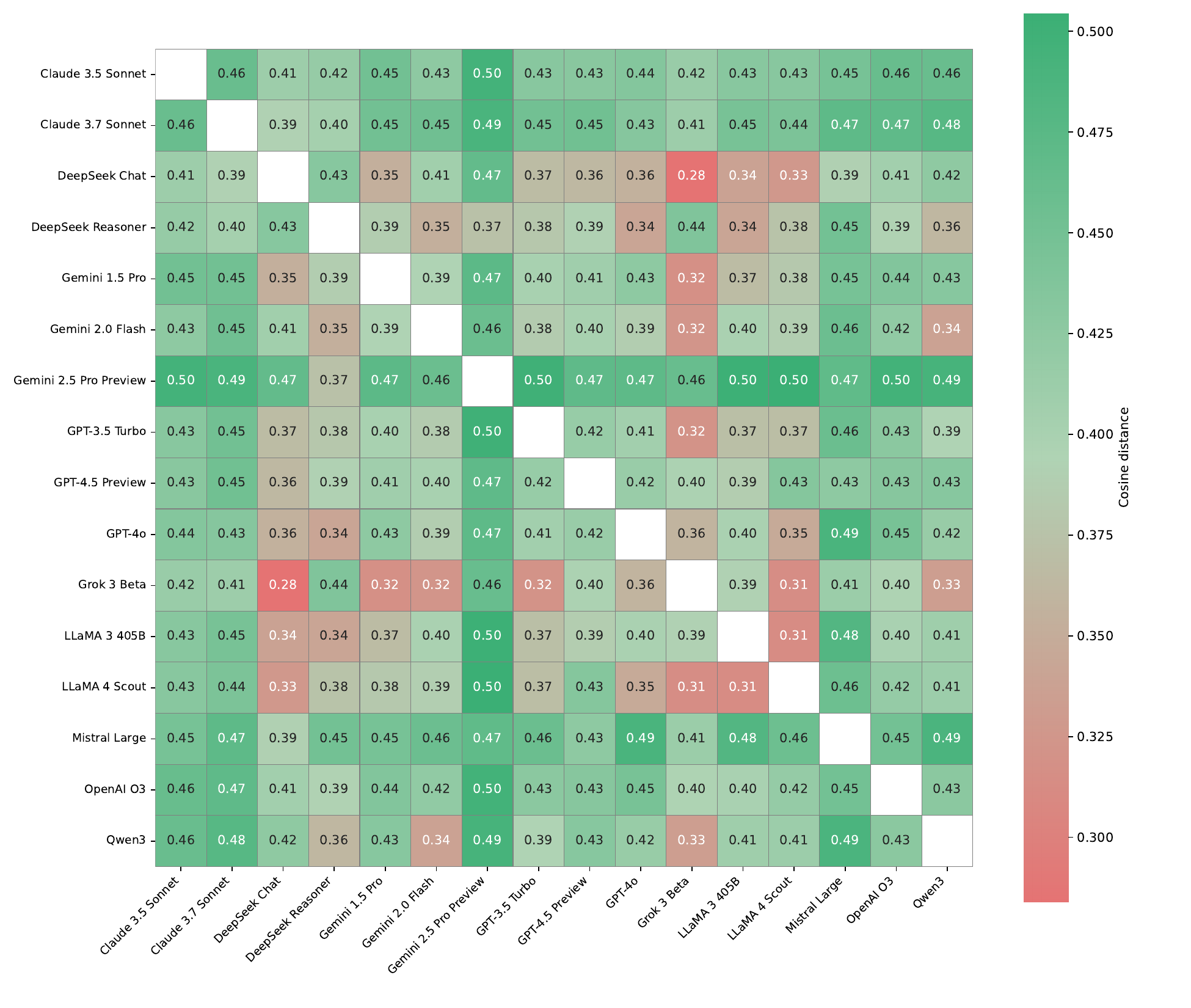}
    \caption{Insights}
  \end{subfigure}\hfill
  \begin{subfigure}[t]{0.48\textwidth}
    \centering
    \includegraphics[width=\linewidth]{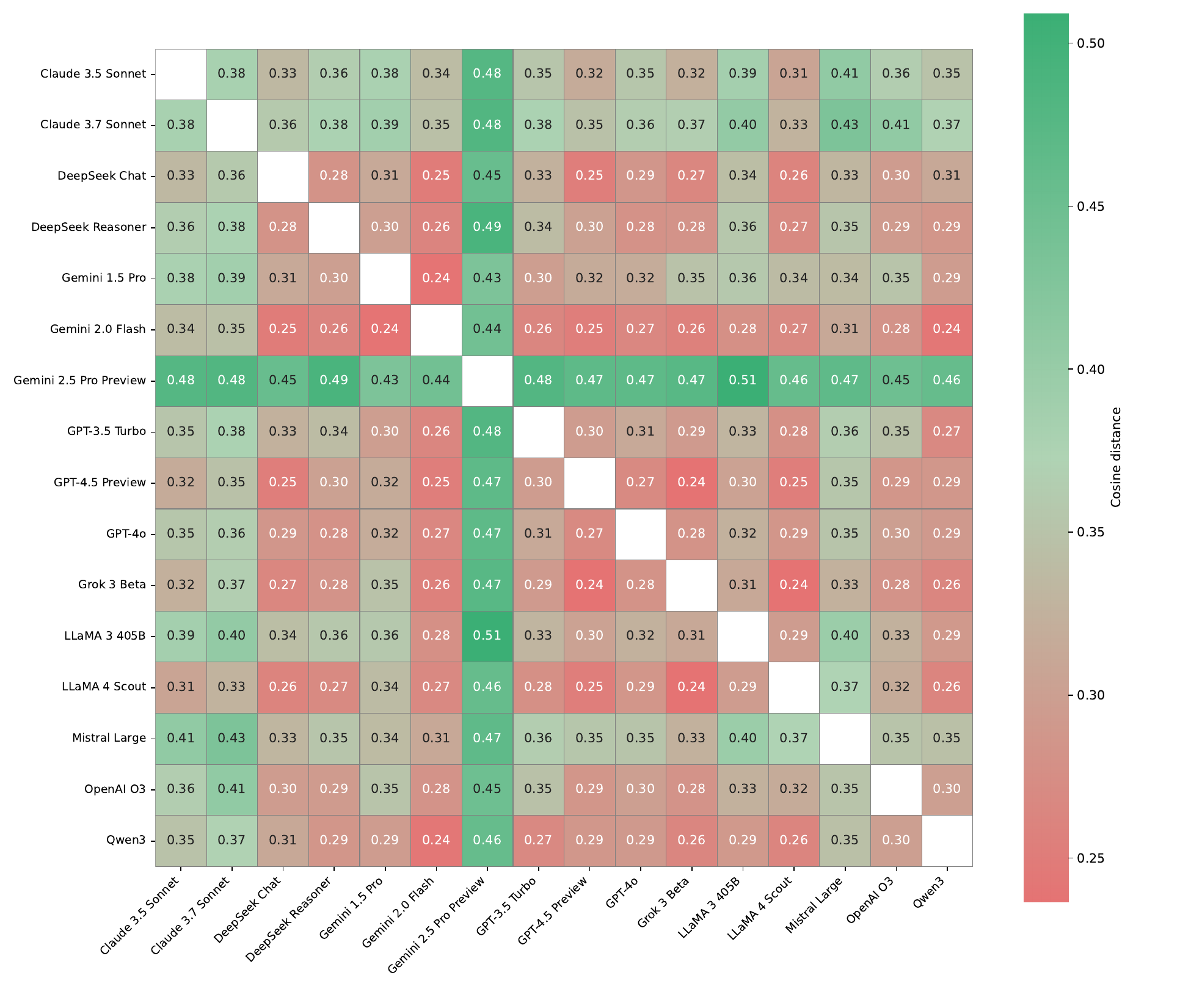}
    \caption{Ideas}
  \end{subfigure}

  \par\medskip

  \begin{subfigure}[t]{0.48\textwidth}
    \centering
    \includegraphics[width=\linewidth]{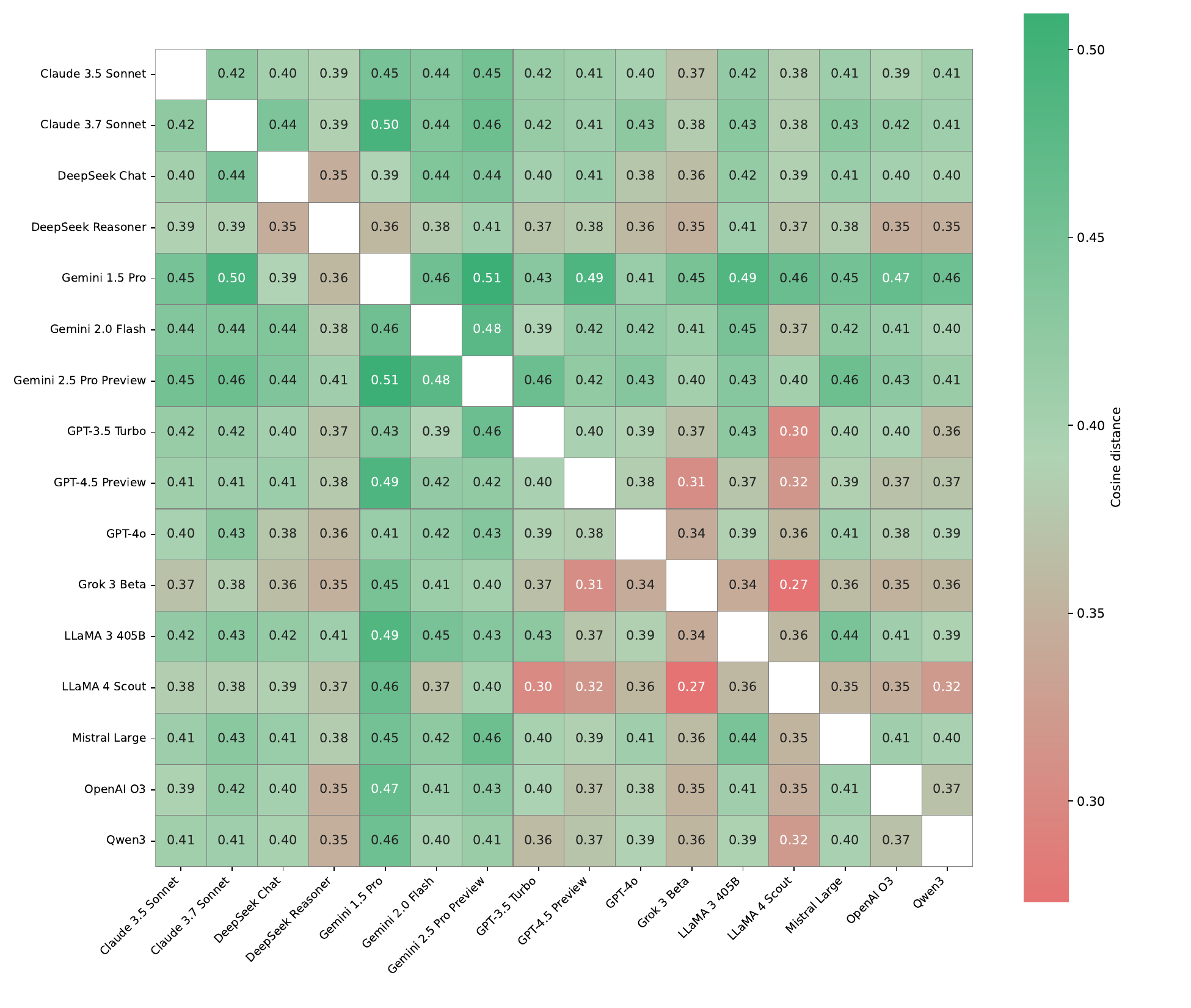}
    \caption{Wild Ideas}
  \end{subfigure}

  \caption{Inter-model diversity: average cosine distance between responses generated by different models for the same prompt. Higher values indicate that the models explore more distinct regions of the idea space.}
  \label{fig:inter_model_diversity}
\end{figure}

Inter-model diversity quantifies how far apart two models’ outputs are for the same prompt, measured as the average cosine distance between their response embeddings (higher means more distinct). As shown in \Cref{fig:inter_model_diversity}, distances are markedly lower for Ideas than for Wild Ideas, which indicates that the framing of the prompt as a request for wild ideas elicits greater divergence in the generative space. Gemini 1.5 Pro is a clear example of prompt sensitivity: it overlaps heavily with peers on Ideas (low distances) but separates sharply on Wild Ideas, suggesting broader exploration when constraints are loosened. By contrast, Mistral maintains consistently high distances across categories. Although its intra-model spread is limited, this persistent orthogonality to other models implies that including Mistral in an ensemble is likely to expand the option set and raise overall ideational novelty.

\subsubsection{Diversity across prompt types}

\begin{figure}[H]
    \centering
    \includegraphics[width=0.85\textwidth]{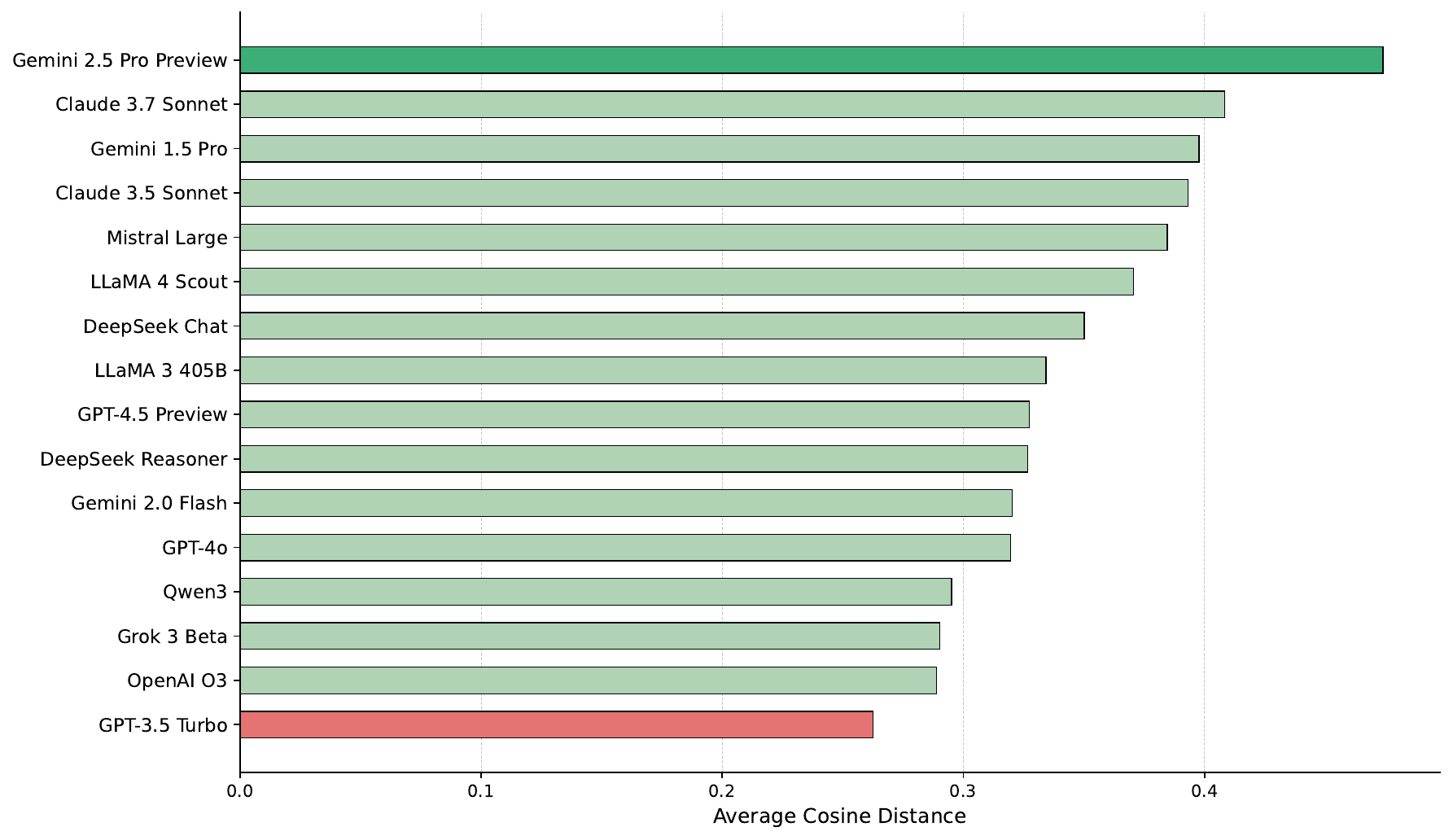}
    \caption{Average nearest-neighbour cosine distance between each model’s \emph{Ideas} and \emph{Wild Ideas} response sets (symmetrised; higher indicates a larger semantic shift).}
    \label{fig:average_prompt_type_distance}
\end{figure}

To quantify how strongly a model reshapes its outputs when switching from \emph{Ideas} to \emph{Wild Ideas}, we computed a symmetrised nearest-neighbour cosine distance between the two response sets for each model. For every \emph{Ideas} response we found its nearest neighbour in \emph{Wild Ideas}, and vice versa, then average the two directional means. Larger values indicate that the model moves into a different lexical and conceptual region rather than reusing the same templates.

As shown in \Cref{fig:average_prompt_type_distance}, Gemini 2.5 Pro exhibits the largest shift ($0.474$), suggesting strong adaptation to more divergent prompts. Gemini 1.5 Pro is also high ($0.398$), reinforcing that this family responds to looser brief constraints. At the lower end, GPT-3.5 Turbo shows the smallest shift ($0.262$), and OpenAI O3 is similarly low ($0.289$), indicating that simple prompt reframing has limited influence on their output space; a wider spread of ideas may require higher sampling temperatures or explicit stylistic cues. Overall, models differ not only in how many distinct options they produce for a single prompt (intra-prompt diversity), but also in how much they change those options when the brief becomes more open (inter-prompt shift). Higher inter-prompt distances indicate greater sensitivity to prompt framing and a wider range of new ideas.

\subsubsection{Brand Diversity}

\begin{figure}[H]
  \centering
  \begin{subfigure}[t]{0.32\textwidth}
    \centering
    \includegraphics[width=\linewidth]{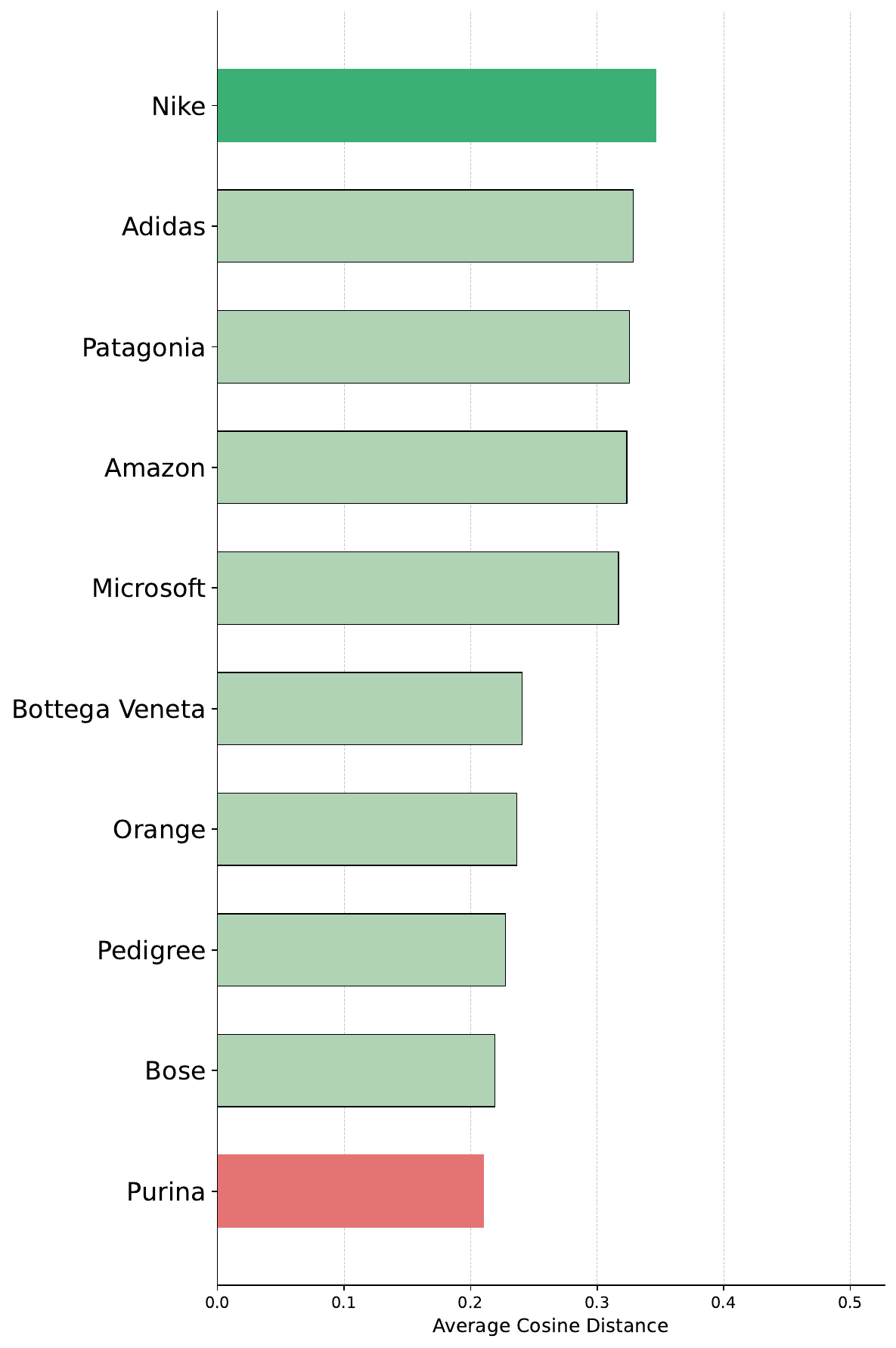}
    \caption{Insights}
  \end{subfigure}\hfill
  \begin{subfigure}[t]{0.32\textwidth}
    \centering
    \includegraphics[width=\linewidth]{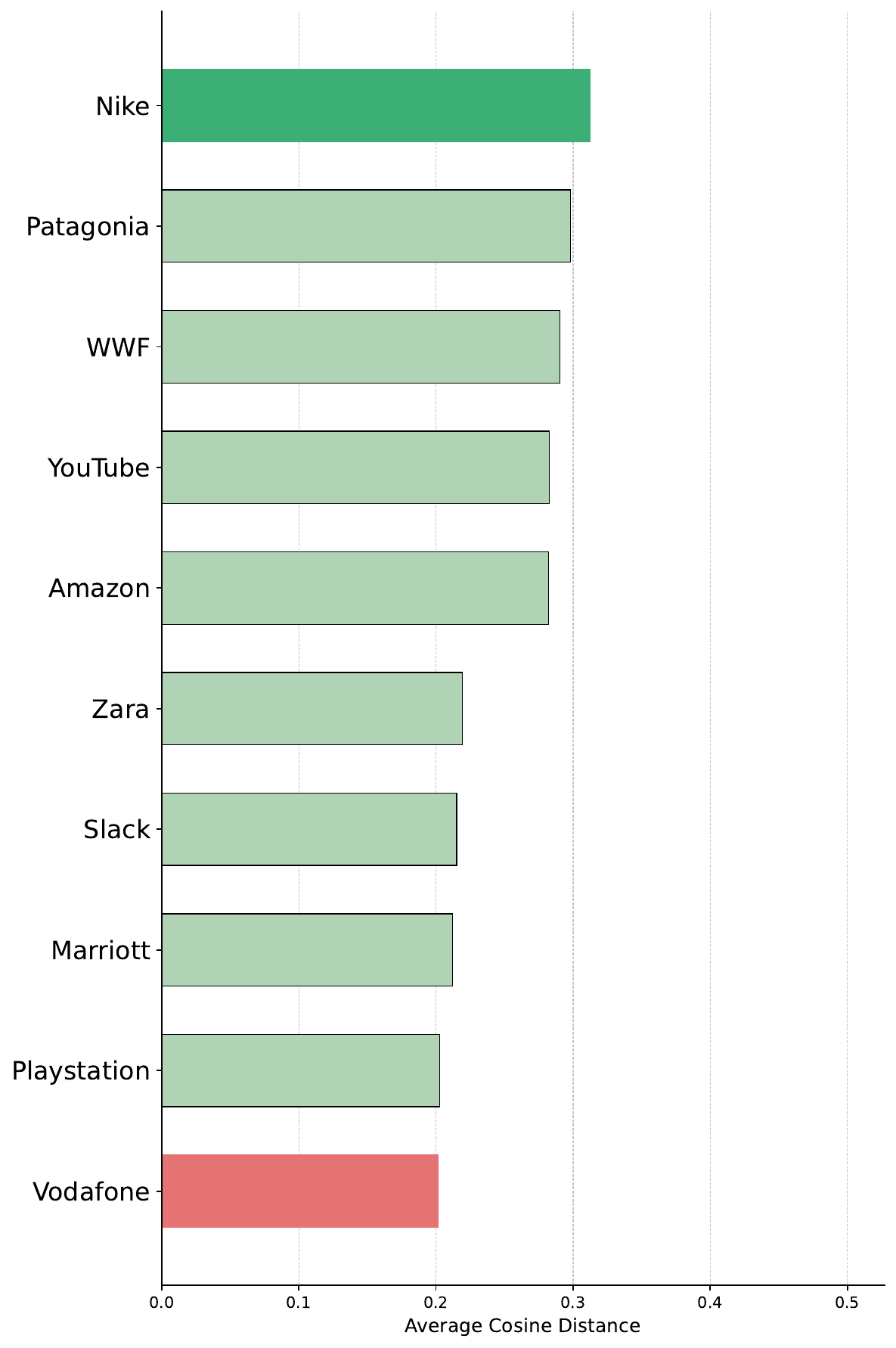}
    \caption{Ideas}
  \end{subfigure}\hfill
  \begin{subfigure}[t]{0.32\textwidth}
    \centering
    \includegraphics[width=\linewidth]{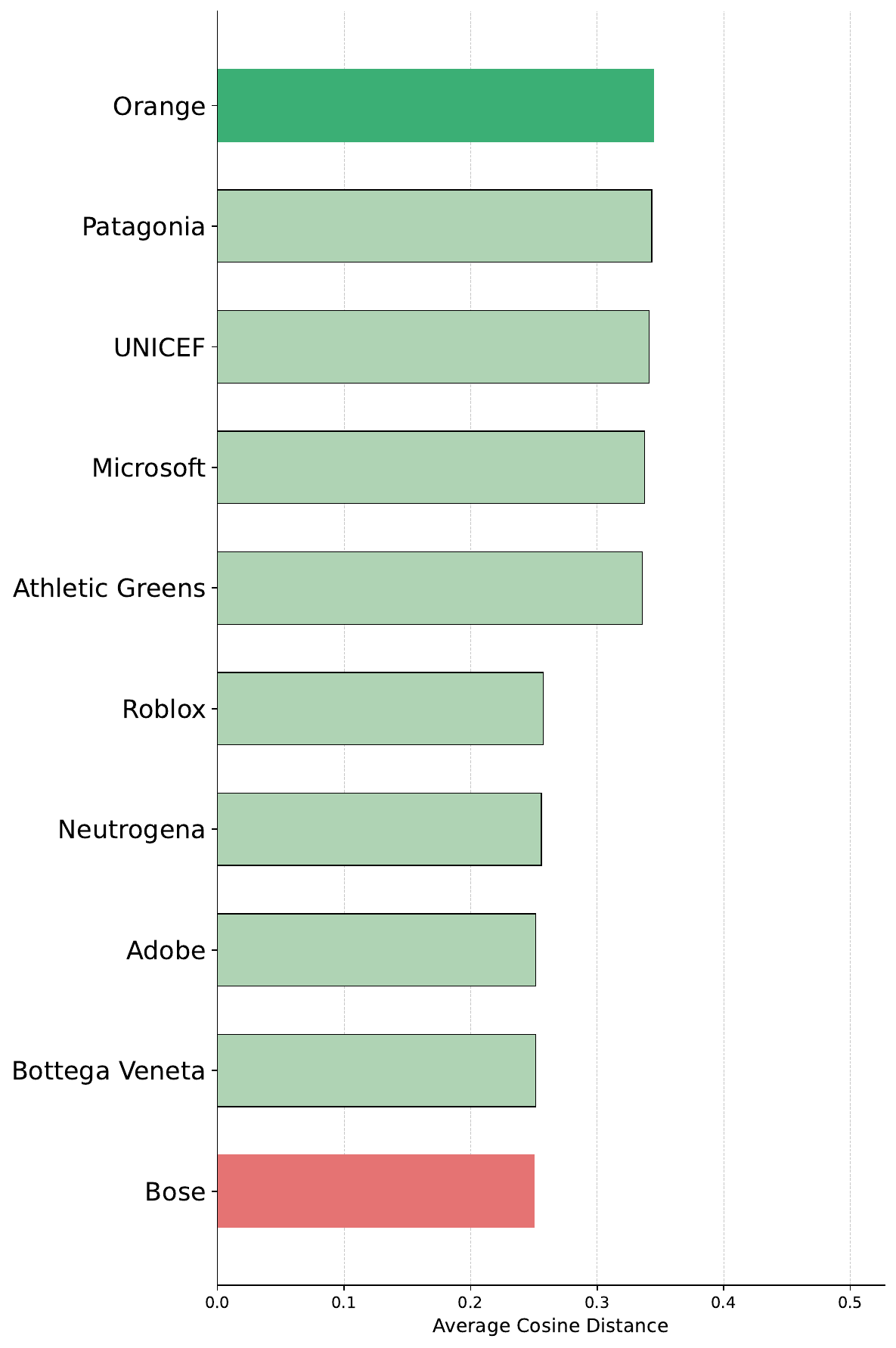}
    \caption{Wild Ideas}
  \end{subfigure}

  \caption{Top vs. bottom brand sets: inter-model diversity by prompt type. Higher average cosine distance indicates greater semantic separation.}
  \label{fig:top_bottom_brands_diversity}
\end{figure}

Brand-level cosine distances vary much more widely, from about $0.19$ to $0.34$ across the three prompt types. This dispersion suggests that brand identity interacts with model priors in non-uniform ways. Two mechanisms likely contribute:
\begin{itemize}
  \item \textbf{Training data coverage.} Well-known consumer brands (for example, Nike and Patagonia) are heavily represented in public text. Models learn richer, more varied associations for these brands, which can be recombined into more distinct responses, increasing cosine distances.
  \item \textbf{Constraints embedded in brand positioning.} Strong, distinctive positionings (for example, sustainability for Patagonia and performance for Nike) tighten the feasible space but can also encourage more polarized exploration when moving from conventional ideas to wild ideas or concise insights. In contrast, commodity or B2B brands (for example, Vodafone and Purina) provide fewer unique anchors, which yields greater semantic convergence and lower diversity.
\end{itemize}

\subsection{Implications}

\begin{itemize}
  \item \textbf{Model selection.} If breadth of ideas is the priority, Gemini 2.5 Pro and Claude 3.7 Sonnet exhibit higher intra-prompt diversity across tasks, with Claude 3.5 Sonnet a close alternative.
  
  \item \textbf{Ensemble strategy.} Pair models with high inter-model distance to expand coverage without large cost increases. For example, Mistral with Gemini 1.5 Pro provides complementary option sets.
  
  \item \textbf{Prompt framing.} Reframing from conventional \emph{Ideas} to \emph{Wild Ideas} reliably increases divergence between models. Use this shift to surface novel directions, then refine with tighter briefs.
  
\item \textbf{Brand context.} Category effects are small on average, but brand-specific cues can widen or narrow the idea space. Individual brand testing may be warranted.
    
  \item \textbf{Quality control.} High cosine distance indicates diversity, not utility. Combine diversity measures with quality filters (human review) that score novelty, appropriateness to brief, and feasibility.
\end{itemize}

\section{LLM-as-judge}
\subsection{Introduction}
Human evaluation of large language model outputs is accurate but expensive and slow, which has motivated the use of LLM-as-judge that score or rank responses automatically. Prior work shows that strong judges can approximate human preferences on open-ended tasks and multi-turn dialogues, for example GPT-4 on MT-Bench and Chatbot Arena reports agreement levels comparable to human-human agreement \cite{zheng2023judgingllmasajudgemtbenchchatbot}. Other studies propose rubric-driven evaluators such as G-Eval and Prometheus, which improve correlation with human ratings in summarization, dialogue, and long-form evaluation \cite{lan2024criticevalevaluatinglargelanguage}. 

At the same time, several analyses caution that judge performance varies by task, dataset, and prompt design, and that evaluators may exhibit position or self-enhancement biases and reduced reliability on model-generated text \cite{bavaresco2025llmsinsteadhumanjudges}. These findings suggest that LLM-as-judge requires task-specific validation.

In this part, we evaluate three LLM-as-judge models under three judging prompts on brand-focused creative tasks. We compare judge-derived rankings with human rankings from Part A. This design tests whether automated judgments recover human preferences in a domain where novelty and brief fit are critical.

\subsection{Methods}

\subsubsection{Model responses}
We evaluated the same $11{,}012$ response pairs used in the human study. Each pair contained two anonymized model outputs to the same brand brief, drawn from three tasks: \emph{Insight}, \emph{Ideas}, and \emph{Wild Ideas}.

\subsubsection{Judge models}

\paragraph{Judge set and rationale.}
We used three independent LLM judges (GPT-4o, Claude 4 Sonnet, and Gemini 2.0 Flash) to increase validity and robustness (see \Cref{tab:judge_models}). Using multiple judges serves three goals:
\begin{enumerate}
  \item estimate cross-judge reliability and the stability of model rankings.
  \item separate sensitivity to judge identity from sensitivity to prompt wording by crossing judges with the three judging prompts.
  \item avoid dependence on a single proprietary LLM.
\end{enumerate}

\begin{table}[H]
  \centering
  \caption{LLM-as-judge models and API identifiers.}
  \label{tab:judge_models}
  \begin{tabular}{lll}
    \toprule
    \textbf{Judge name} & \textbf{API model ID}        & \textbf{Vendor} \\
    \midrule
    GPT-4o              & gpt-4o-2024-11-20            & OpenAI \\
    Claude 4 Sonnet     & claude-sonnet-4-20250514     & Anthropic \\
    Gemini 2.0 Flash    & models/gemini-2.0-flash-001  & Google \\
    \bottomrule
  \end{tabular}
\end{table}

\subsubsection{Judge prompts}
We tested three system prompts for the judge:
\begin{itemize}
  \item \textbf{Creative Strategist:} based on a creative-brief template drafted by agency strategists.
  \item \textbf{Surprising Ideas:} explicitly rewards surprising and unconventional ideas that remain on brief.
  \item \textbf{EQ Benchmark:} adapted from the EQ-Bench creativity prompt, balancing creativity and craft.
\end{itemize}
Full prompt text is provided in the appendix \Cref{app:judge_prompts}.

\subsubsection{Dual-vote aggregation}
Each pair was evaluated twice, with the display order reversed on the second pass. If the same response was selected in both orders, it was counted as the winner; otherwise, we recorded a tie.

\subsubsection{Alignment measurement}
We quantified how closely each LLM-as-judge reproduces human preferences using \textbf{Spearman’s rank correlation} ($\rho$) between judge- and human-derived model rankings. For each prompt type $P \in \{\textit{Insight}, \textit{Ideas}, \textit{Wild Ideas}\}$:
\begin{enumerate}[nosep,leftmargin=*]
  \item Fit Bradley-Terry models to the human pairs to obtain per-model scores $\hat{\theta}^{(H,P)}_m$, and to the judge pairs (after dual-vote aggregation) to obtain $\hat{\theta}^{(J,P)}_m$.
  \item Convert scores to ranks $r^{(H,P)}_m=\mathrm{rank}(\hat{\theta}^{(H,P)}_m)$ and $r^{(J,P)}_m=\mathrm{rank}(\hat{\theta}^{(J,P)}_m)$
  \item Compute
  \[
  \rho(P)\;=\;\mathrm{corr}_s\!\bigl(\{r^{(H,P)}_m\},\,\{r^{(J,P)}_m\}\bigr)
  \]
  on the intersection of models scored by both methods.
\end{enumerate}
Here, $\rho=1$ indicates perfect ordinal agreement, $\rho=0$ indicates no monotonic association, and $\rho<0$ indicates inverse ordering.

\subsection{Results}
\subsubsection{Creative Strategist System Prompt}
\begin{figure}[H]
  \centering
  \begin{subfigure}[t]{0.24\textwidth}
    \centering
    \includegraphics[width=\linewidth]{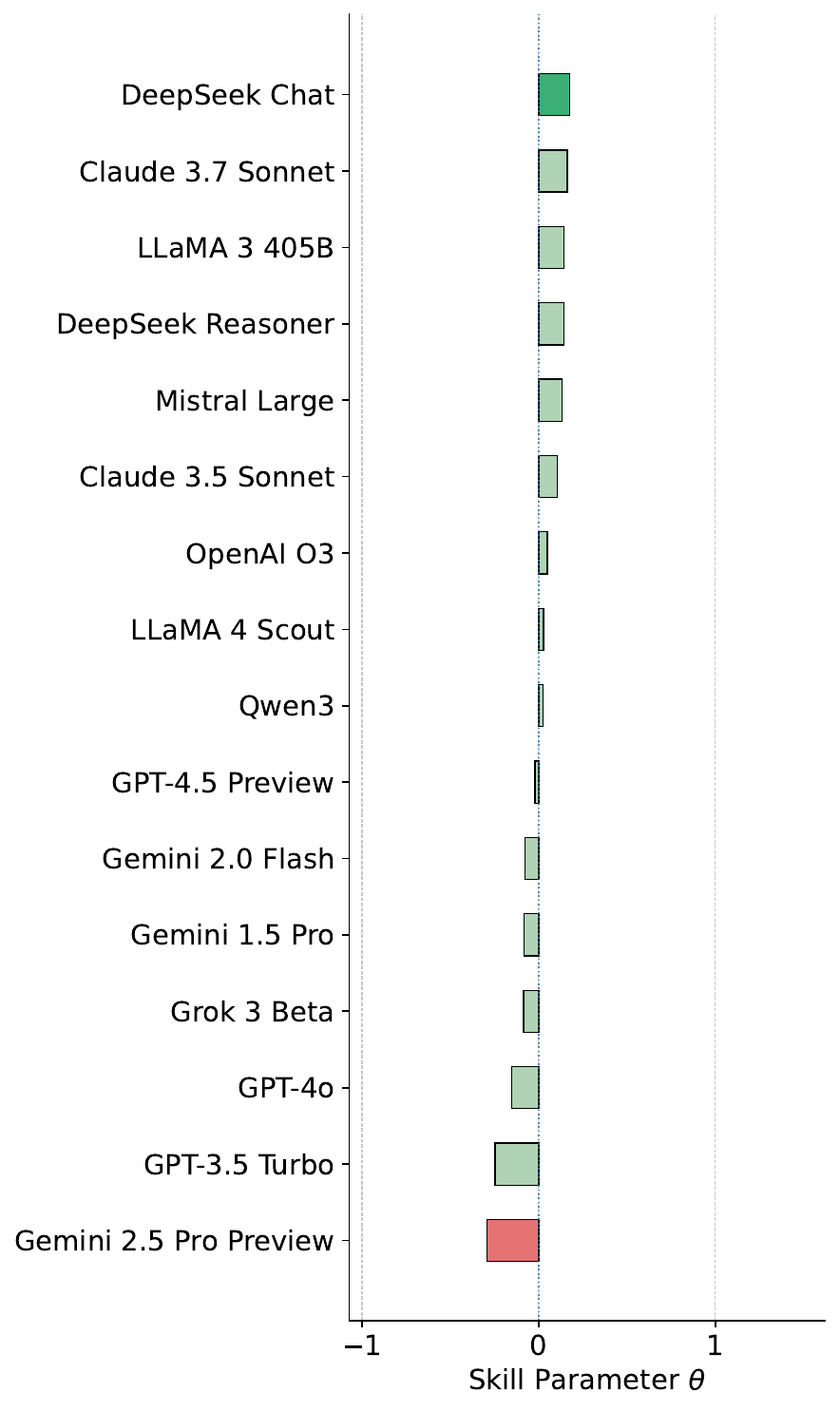}
    \caption{Human raters}
  \end{subfigure}\hfill
  \begin{subfigure}[t]{0.24\textwidth}
    \centering
    \includegraphics[width=\linewidth]{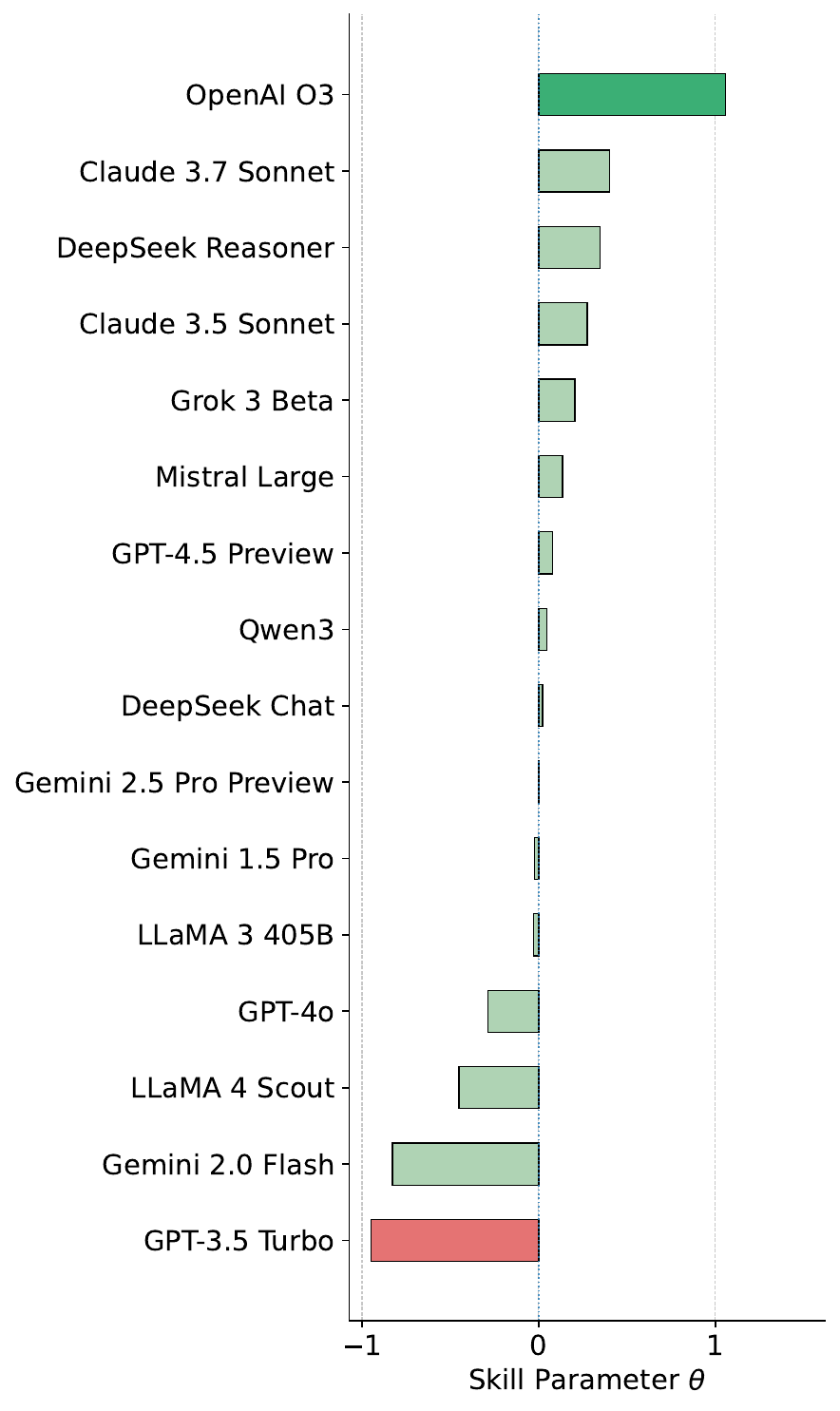}
    \caption{Claude 4 Sonnet (judge)}
  \end{subfigure}\hfill
  \begin{subfigure}[t]{0.24\textwidth}
    \centering
    \includegraphics[width=\linewidth]{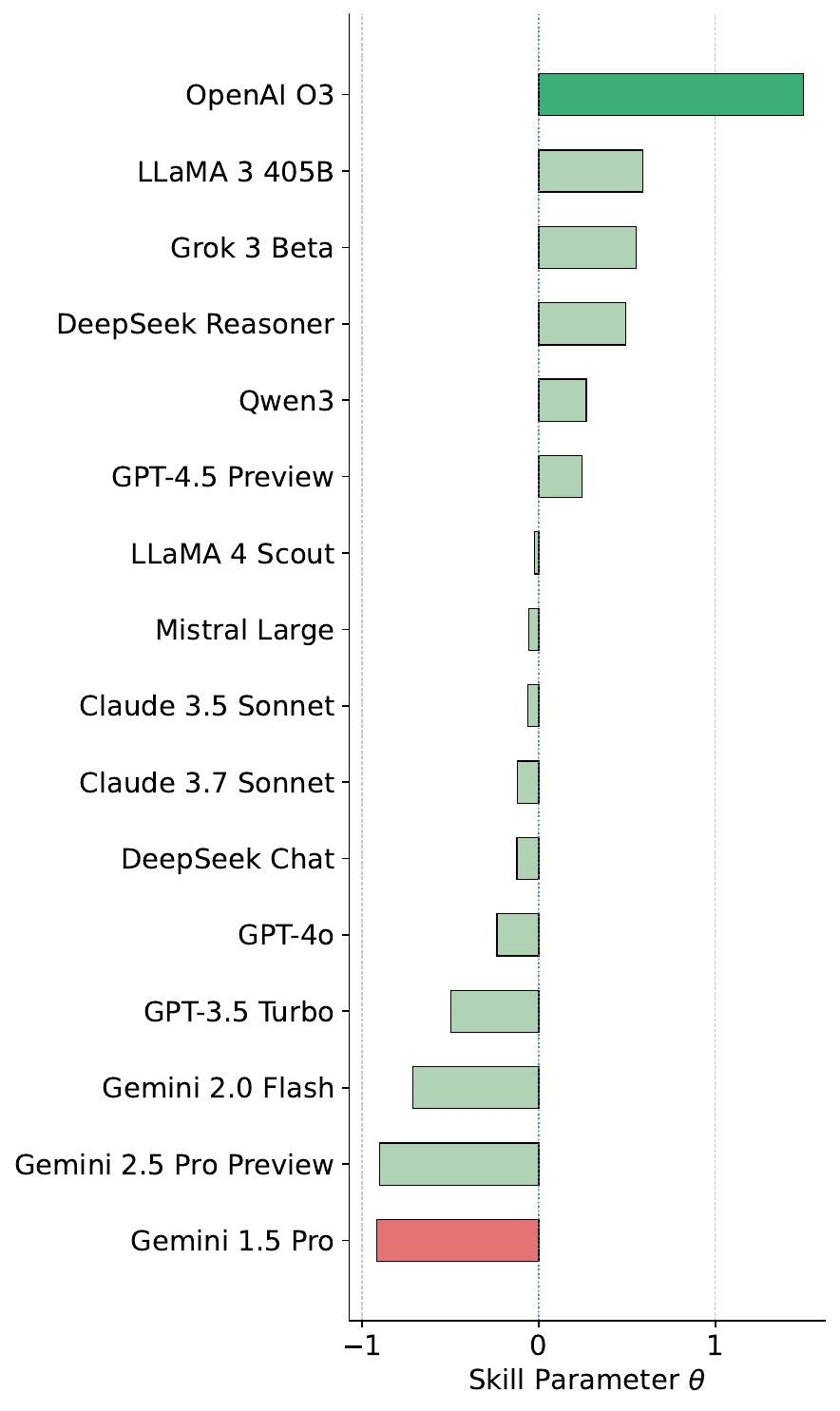}
    \caption{GPT-4o (judge)}
  \end{subfigure}\hfill
  \begin{subfigure}[t]{0.24\textwidth}
    \centering
    \includegraphics[width=\linewidth]{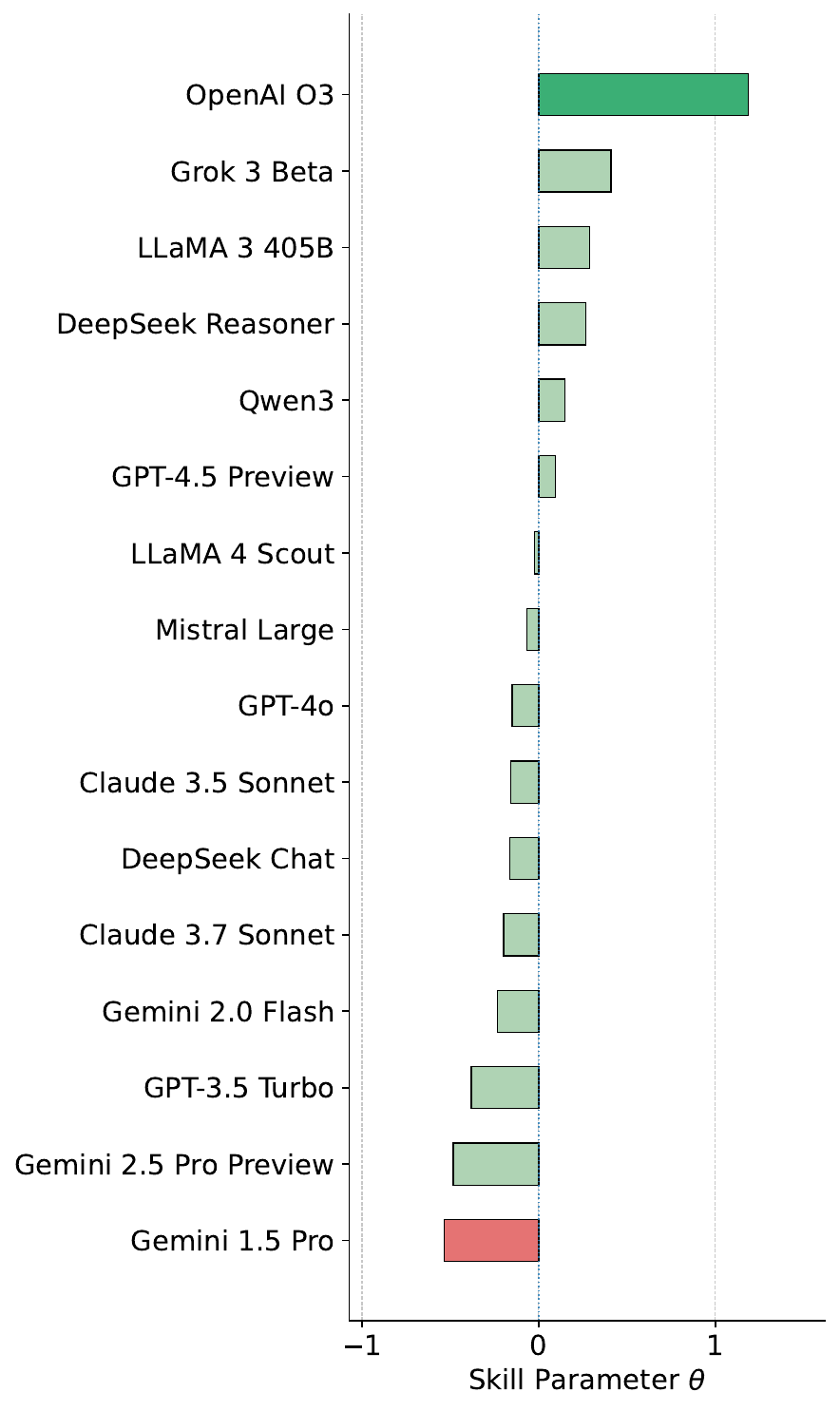}
    \caption{Gemini 2.0 Flash (judge)}
      \end{subfigure}

  \caption{Ideas task under the Creative Strategist system prompt. Bradley-Terry strengths ($\theta$) from human raters and three LLM-as-judge evaluators. Higher values indicate stronger preference.}
  \label{fig:bt_ideas_creative_judges_row}
\end{figure}

\begin{table}[H]
  \centering
  \small
\caption{Creative Strategist judge prompt: Spearman’s rank correlation ($\rho$) between LLM-as-judge and human model rankings by prompt type (higher indicates closer alignment with human rankings).}
  \label{tab:creative_spearman}
  \setlength{\tabcolsep}{8pt}
  \renewcommand{\arraystretch}{1.15}
  \begin{tabular}{lccc}
    \toprule
    \textbf{Prompt} & \textbf{Claude 4 Sonnet} & \textbf{GPT-4o} & \textbf{Gemini 2.0 Flash} \\
    \midrule
    Ideas      & 0.482 & 0.432 & 0.321 \\
    Wild Ideas & 0.597 & 0.409 & 0.338 \\
    Insight    & 0.309 & 0.168 & 0.129 \\
    \bottomrule
  \end{tabular}
\end{table}

\noindent\textit{What this judge rewards.}
A single clear idea stated directly, a confident tone, and simple structure. Entries that lead with the core thought, avoid hedging, and supply a brief rationale score higher.

\noindent\textit{Agreement with humans.}
The \emph{Ideas} panels in \Cref{fig:bt_ideas_creative_judges_row} illustrate the relative positions under this rubric, and \Cref{tab:creative_spearman} summarises alignment with human rankings. Agreement is at best moderate and varies by both judge and prompt type, which means judge choice is itself a major source of variance. In practice, LLM-as-judge is not a drop-in substitute for expert assessment here; it introduces its own preferences and is, at most, a coarse filter rather than a ranking authority.

\noindent\textit{Ranking patterns.}
Across \emph{Insight}, \emph{Ideas}, and \emph{Wild Ideas}, the three judges often place OpenAI O3 near the top, consistent with the rubric’s emphasis on clarity, structure, and brief adherence, although this association is descriptive rather than causal. The Gemini 2.0 Flash judge rates LLaMA 3.1 405B comparatively highly on \emph{Ideas}, suggesting a stylistic match to that evaluator. GPT-3.5 Turbo typically appears in the lower tail under both human and judge rankings. Effect sizes are modest and remain brand- and prompt-dependent. Corresponding panels for \emph{Insights} and \emph{Wild Ideas} are provided in the appendix (\Cref{fig:bt_insights_judges_row,fig:bt_wildideas_judges_row}).

\subsubsection{Surprise Prompt}
\begin{figure}[H]
  \centering
  \begin{subfigure}[t]{0.24\textwidth}
    \centering
    \includegraphics[width=\linewidth]{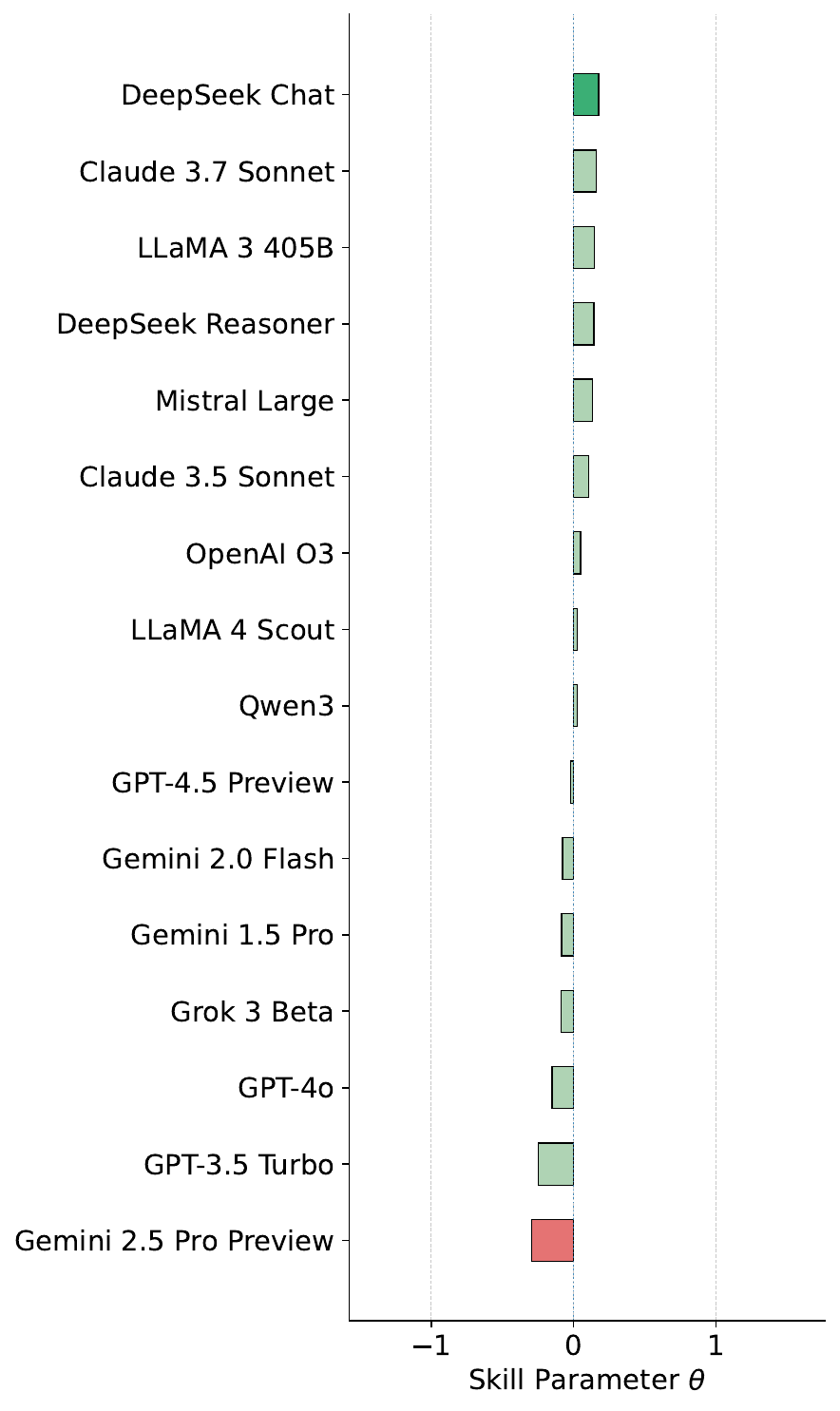}
    \caption{Human raters}
  \end{subfigure}\hfill
  \begin{subfigure}[t]{0.24\textwidth}
    \centering
    \includegraphics[width=\linewidth]{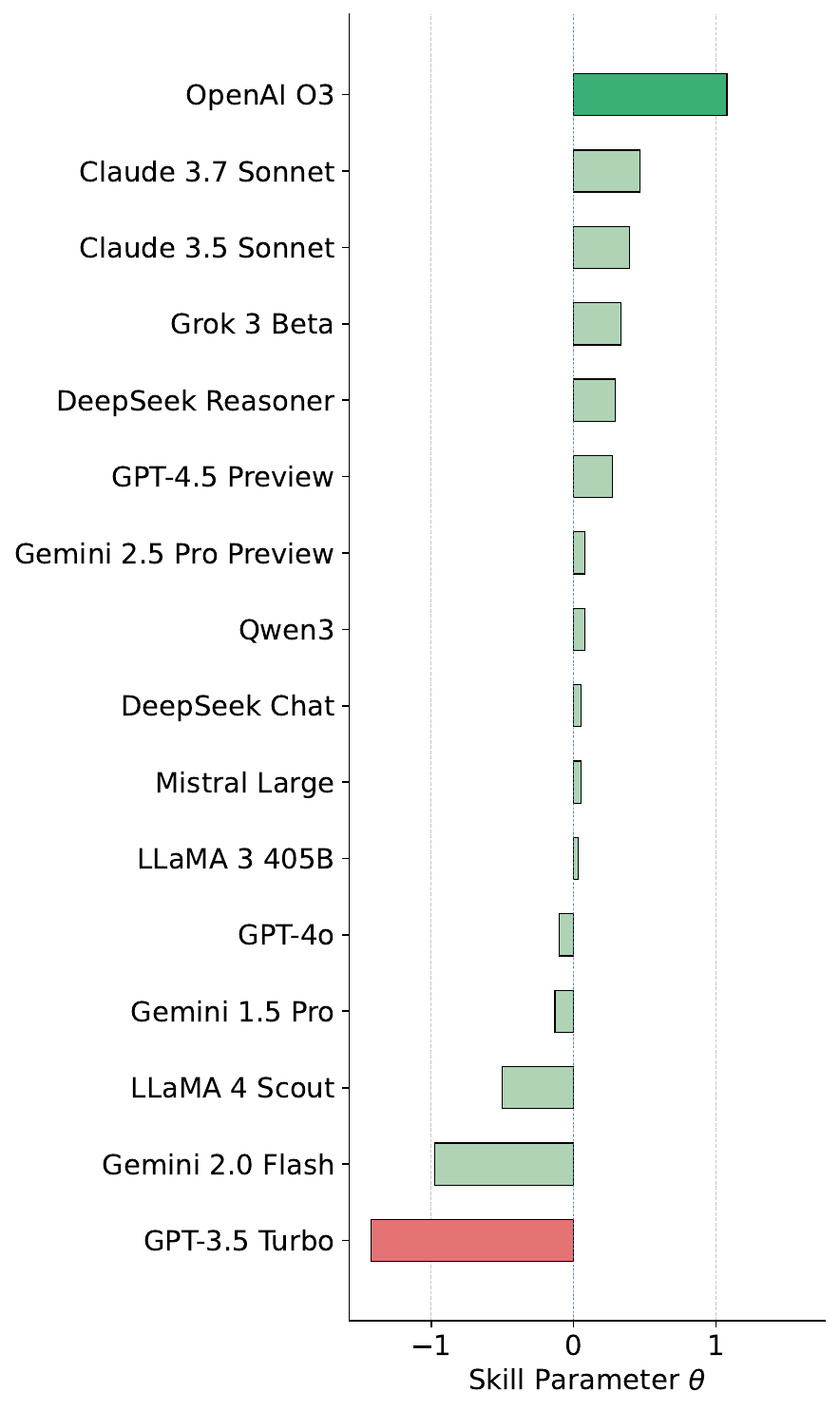}
    \caption{Claude 4 Sonnet (judge)}
  \end{subfigure}\hfill
  \begin{subfigure}[t]{0.24\textwidth}
    \centering
    \includegraphics[width=\linewidth]{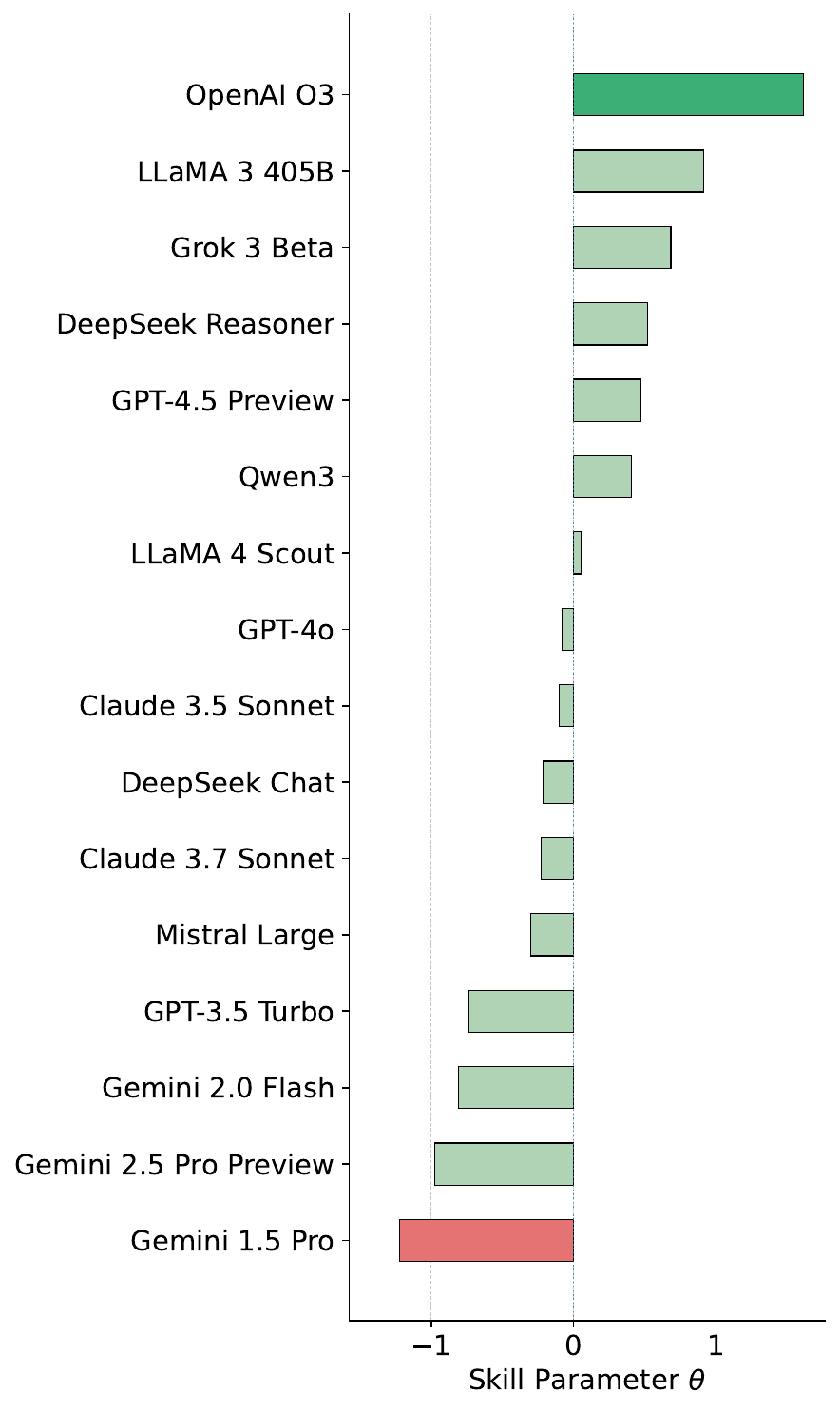}
    \caption{GPT-4o (judge)}
  \end{subfigure}\hfill
  \begin{subfigure}[t]{0.24\textwidth}
    \centering
    \includegraphics[width=\linewidth]{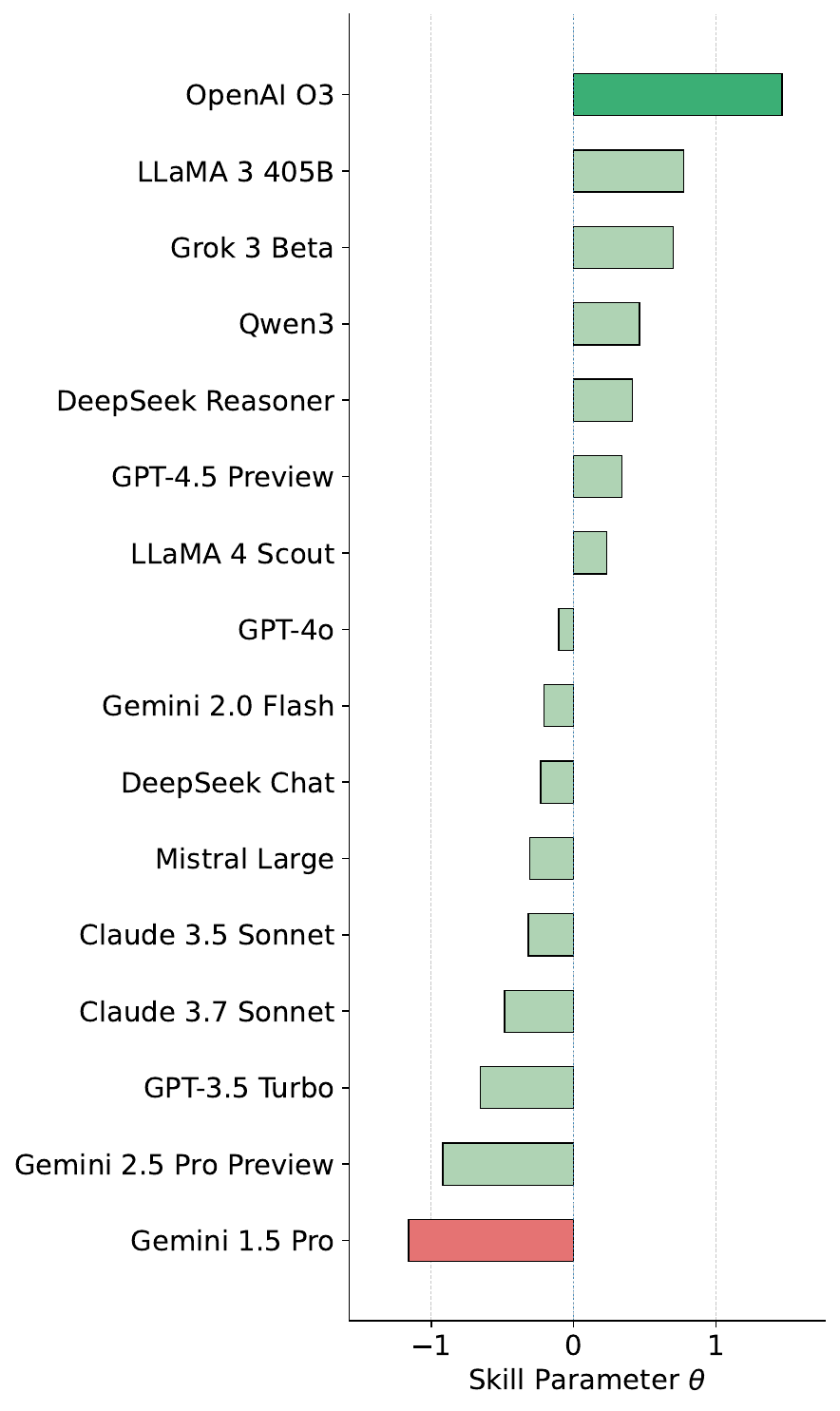}
    \caption{Gemini 2.0 Flash (judge)}
  \end{subfigure}
  \caption{Ideas task under the Surprise judge prompt. Bradley-Terry strengths ($\theta$) from human raters and three LLM-as-judge evaluators.}
  \label{fig:surprise_ideas_row}
\end{figure}

\noindent\textit{What this judge rewards.}
Originality above all. Unusual, contrarian, or counter-intuitive moves receive credit if the response stays on brief and remains intelligible. Minor rough edges in phrasing or structure are tolerated when the underlying idea is strong.

\begin{table}[H]
  \centering
  \small
  \caption{Surprise judge prompt: Spearman’s rank correlation ($\rho$) between LLM-as-judge and human model rankings by prompt type (higher indicates closer alignment with human rankings).}
  \label{tab:surprise_spearman}
  \setlength{\tabcolsep}{8pt}
  \renewcommand{\arraystretch}{1.15}
  \begin{tabular}{lccc}
    \toprule
    \textbf{Prompt} & \textbf{Claude 4 Sonnet} & \textbf{GPT-4o} & \textbf{Gemini 2.0 Flash} \\
    \midrule
    Ideas      & 0.359 & 0.326 & 0.247 \\
    Wild Ideas & 0.597 & 0.409 & 0.265 \\
    Insight    & 0.321 & 0.218 & 0.041 \\
    \bottomrule
  \end{tabular}
\end{table}

\noindent\textit{Agreement with humans.}
The correlations in \Cref{tab:surprise_spearman} indicate at best moderate alignment and strong dependence on both judge and prompt type. Agreement is highest for \emph{Wild Ideas} with Claude 4 Sonnet ($\rho=0.597$) and weakest for \emph{Insight} with Gemini 2.0 Flash ($\rho=0.041$). This pattern shows that changing the judging model or the rubric emphasis materially re-orders systems, so LLM-as-judge should be treated as a coarse filter rather than a substitute for expert assessment.

\noindent\textit{Ranking patterns.}
\Cref{fig:surprise_ideas_row} shows that orderings under the Surprise rubric are broadly similar to those under the Creative Strategist rubric, with small shifts at the margins. One consistent movement is a lift for LLaMA 3.1 on \emph{Insight} when originality is emphasised (see \Cref{fig:surprise_insights_row}), suggesting that an originality-first judge favours unconventional yet coherent responses. Corresponding panels for \emph{Insights} and \emph{Wild Ideas} are provided in the appendix (\Cref{fig:surprise_insights_row,fig:surprise_wildideas_row}).

\subsubsection{EQ Bench system prompt}
\begin{figure}[H]
  \centering
  \begin{subfigure}[t]{0.24\textwidth}
    \centering
    \includegraphics[width=\linewidth]{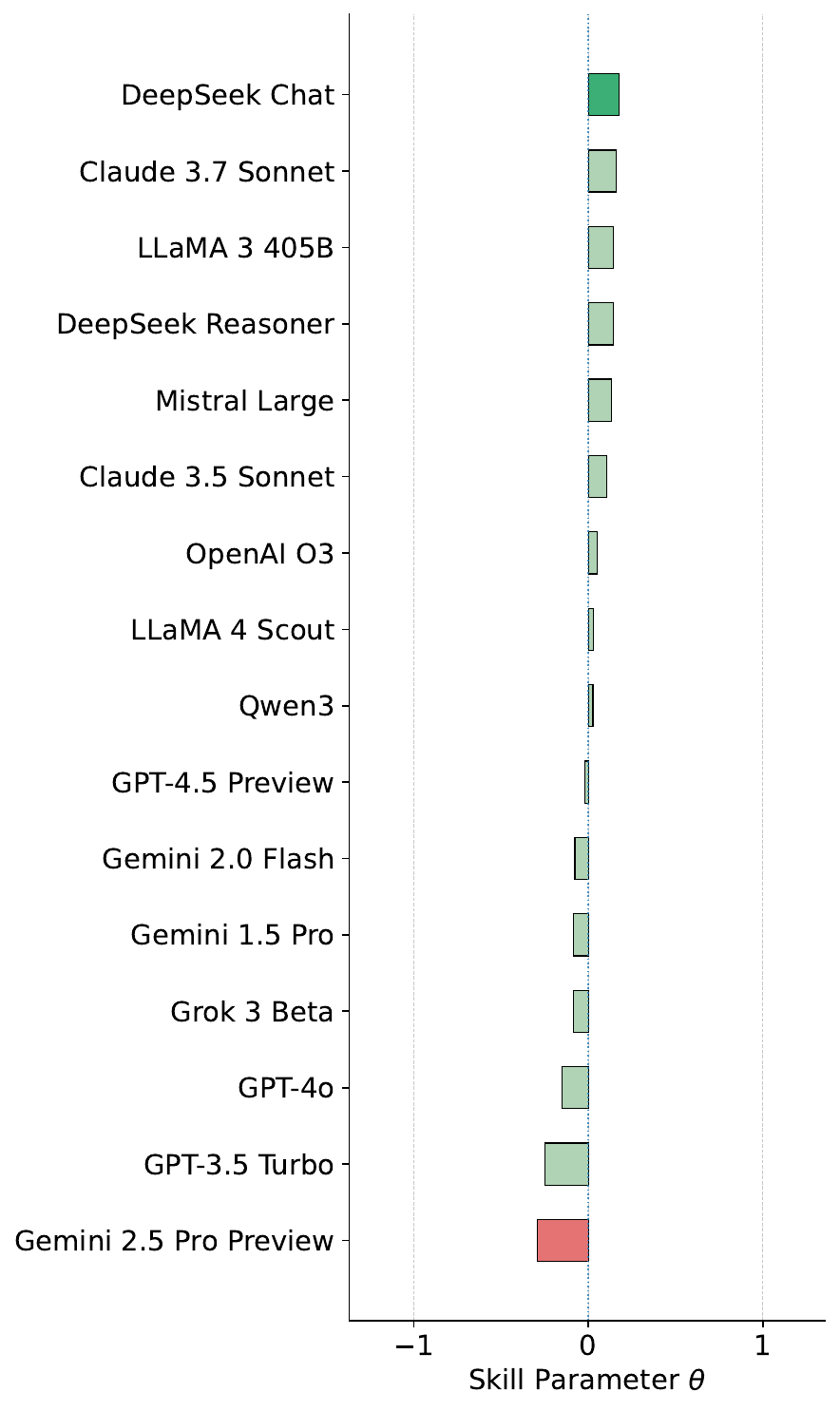}
    \caption{Human raters}
  \end{subfigure}\hfill
  \begin{subfigure}[t]{0.24\textwidth}
    \centering
    \includegraphics[width=\linewidth]{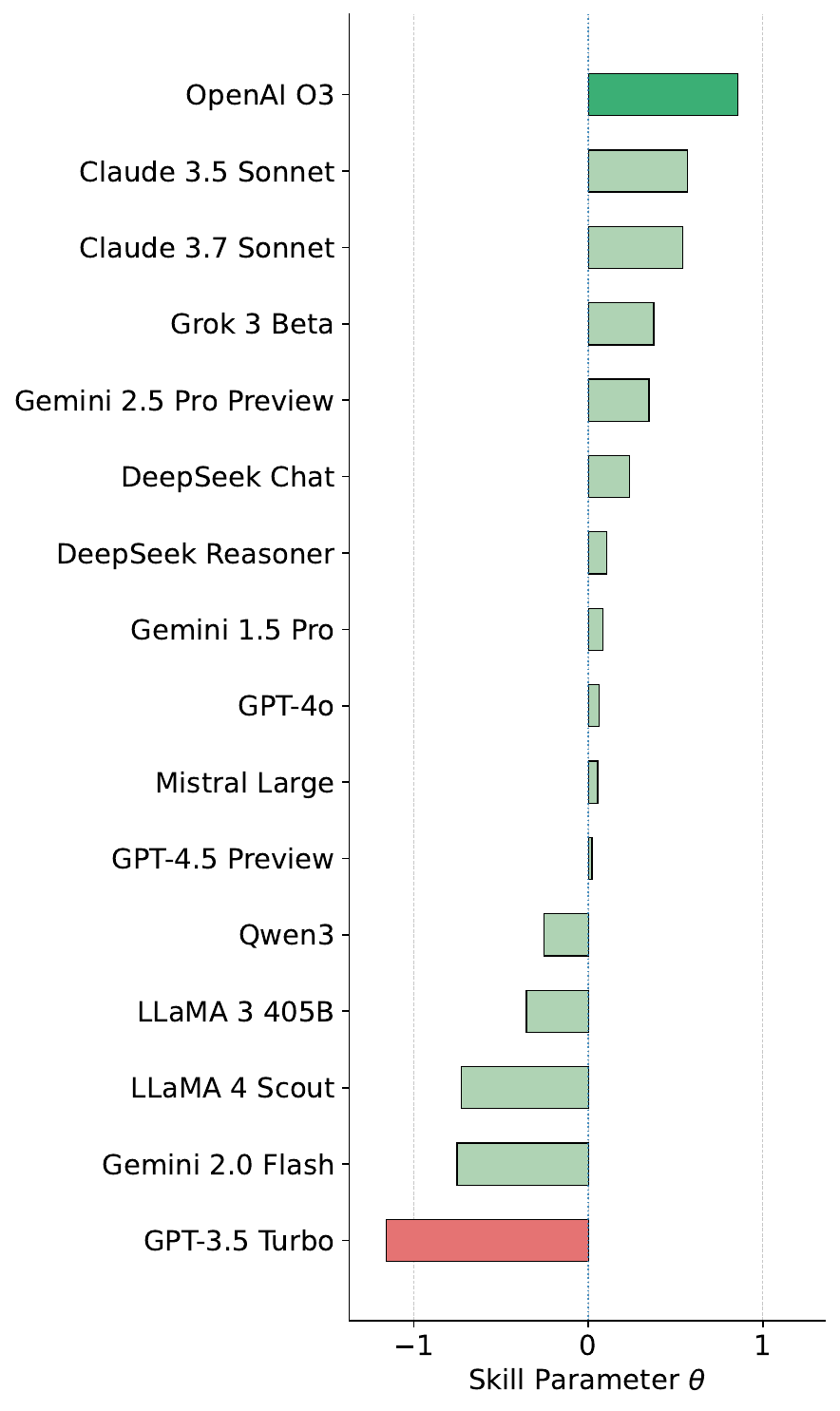}
    \caption{Claude 4 Sonnet (judge)}
  \end{subfigure}\hfill
  \begin{subfigure}[t]{0.24\textwidth}
    \centering
    \includegraphics[width=\linewidth]{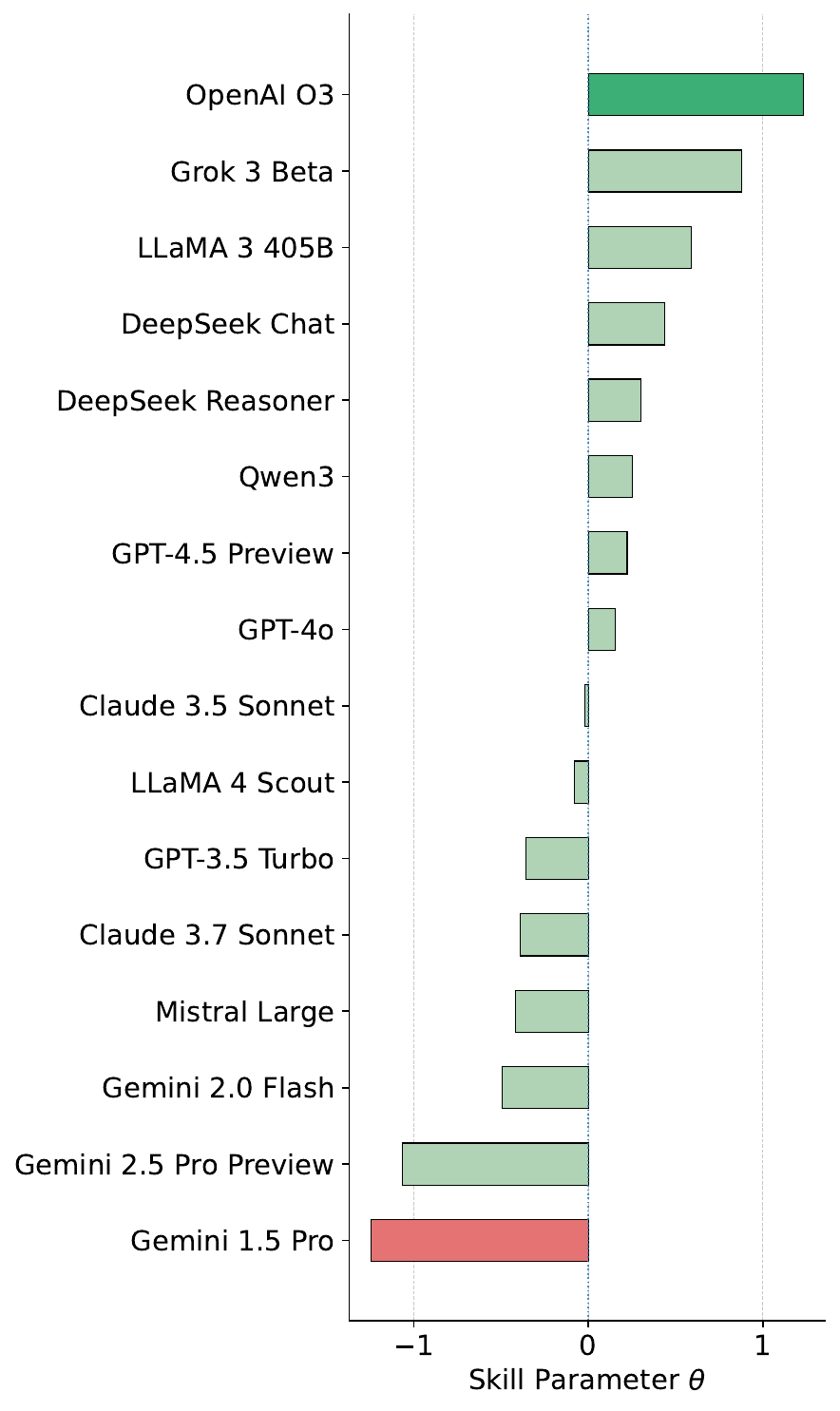}
    \caption{GPT-4o (judge)}
  \end{subfigure}\hfill
  \begin{subfigure}[t]{0.24\textwidth}
    \centering
    \includegraphics[width=\linewidth]{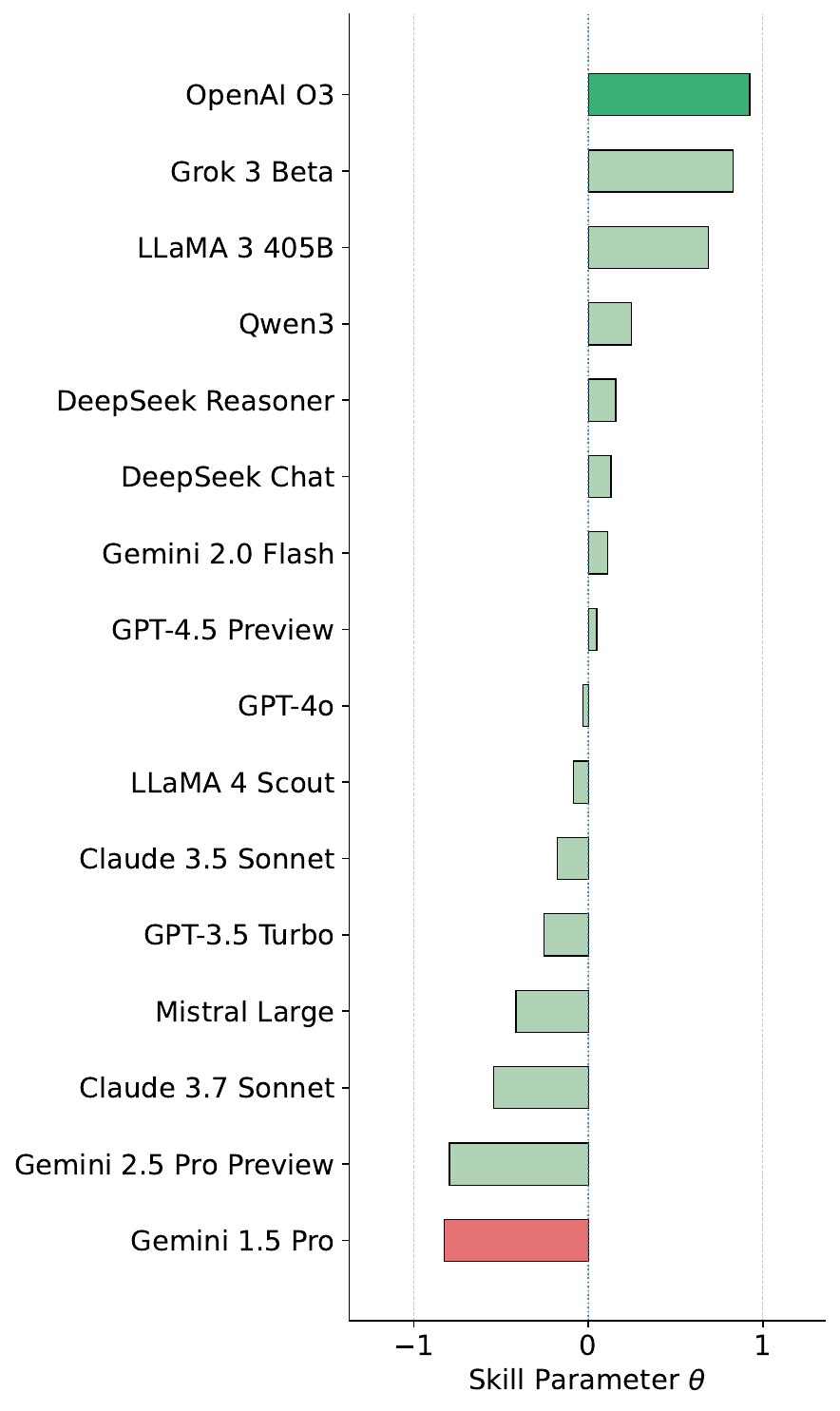}
    \caption{Gemini 2.0 Flash (judge)}
  \end{subfigure}

  \caption{Ideas task under the EQ Bench system prompt. Bradley-Terry strengths ($\theta$) from human raters and three LLM-as-judge evaluators.}
  \label{fig:eqbench_ideas_row}
\end{figure}
\begin{table}[H]
  \centering
  \caption{EQ Bench judge prompt: Spearman’s rank correlation ($\rho$) between LLM-as-judge and human model rankings by prompt type (higher indicates closer alignment with human rankings).}
   \label{tab:eqbench_spearman}
  \small
  \begin{tabular}{lccc}
    \toprule
    \textbf{Prompt} & \textbf{Claude 4 Sonnet} & \textbf{GPT-4o} & \textbf{Gemini 2.0 Flash} \\
    \midrule
    Ideas       & 0.238 & 0.353 & 0.241 \\
    Wild Ideas  & 0.629 & 0.565 & 0.300 \\
    Insight     & 0.329 & 0.521 & 0.144 \\
    \bottomrule
  \end{tabular}
\end{table}
\noindent\textit{What this judge rewards.}
A balanced mix of creativity and craft. The rubric emphasizes authenticity, originality, writing quality, coherence, instruction following, and overall atmosphere. It discourages verbosity, gratuitous metaphor, and showy vocabulary. 

\noindent\textit{Agreement with humans.}
The correlations in \Cref{tab:eqbench_spearman} indicate that alignment is task- and judge-dependent. Agreement is strongest for \emph{Wild Ideas} (up to $\rho=0.629$ for Claude 4 Sonnet; $\rho=0.565$ for GPT-4o) and weaker for \emph{Ideas} and \emph{Insight} (often $\rho<0.35$, and as low as $0.144$ for Gemini 2.0 Flash on \emph{Insight}). In practice, changing the evaluating model or rubric emphasis materially re-orders systems, so LLM-as-judge should be treated as a coarse filter rather than a substitute for expert assessment.

\noindent\textit{Ranking patterns.}
\Cref{fig:eqbench_ideas_row} shows orderings that are broadly consistent with the other judge prompts, with only modest movements where models trade originality for polish or vice versa. Panels for \emph{Insights} and \emph{Wild Ideas} under the same rubric are provided in the appendix (\Cref{fig:eqbench_insights_row,fig:eqbench_wildideas_row}).

\subsection{Implications}

\begin{itemize}

\item \textbf{LLM-as-judge systems do not reliably mirror human judgements.} Across prompts, rank alignment with human panels is weak to moderate and unstable (e.g., Spearman $\rho$ typically $\le 0.35$ and varies by judge and prompt type). Changing the judge often reorders models. These systems should not be used to select or rank models and must not substitute for expert review.
  \item \textbf{Make rubric choices explicit.} Prompts that reward clarity, brevity, and structure will favour models like O3. State the intended objective of the judge, disclose criteria, and publish per criterion scores, not only ranks.
\item \textbf{Use multiple LLM judges.} Different judge models can exhibit systematic preferences for particular generators. For example, Gemini 2.0 Flash rates LLaMA3 405B comparatively highly, and several judges often prefer O3. Try judges from diverse vendors, report cross-judge variance, and aggregate by median rank or majority vote. Calibrate periodically against human anchors to detect drift.
\end{itemize}

\section{Conventional creativity tests}
\subsection{Introduction}
Earlier parts evaluated large language models on brand-focused briefs. Here we assess general creative ability outside the marketing domain and ask whether performance on conventional creativity tests transfers to domain-specific tasks. We follow the standard view that creativity requires both novelty and appropriateness \cite{runco2012standard}, and we ground our design in the Torrance Tests of Creative Thinking (TTCT) and Guilford’s divergent-thinking framework, which emphasize fluency, flexibility, originality, and elaboration \cite{torrance1966torrance,guilford1961three}. 

In this part we adapt four open-ended task types. \textit{Task I, Creative Reuse and Improvement,} asks models to repurpose or upgrade familiar objects or situations in varied, nonredundant ways, linking primarily to flexibility and originality. \textit{Task II, Implications and Adaptations,} extends a given premise into concrete consequences or applications across contexts, targeting fluency and flexible perspective shifts. \textit{Task III, Speculative Narrative,} elicits short scenarios that introduce unexpected angles while remaining coherent and intelligible, stressing originality with sufficient elaboration to maintain sense. \textit{Task IV, Practical Innovation,} requests implementable concepts that solve a stated problem, balancing novelty with clarity and usefulness. These tasks probe core divergent-thinking capacities that are not tied to a brand brief, allowing us to compare model rankings on conventional tests with the marketing results reported earlier.

A central question is transfer. Creativity shows meaningful domain and task specificity, so strong scores on general divergent-thinking measures may not fully predict brand-constrained performance \cite{baer2015domain}. By evaluating the same set of models on both types of tasks, we quantify alignment and divergence, examine where rubric choice drives differences, and identify models whose strengths on conventional tests coincide with human preference in marketing contexts.

\subsection{Methods}

\subsubsection{Benchmark design}
We adapted open-ended tasks from the Torrance Tests of Creative Thinking (TTCT) and related divergent-thinking assessments for large language model (LLM) evaluation, grounding the design in established dimensions of fluency, flexibility, originality, and elaboration \cite{torrance1966torrance,guilford1961three,runco2012standard}. All prompts are listed in \Cref{app:creativity_tasks}.

\subsubsection{Task types and example prompts}
\paragraph{Task I: Creative Reuse and Improvement.}
Assesses applied divergent thinking on everyday objects. The goal is to produce many distinct, category-spanning, and novel ideas as a numbered list.\\
\textit{Examples:} “List as many alternative uses for a bicycle inner tube as you can.”, “Suggest as many improvements as you can to make public libraries more engaging and relevant in the digital age.”

\paragraph{Task II: Implications and Adaptations.}
Explores hypothetical premises and cross-context adaptations, emphasizing perspective shifts and concrete consequences.\\
\textit{Examples:} “What if humans had a ‘patience meter’ visible above their heads, how would daily interactions change?”, “If rain began falling upward, how would cities redesign streets and buildings?”

\paragraph{Task III: Speculative Narrative.}
Elicits short fiction and world-building that combine novelty with coherent detail.\\
\textit{Examples:} “Suppose you found a remote control that could pause, rewind, and fast-forward your own life. Write about a day you use it.”, “Write a story based on the prompt ‘The city that dreamed’.”

\paragraph{Task IV: Practical Innovation.}
Targets inventive but actionable solutions to stated constraints, balancing novelty with usefulness.\\
\textit{Examples:} “Design a system to efficiently and fairly distribute food in a large, isolated community after a natural disaster.”, “Propose a low-cost approach to reduce noise pollution in a dense urban apartment building.”

\subsubsection{Creativity dimensions}
We scored each response on four dimensions commonly used in creativity research:
\begin{itemize}
  \item \textbf{Fluency:} number of distinct ideas. Repetitions or paraphrases count once.
  \item \textbf{Flexibility:} range of categories or perspectives, evidence of shifting frames or approaches.
  \item \textbf{Originality:} novelty or unexpectedness relative to common responses.
  \item \textbf{Elaboration:} depth, specificity, and development, including concrete details or plausible mechanisms.
\end{itemize}

\subsubsection{Task-dimension mapping}
Each task targeted a specific subset of creativity dimensions, as summarised in \Cref{tab:task_dim_map}. Task I was scored on fluency, flexibility, and originality. Task II assessed flexibility, originality, and elaboration. Task III emphasised originality and elaboration. Task IV focused on originality under practical constraints, without requiring high fluency or elaboration.
\begin{table}[H]
  \centering
  \caption{Creativity dimensions targeted by each task.}
  \label{tab:task_dim_map}
  \begin{tabular}{lcccc}
    \toprule
    \textbf{Task} & \textbf{Fluency} & \textbf{Flexibility} & \textbf{Originality} & \textbf{Elaboration} \\
    \midrule
    Task I: Creative Reuse and Improvement   & Yes & Yes & Yes & No  \\
    Task II: Implications and Adaptations    & No  & Yes & Yes & Yes \\
    Task III: Speculative Narrative          & No  & No  & Yes & Yes \\
    Task IV: Practical Innovation            & No  & No  & Yes & No  \\
    \bottomrule
  \end{tabular}
\end{table}

\subsubsection{LLM-as-judge rubric}
LLM judges rated each response on the four dimensions above using a 1 to 5 scale per dimension. The rubric text was presented alongside the task prompt and model response. When uncertain, judges were instructed to assign the lower of two candidate ratings. We adapted judging prompts from prior literature on rubric-guided LLM evaluation, full wording appears in the appendix.

\subsubsection{Judging LLMs}
We used three independent judges to reduce shared biases and to estimate cross judge reliability. Across judges, the rank ordering of models was broadly similar for all four tasks. For brevity, we present results from the OpenAI judge (GPT-4o) in the main text. Full results for Claude 4 Sonnet and Gemini 2.0 Flash are provided in the \Cref{app:part_d_claude_judge_results} and \Cref{app:part_d_gemini_judge} .

\begin{table}[H]
  \centering
  \caption{LLM judges used for conventional creativity tests.}
  \label{tab:judge_models_partD}
  \begin{tabular}{lll}
    \toprule
    \textbf{Judge name} & \textbf{API model ID}        & \textbf{Provider} \\
    \midrule
    GPT-4o              & gpt-4o-2024-11-20            & OpenAI \\
    Claude 4 Sonnet     & claude-sonnet-4-20250514     & Anthropic \\
    Gemini 2.0 Flash    & models/gemini-2.0-flash-001  & Google \\
    \bottomrule
  \end{tabular}
\end{table}

\subsection{Results}

\subsubsection{Task I: Creative Reuse and Improvement}
\begin{figure}[H]
  \centering
   \begin{subfigure}[t]{0.24\textwidth}
    \centering
    \includegraphics[width=\linewidth]{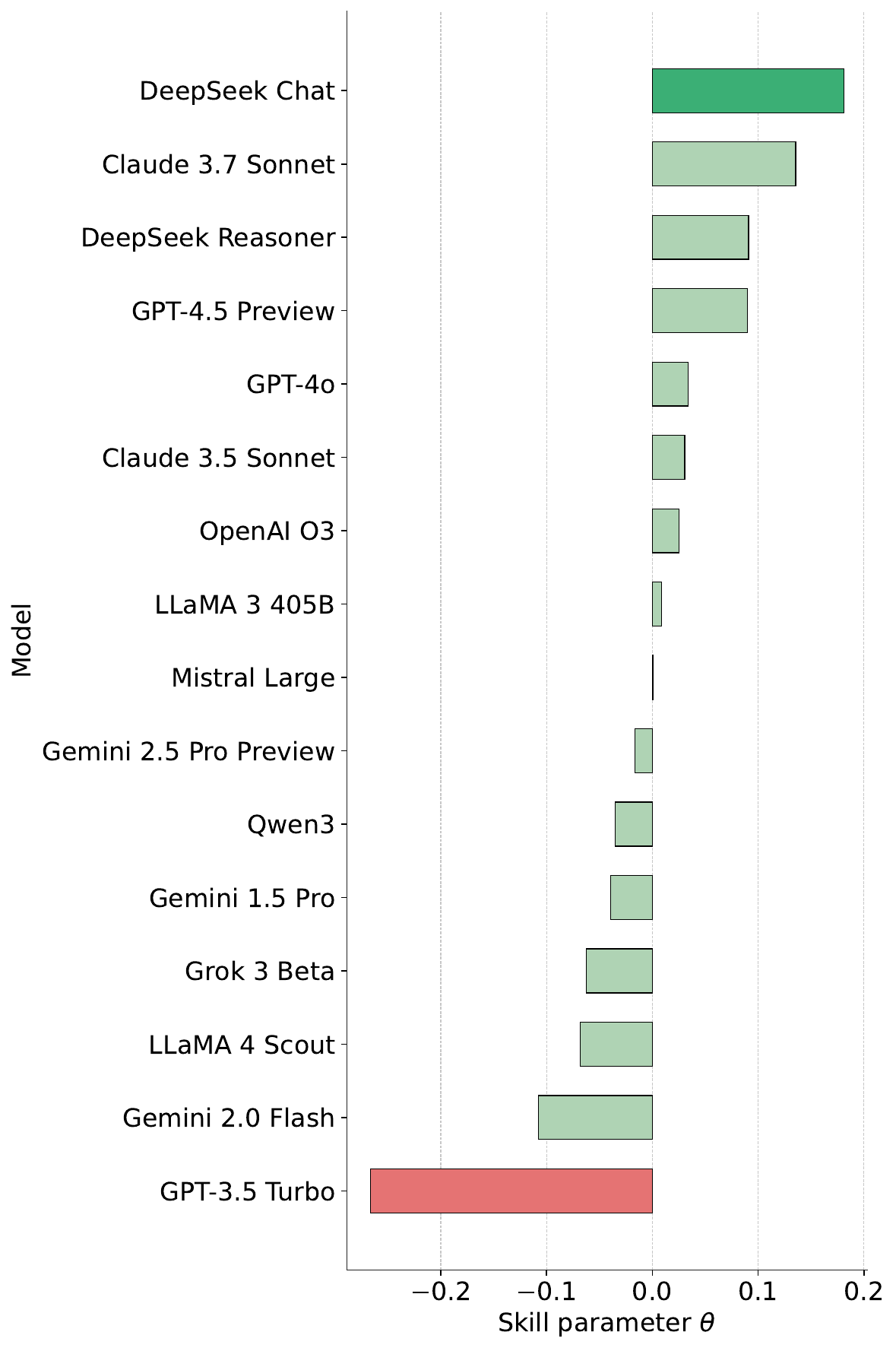}
    \caption{Human ratings}
  \end{subfigure}
  \begin{subfigure}[t]{0.24\textwidth}
    \centering
    \includegraphics[width=\linewidth]{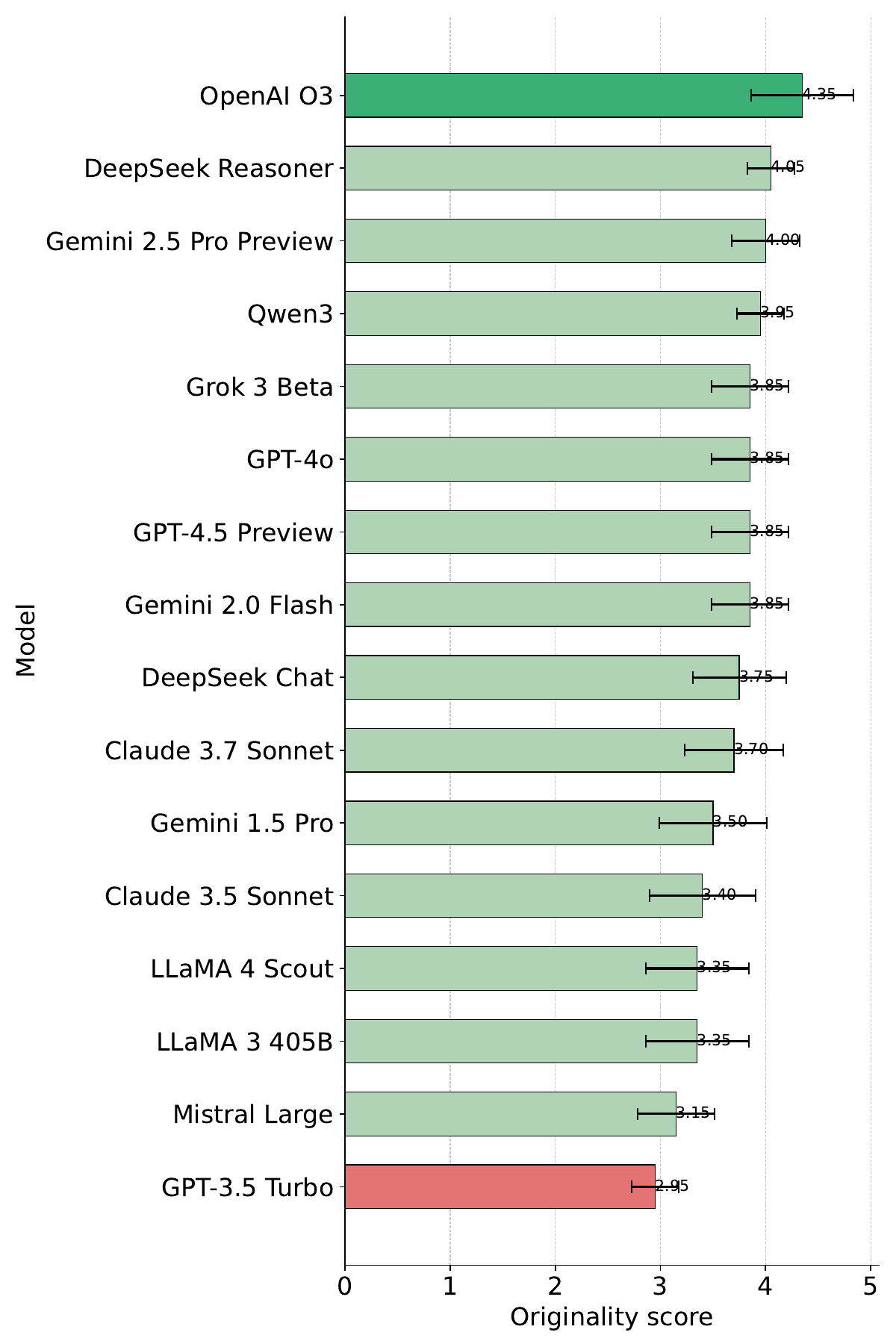}
    \caption{Originality}
    \label{fig:task1_openai_originality}
  \end{subfigure}\hfill
  \caption{Task I (Creative Reuse and Improvement). Comparison of originality scores and Human rating Bradley-Terry skill parameters across models. Spearman $\rho = 0.328$ between Originality and Human Model ranking from Part A}
  \label{fig:task_1_orig_bt_comparison}
\end{figure}

As shown in \Cref{tab:task_1_orig_bt_comparison}, Fluency and Flexibility are at or near ceiling (means $\approx 5.0$ with near-zero SDs), so they contribute little discrimination on this task. The informative signal comes from Originality, where means span $2.95$ to $4.35$ with small SDs ($\approx 0.22$-$0.50$). OpenAI O3 leads on Originality at $4.35\pm0.49$, outperforming the next highest mean (DeepSeek Reasoner at $4.05\pm0.22$) by about $0.30$. Reasoning-oriented models such as DeepSeek Reasoner and Qwen3 also rank relatively high, while Mistral Large and GPT-3.5 Turbo define the lower bound. Elaboration shows some spread but is partly constrained by brevity requirements, so it is less diagnostic here. The correlation between human Bradley-Terry skill parameters and originality means for Task~A is modest and non-significant (Spearman $\rho = 0.328$). This indicates that performance on Creative Reuse and Improvement originality ratings does not strongly align with human preference in marketing and advertising-related creativity tasks.

\subsubsection{Task II: Implications and Adaptations}
\begin{figure}[H]
  \centering
  \begin{subfigure}[t]{0.24\textwidth}
    \centering
    \includegraphics[width=\linewidth]{figures_part_d_bradley_terry_score_best_worst.pdf}
    \caption{Human ratings}
  \end{subfigure}\hfill
  \begin{subfigure}[t]{0.24\textwidth}
    \centering
    \includegraphics[width=\linewidth]{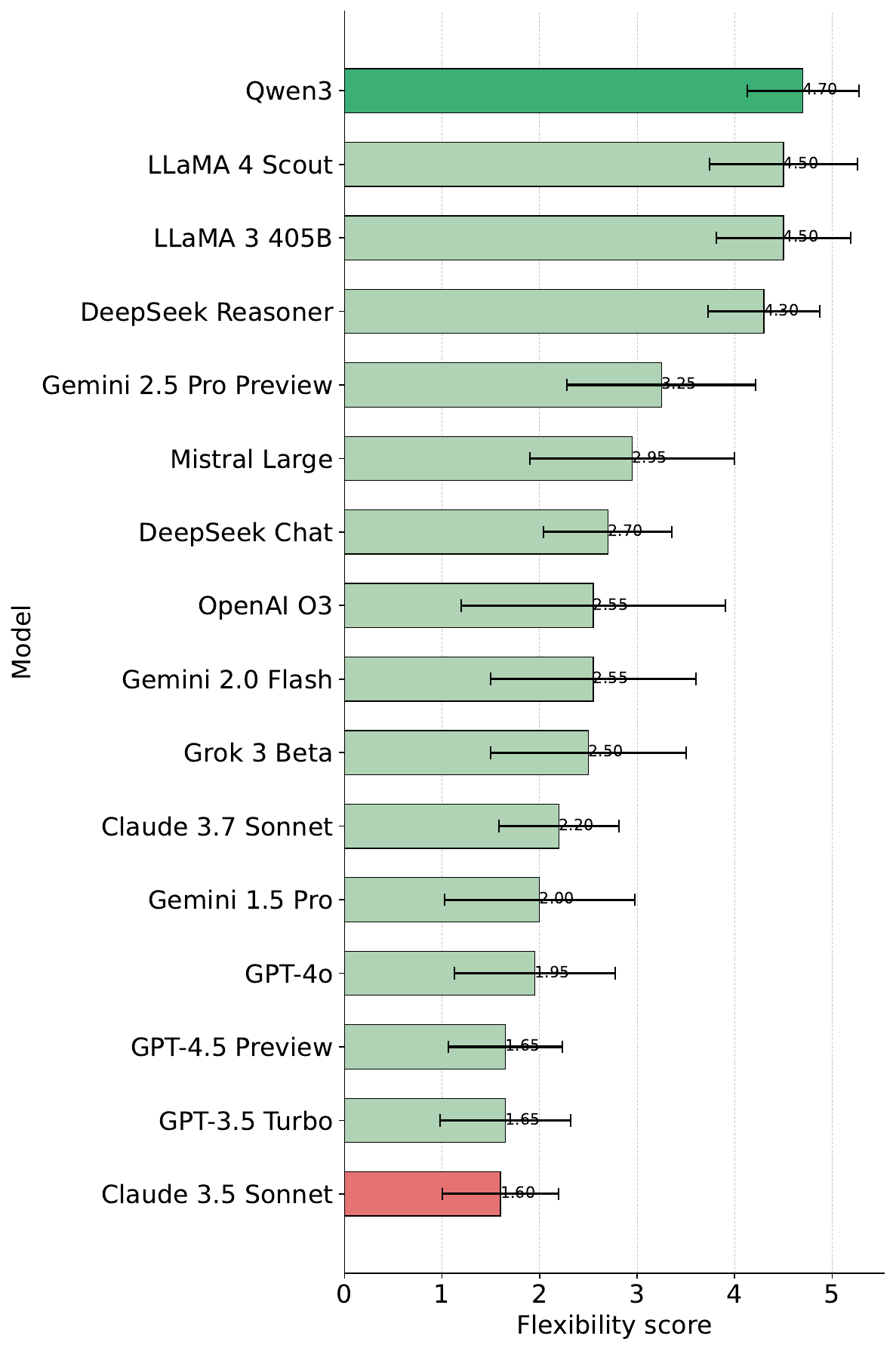}
    \caption{Flexibility}
  \end{subfigure}\hfill
  \begin{subfigure}[t]{0.24\textwidth}
    \centering
    \includegraphics[width=\linewidth]{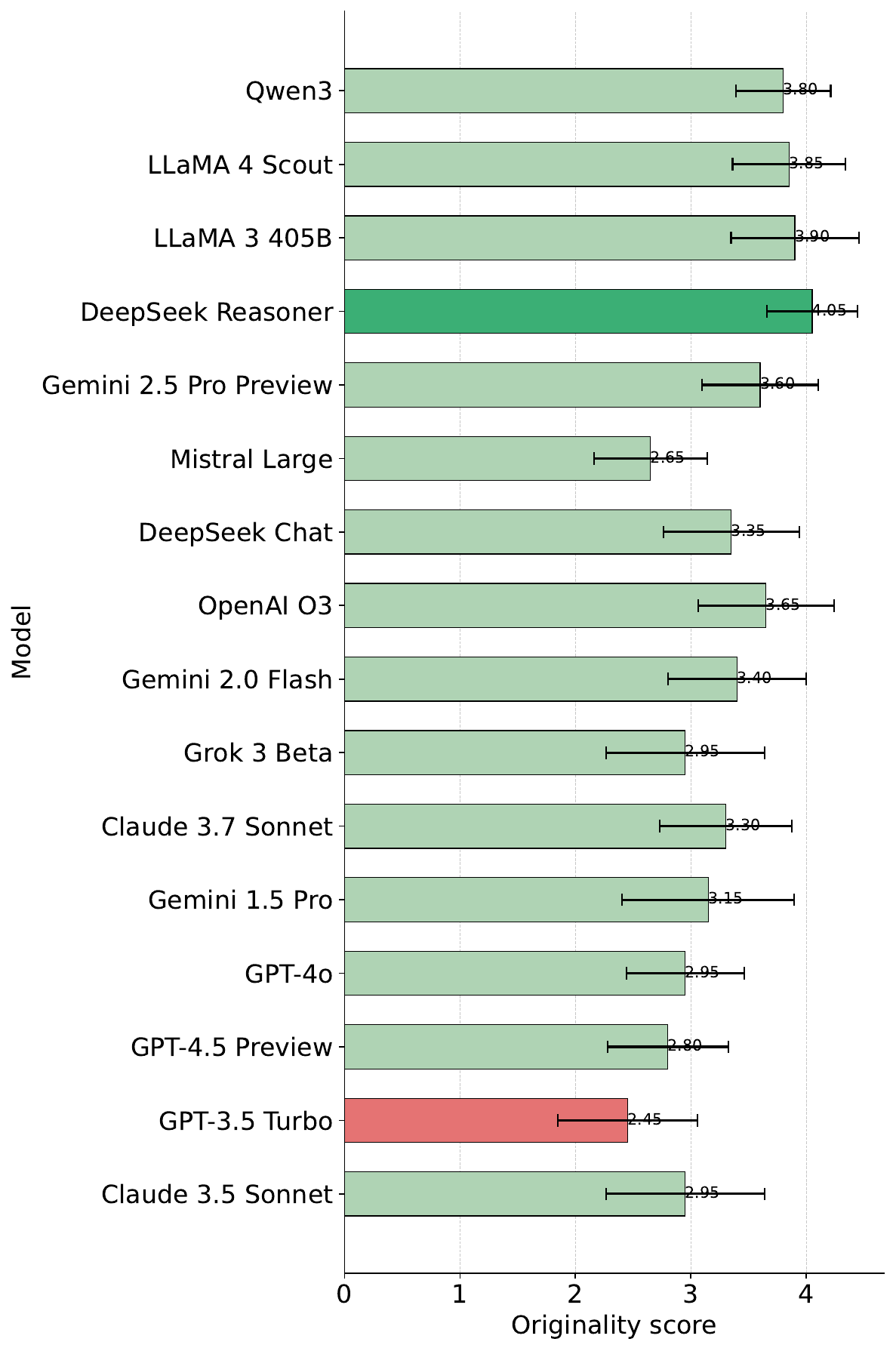}
    \caption{Originality}
  \end{subfigure}\hfill
  \begin{subfigure}[t]{0.24\textwidth}
    \centering
    \includegraphics[width=\linewidth]{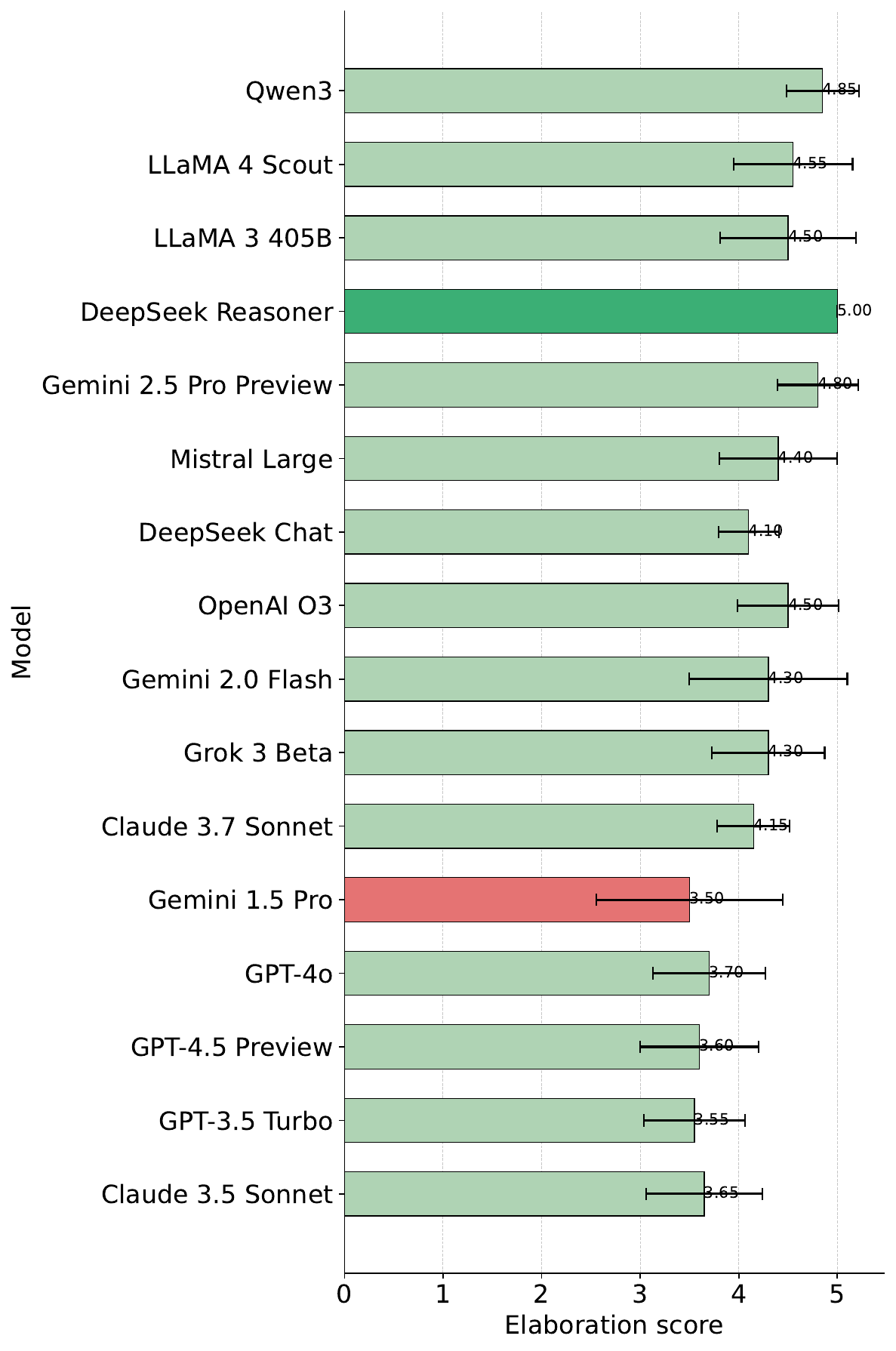}
    \caption{Elaboration}
  \end{subfigure}
  \caption{Task II (Implications and Adaptations). Comparison of human ratings (Bradley-Terry) and OpenAI-judge dimension scores (mean $\pm$ SD) by model.}
  \label{fig:task2_openai_human_compare}
\end{figure}

Unlike Task I, Task II separates models on \emph{flexibility} and \emph{elaboration} in addition to \emph{originality}. As visible in \Cref{fig:task2_openai_human_compare} (full table in the appendix, \Cref{tab:task2_openai_scores}), DeepSeek Reasoner shows the strongest all-round profile with flexibility $4.30\pm0.57$, originality $4.05\pm0.39$, and elaboration $5.00\pm0.00$, indicating broad fra  -ming with fully developed consequences. Qwen3 leads flexibility at $4.70\pm0.57$ and also exhibits high elaboration ($4.85\pm0.37$). The LLaMA 3 and LLaMA 4 variants present balanced breadth and depth across dimensions. OpenAI O3 provides consistently detailed analyses (elaboration $4.50\pm0.51$) but lower and more variable flexibility (SD $\approx 1.36$), consistent with sensitivity to prompt framing. At the lower end, GPT-3.5 Turbo and Claude 3.5 Sonnet show constrained flexibility and reduced originality. Flexibility variances are larger here than in Task I, suggesting that scenario framing steers the range of perspectives considered. By contrast, top models show small variances in elaboration, indicating dependable depth once a perspective is selected. Correlations with human Bradley-Terry skill parameters are negligible across all three dimensions: flexibility (Spearman $\rho = -0.077$), originality ($\rho = 0.125$), and elaboration ($\rho = 0.002$). These results reinforce the finding from Task I, performance on standard creativity test dimensions does not align with human preferences in marketing and advertising creativity. Human evaluators appear to prioritize qualities beyond flexibility, originality, and elaboration as scored in isolated tasks.

\subsubsection{Task III: Speculative Narrative}

\begin{figure}[H]
  \centering
  \begin{subfigure}[t]{0.24\textwidth}
    \centering
    \includegraphics[width=\linewidth]{figures_part_d_bradley_terry_score_best_worst.pdf}
    \caption{Human ratings}
  \end{subfigure}
  \begin{subfigure}[t]{0.24\textwidth}
    \centering
    \includegraphics[width=\linewidth]{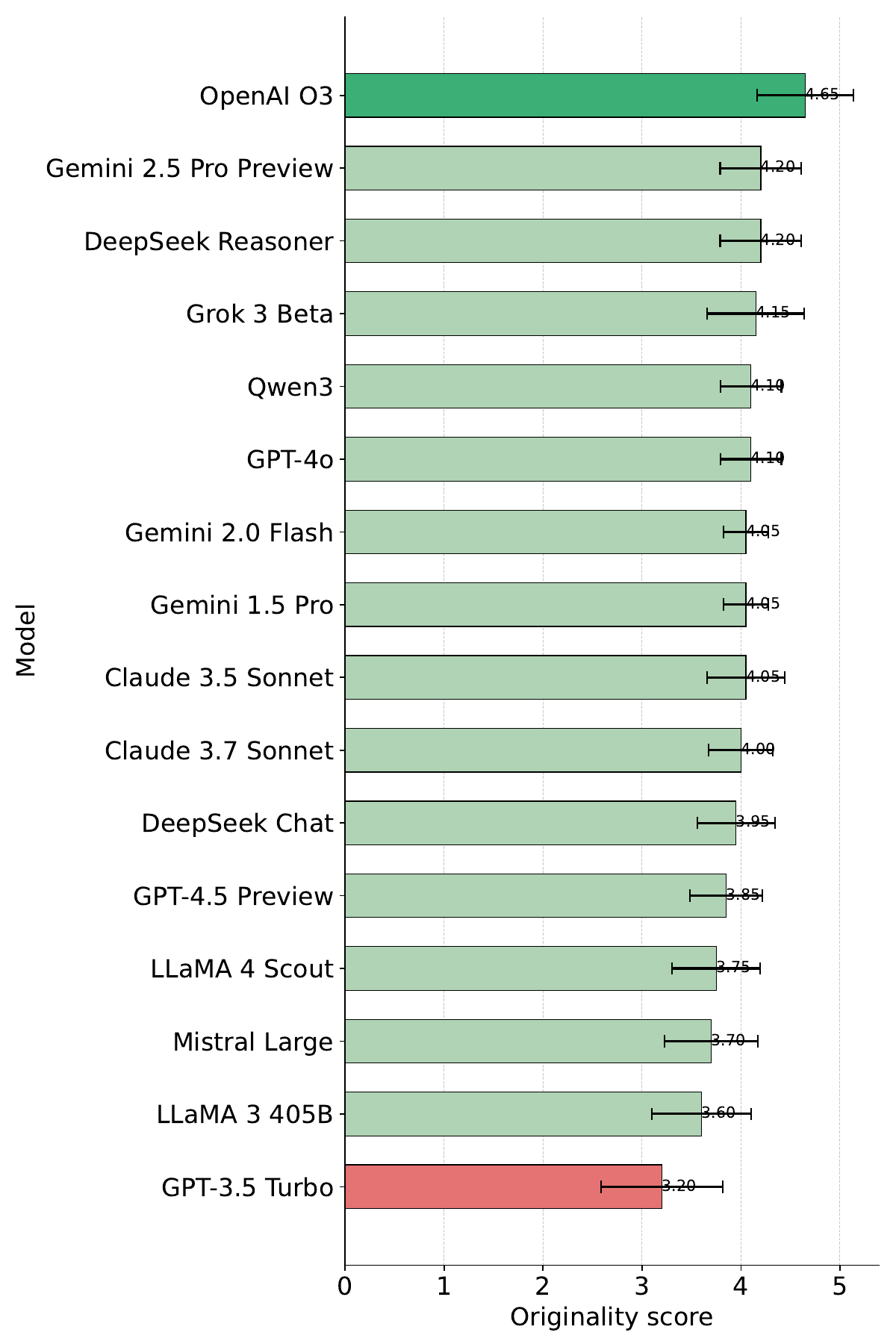}
    \caption{Originality}
  \end{subfigure}
  \begin{subfigure}[t]{0.24\textwidth}
    \centering
    \includegraphics[width=\linewidth]{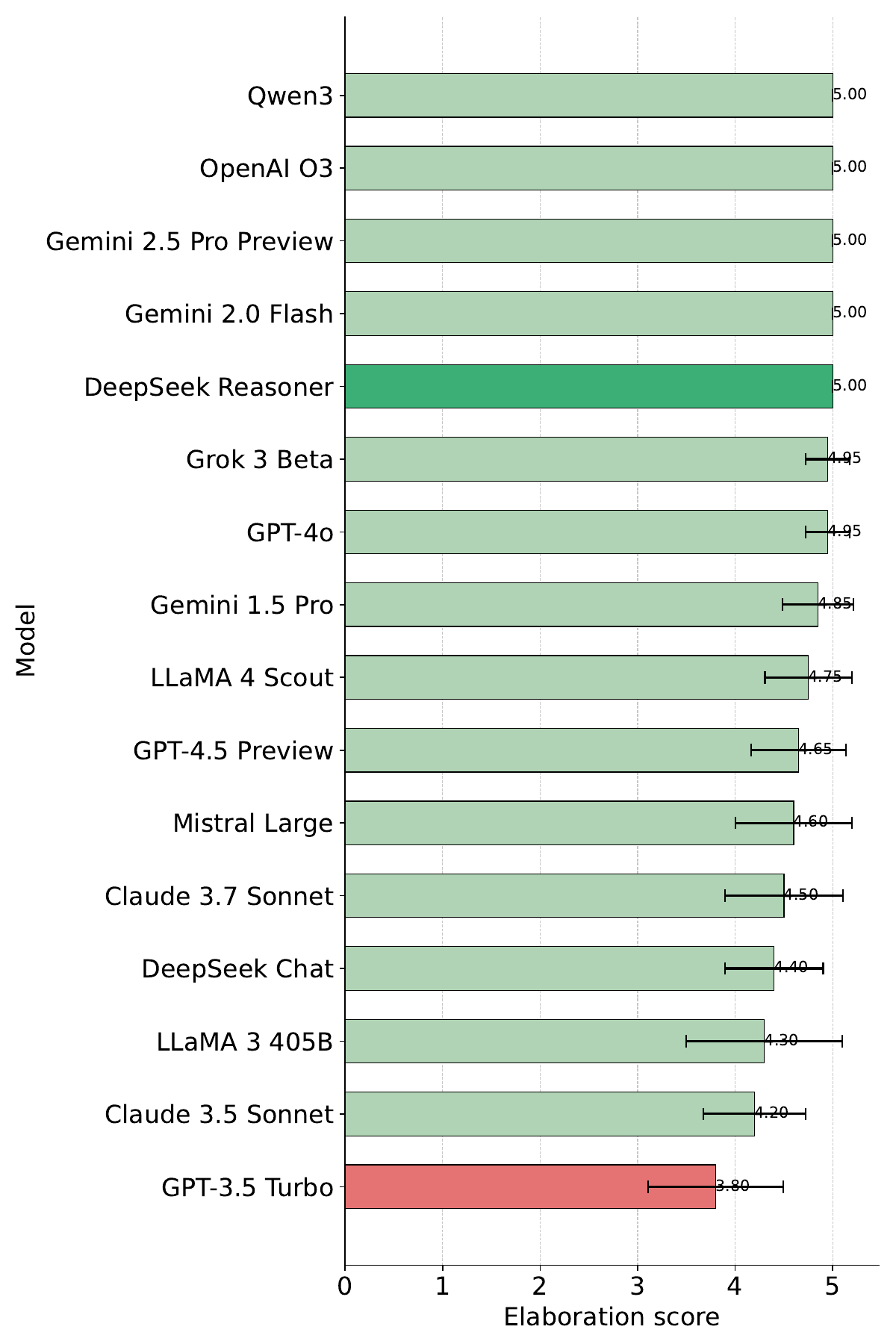}
    \caption{Elaboration}
  \end{subfigure}\hfill
  \caption{Task III (Speculative Narrative). Comparison of human ratings (Bradley-Terry) and OpenAI-judge dimension scores (mean $\pm$ SD) by model.}
  \label{fig:task3_openai_human_panels}
\end{figure}

This task rewards novelty together with sustained scene development. As shown in \Cref{fig:task3_openai_human_panels} (see also \Cref{tab:task3_openai_scores} in the appendix), elaboration is near ceiling for many models (several at $5.00$ with very small variance) indicating that once a narrative direction is chosen, most LLMs maintain coherent, detailed prose. Differentiation therefore arises primarily from originality. OpenAI O3 attains the highest originality ($\approx 4.65$). A strong tier near $4.20$ includes Gemini 2.5 Pro Preview and DeepSeek Reasoner. GPT-4o, Grok 3 Beta, Qwen3, Gemini 1.5 Pro, and Gemini 2.0 Flash cluster around $4.0$ with high elaboration, making them reliable storytellers requiring limited editing. LLaMA 3 405B, LLaMA 4 Scout, and Mistral Large sit somewhat lower on originality while maintaining substantial elaboration, and GPT-3.5 Turbo marks the lower bound on both dimensions. Correlations with human Bradley-Terry skill parameters remain weak: originality (Spearman $\rho = 0.199$) and elaboration ($\rho = -0.099$).

\subsubsection{Task IV: Practical Innovation}

\begin{figure}[H]
  \centering
  \begin{subfigure}[t]{0.24\textwidth}
    \centering
    \includegraphics[width=\linewidth]{figures_part_d_bradley_terry_score_best_worst.pdf}
    \caption{Human ratings}
  \end{subfigure}
  \begin{subfigure}[t]{0.24\textwidth}
    \centering
    \includegraphics[width=\linewidth]{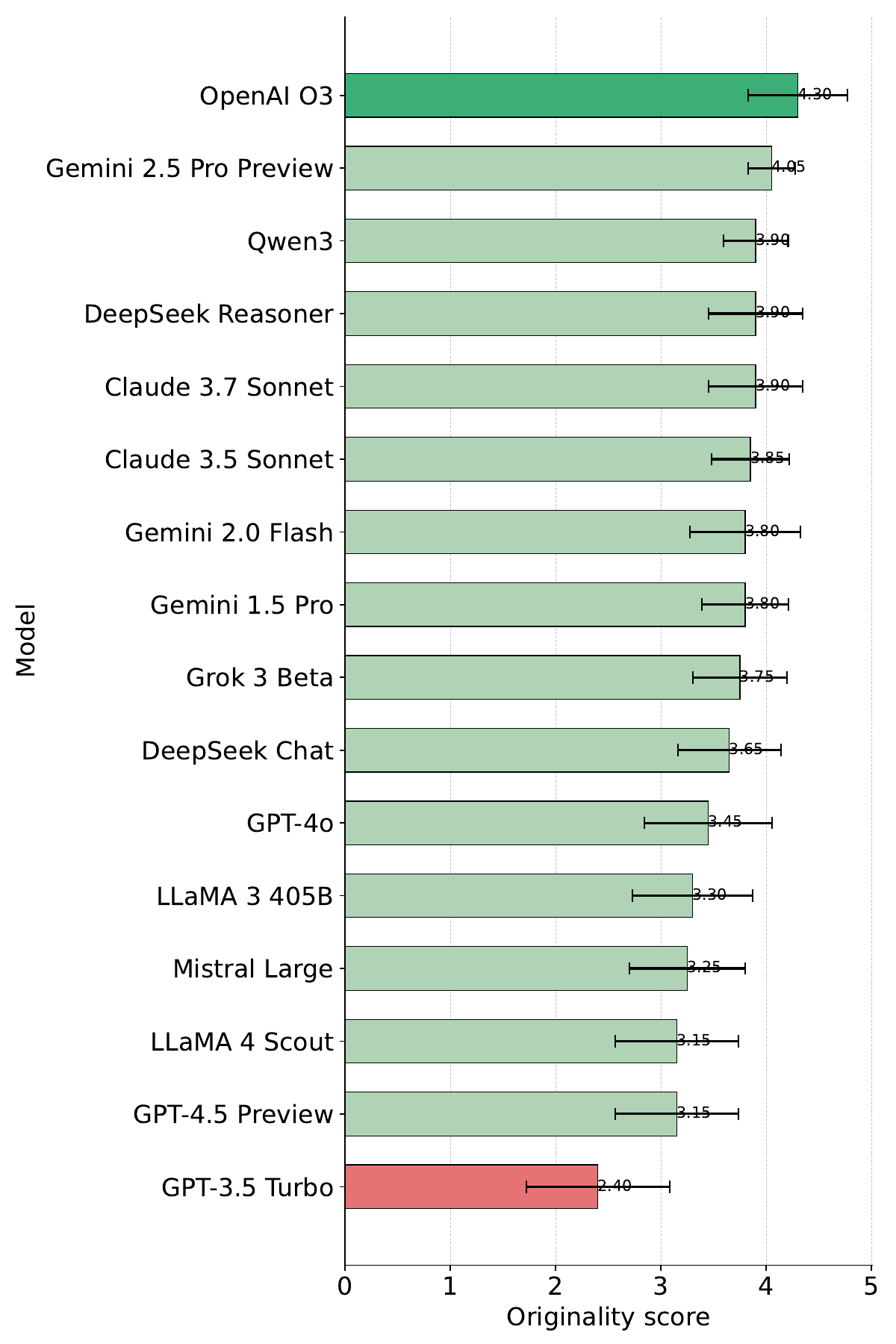}
    \caption{Originality}
  \end{subfigure}
  \caption{Task IV (Practical Innovation). Comparison of human ratings (Bradley-Terry) and OpenAI-judge originality scores (mean $\pm$ SD) by model.}
  \label{fig:task4_openai_originality}
\end{figure}

As shown in \Cref{fig:task4_openai_originality} (see \Cref{tab:task4_openai_human_originality} in the appendix), OpenAI O3 attains the highest originality score ($4.30\pm0.47$), indicating frequent novel solutions within practical constraints. Gemini 2.5 Pro Preview follows at $4.05\pm0.22$ with notably low variance, suggesting consistent inventive output. Qwen3 and DeepSeek Reasoner form the next tier near $3.9$, and both Claude Sonnet variants cluster between $3.85$-$3.90$. A mid group spans $3.25$-$3.80$ (Gemini 1.5 Pro, Gemini 2.0 Flash, Grok 3 Beta, DeepSeek Chat, GPT-4o, Mistral Large, LLaMA 3 405B). GPT-4.5 Preview and LLaMA 4 Scout sit lower around $3.15$, and GPT-3.5 Turbo marks the floor at $2.40\pm0.68$ with the largest spread. Correlation with human Bradley-Terry parameters is again weak (originality: Spearman $\rho = 0.323$), reinforcing the limited alignment between conventional creativity metrics and human preference

\subsection{Implications}

\paragraph{Cross-task patterns.}
Most models generate many ideas and coherent prose. Differences arise in breadth and originality. Task I hits ceilings on Fluency and Flexibility, and Task III nears a ceiling on Elaboration, so Originality carries the signal. Task II separates models on all three axes.

\paragraph{Model profiles.}
OpenAI O3 attains the highest originality, regularly proposing less common solution paths (with moderate variance). DeepSeek Reasoner and Qwen3 shift perspectives widely in Task II and stay competitive in narrative and innovation work. Gemini 2.5 Pro Preview shows slightly lower peak originality but low variance and strong elaboration, which suits dependable content.

\paragraph{Divergence from human ratings.}
Standard creativity metrics show weak or inconsistent correlations with Part A human preference judgments. This suggests that success on abstract dimensions (e.g., originality or elaboration in standard tasks) does not directly translate into perceived value in marketing and advertising creativity. Likely drivers of the gap include differences in construct validity (what “creativity” means in psychometric tests vs.  brand work), and stylistic preferences that influence human judgments. For applied evaluation, domain-specific human ratings remain important, with LLM-based scoring best positioned as a complementary diagnostic tool rather than a standalone substitute.

\section{Limitations}
\label{sec:limitations}

This study has limitations that affect how the results should be interpreted.

\begin{enumerate}[label=\textbf{L\arabic*.}, leftmargin=*]
  \item \textbf{Scope and language.} Prompts and ratings are English-first. Applicability to other languages, markets, and culturally specific briefs is untested.
  \item \textbf{Brand and task coverage.} Although the benchmark covers 100 brands across 12 categories and three prompt types, it does not exhaust the space of real briefs. Results may differ for longer formats, multimodal artifacts, channel-specific constraints, or production workflows.

  \item \textbf{Decoding settings and model drift.} All generations used a fixed sampling temperature of $1.0$ with provider default parameters. Different temperatures, system prompts, or sampling schemes can change rankings. Models and APIs update frequently, so results are a snapshot in time.

\end{enumerate}

\section{Conclusion}

This study introduces a domain-grounded benchmark for marketing creativity and applies it to a wide set of contemporary LLMs. Three findings stand out:

\begin{itemize}[leftmargin=*]
  \item \textbf{No clear winner.} In head-to-head comparisons, win rates between models are small, closer to coin flips than decisive victories. Even the strongest systems only modestly outperform the weakest on average, with win rates topping out around 61\% in head-to-head matchups.
  \item \textbf{LLMs are not good judges.} When asked to rank creative work, automated judges display stable, judge-specific preferences that do not reliably imitate human evaluations, and they often express unwarranted confidence. This limits the validity of “agentic” set-ups that rely on model-based adjudication without human oversight.
  \item \textbf{Variance matters.} Because creative preference is subjective and judges do not track experts well, optimising for variation is more useful than chasing small average quality gaps. Keeping humans in the loop is essential for critical decisions. In our data, Claude~3.7 and Gemini~2.5~Pro provide especially strong variation, which makes them effective for idea exploration.
\end{itemize}

\noindent For practitioners, four recommendations follow:

\begin{enumerate}[label=\textbf{(\arabic*)}, leftmargin=*]
  \item \textbf{Choose for fit, not rank.} With small gaps between models, prioritise brand-voice control, workflow integration, latency, cost, and team preference rather than leaderboard position.
  \item \textbf{Value variation.} Use LLMs to expand the pool of ideas, then rely on human judgment for selection. Claude~3.7 and Gemini~2.5~Pro are strong options for producing diverse alternatives.
  \item \textbf{Be wary of overconfidence.} Models can accelerate evaluation pipelines, but they should not replace human review when deciding what will resonate.
  \item \textbf{Keep experimenting with prompts.} Small wording shifts can unlock additional range; for example, reframing \emph{Ideas} as \emph{Wild Ideas} reliably pushes outputs into less conventional territory.
\end{enumerate}

\noindent Overall, the evidence supports a pragmatic stance: use LLMs as accelerators of idea exploration, not arbiters of creative value. Combine models where helpful, and anchor final decisions in targeted human evaluation.

\section*{Acknowledgements}

We thank our Industry Advisors for their guidance and advocacy: 
\textbf{Zoe Scaman} (Founder, Bodacious), 
\textbf{James Hurman} (Founding Partner, Previously Unavailable), 
\textbf{Tom Roach} (VP Brand Strategy, Jellyfish/Brandtech), 
\textbf{Laurence Green} (Director of Effectiveness, IPA), 
\textbf{Tony Hale} (CEO, Advertising Council Australia), 
\textbf{Gavin McLeod} (Chief Creative Officer, CHEP Network), 
\textbf{Jeremy Lockhorn} (SVP Creative Technologies \& Innovation, 4A's), 
\textbf{Dagmara Szulce} (Managing Director, IAA Global Office), 
\textbf{Kevin Swanepoel} (CEO, The One Club for Creativity), 
and \textbf{Paul Drake} (Foundation Director, D\&AD).

We acknowledge and appreciate the support of leading advertising and marketing organisations around the world: 
\textbf{American Association of Advertising Agencies (4A's)}, 
\textbf{Advertising Council Australia (ACA)}, 
\textbf{APG}, 
\textbf{D\&AD}, 
\textbf{Institute of Practitioners in Advertising (IPA)}, 
\textbf{International Advertising Association (IAA)}, 
and \textbf{The One Club for Creativity}.

We are especially grateful to \textbf{Maddie Gross}, \textbf{Thomas Taylor}, \textbf{Ronan Murphy}, and \textbf{Eden Payne} for their invaluable contributions and support throughout the project.

Finally, we thank the many strategists, creatives, and marketers who contributed judgments on the platform. Your participation makes this collective, transparent benchmark possible.

\printbibliography

\renewcommand\theequation{\Alph{section}\arabic{equation}} %
\counterwithin*{equation}{section} %
\renewcommand\thefigure{\Alph{section}\arabic{figure}} %
\counterwithin*{figure}{section} %
\renewcommand\thetable{\Alph{section}\arabic{table}} %
\counterwithin*{table}{section} %

\begin{appendices}
\appendix

\section{Benchmark Setup}
\subsection{Prompt Templates} \label{app:prompt}

\subsubsection{System Prompt (used for everything)}  
\begin{promptbox}{System Prompt}
You are a world-class brand strategist and creative thinker at a top global agency.
You generate original insights and bold campaign ideas that spark creativity and cultural relevance. Your thinking is strategic, surprising, and never cliché.
You draw from human behaviour, cultural shifts, product truths, audience quirks, and category conventions — wherever the best ideas live.
Your insights are revealing. Your ideas are platformable. Your wild ideas are provocative but strategically grounded. You don’t write slogans — you ignite campaigns.
Please follow these formatting rules in your responses: \newline
- Use plain text only — no lists, markdown, emojis, or formatting \newline
- Respond in a single, concise sentence \newline
- NO PREAMBLES — do not introduce your answer or explain anything \newline
- Do not explain or justify the idea — just give the final output \newline
- Capitalize appropriately and end with punctuation\newline

\end{promptbox}

\subsubsection{User Prompt}
\begin{promptbox}{INSIGHT}
What is a surprising insight about people, culture, category, or product that [BRAND] could build a campaign around? \newline
Keep it under 10 words. Make it a creative springboard, something a strategist would share to spark ideas. Avoid slogans or generic observations.
\end{promptbox}

\begin{promptbox}{IDEAS}
Propose a big, campaignable platform idea for [BRAND]. It should be based on a strategic or cultural truth and work across any channel. \newline
Keep it under 50 words. Do not write a slogan. Make it feel like the beginning of a powerful, elastic campaign, not the end.
\end{promptbox}

\begin{promptbox}{WILD IDEAS}
What is your wildest unconventional campaign idea for [BRAND], something no traditional agency would dare present, but that could hijack culture, spark headlines, and get people talking.\newline
Keep it under 50 words. Make it strange, fresh, or provocative but still creatively smart and on-brand. Avoid generic stunts or randomness. Surprise me in a good way.
\end{promptbox}
\bigskip

\subsection{Brands} \label{app:brands}

\begin{table}[H]
\centering
\footnotesize
\setlength{\tabcolsep}{3pt}
\begin{tabularx}{\textwidth}{@{}YYYYYY@{}}
\toprule
\textbf{Fashion, Apparel \& Footwear} &
\textbf{Beauty, Wellness \& Personal Care} &
\textbf{Food, Beverage \& QSR} &
\textbf{Retail \& Marketplaces} &
\textbf{Luxury \& Prestige} &
\textbf{Tech \& Consumer Electronics} \\
\midrule
Nike       & Dove          & McDonald's & Amazon & Chanel        & Apple \\
Adidas     & Glossier      & Coca-Cola  & IKEA   & Rolex         & Samsung \\
Uniqlo     & Fenty Beauty  & Pepsi      & Walmart& Louis Vuitton & Sony \\
Patagonia  & Olay          & Burger King& Target & Tiffany \& Co.& Google \\
Gucci      & Aesop         & Oatly      & ASOS   & Prada         & Microsoft \\
Dior       & The Ordinary  & Red Bull   & Etsy   & Moncler       & Lenovo \\
Lululemon  & Neutrogena    & Doritos    & Farfetch& Cartier      & Bose \\
Allbirds   & Athletic Greens& Starbucks & H\&M   & Bottega Veneta& Beats by Dre \\
Zara       &               & Chipotle   & Instacart &            & \\
\bottomrule
\end{tabularx}
\caption{Brands by sector, part 1.}
\label{tab:brands-by-sector-1}
\end{table}

\begin{table}[H]
\centering
\footnotesize
\setlength{\tabcolsep}{3pt}
\begin{tabularx}{\textwidth}{@{}YYYYYY@{}}
\toprule
\textbf{Media, Gaming \& Entertainment} &
\textbf{Travel, Automotive \& Transport} &
\textbf{Finance, Fintech \& Insurance} &
\textbf{B2B, SaaS \& Productivity} &
\textbf{Telco, Utilities \& Infrastructure} &
\textbf{Pet, Parenting \& Purpose} \\
\midrule
Netflix    & Tesla          & Amex        & Salesforce & Verizon  & UNICEF \\
PlayStation& BMW            & Mastercard  & HubSpot    & Telstra  & WWF \\
Xbox       & Emirates       & Monzo       & Slack      & Orange   & Huggies \\
HBO        & Airbnb         & Klarna      & Zoom       & Vodafone & Pampers \\
TikTok     & Toyota         & Stripe      & Adobe      & AT\&T    & Purina \\
Roblox     & Delta Airlines & Robinhood   & Canva      & T-Mobile & Pedigree \\
Spotify    & Marriott       & Revolut     & Notion     & BT       & Unesco \\
YouTube    & Uber           & Allianz     & Shopify    & Starlink & Lego \\
           &                &             & Mailchimp  &          & \\
\bottomrule
\end{tabularx}
\caption{Brands by sector, part 2.}
\label{tab:brands-by-sector-2}
\end{table}

\section{Model Evaluation via Human Rating}
\subsection{Interface Design}

\begin{figure}[H]
  \centering
  \includegraphics[width=\textwidth]{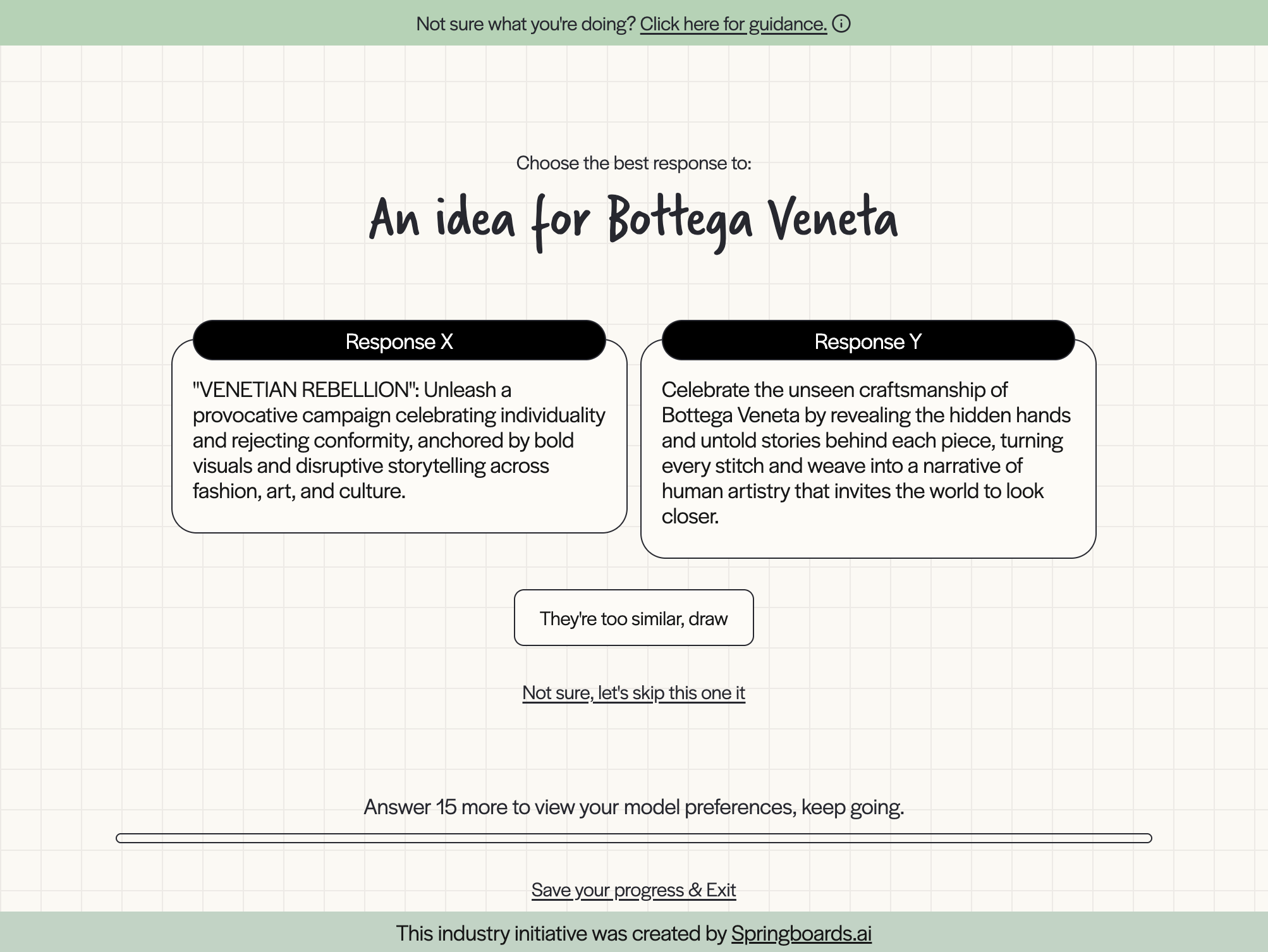}
  \caption{Interface views by participants.}
  \label{fig:interface}
\end{figure}

\subsection{Participant Demographics}\label{app:demographics}
\begin{figure}[H]
\centering

\begin{subfigure}[t]{0.49\textwidth}
  \centering
  \includegraphics[width=\linewidth]{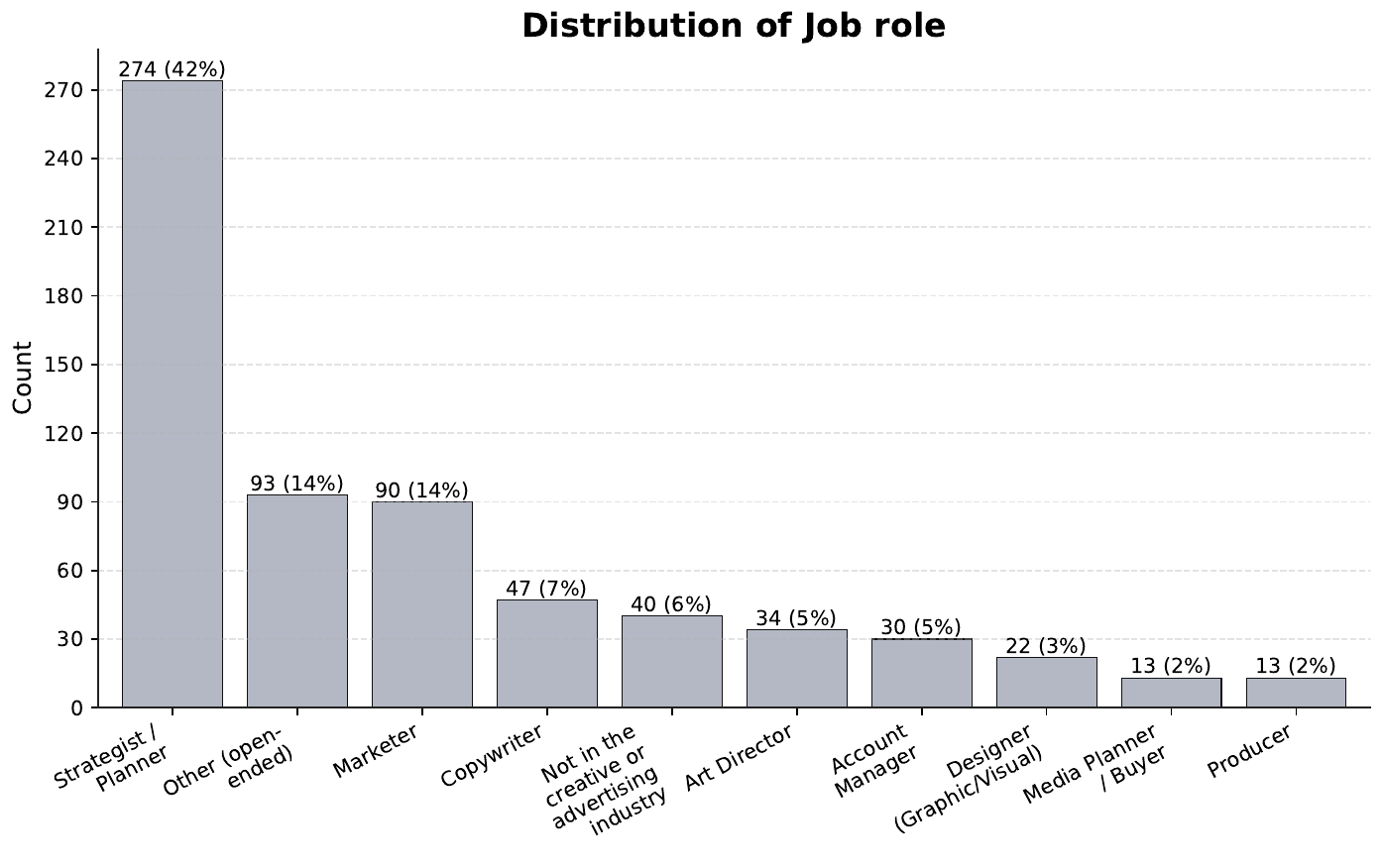}
  \caption{Job role}
  \label{fig:demo-jobrole}
\end{subfigure}
\hfill
\begin{subfigure}[t]{0.49\textwidth}
  \centering
  \includegraphics[width=\linewidth]{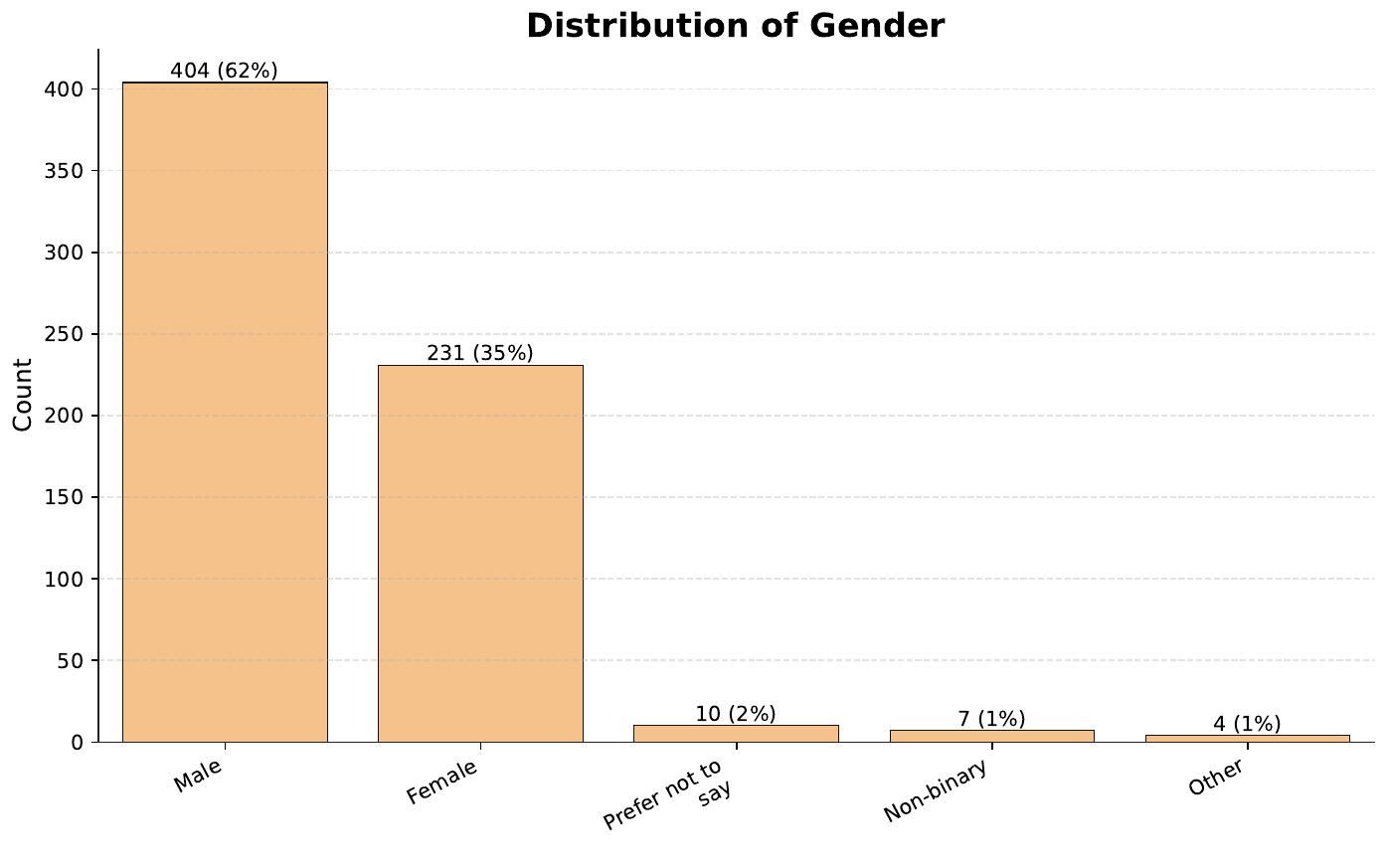}
  \caption{Gender}
  \label{fig:demo-gender}
\end{subfigure}

\vspace{0.5em}

\begin{subfigure}[t]{0.49\textwidth}
  \centering
  \includegraphics[width=\linewidth]{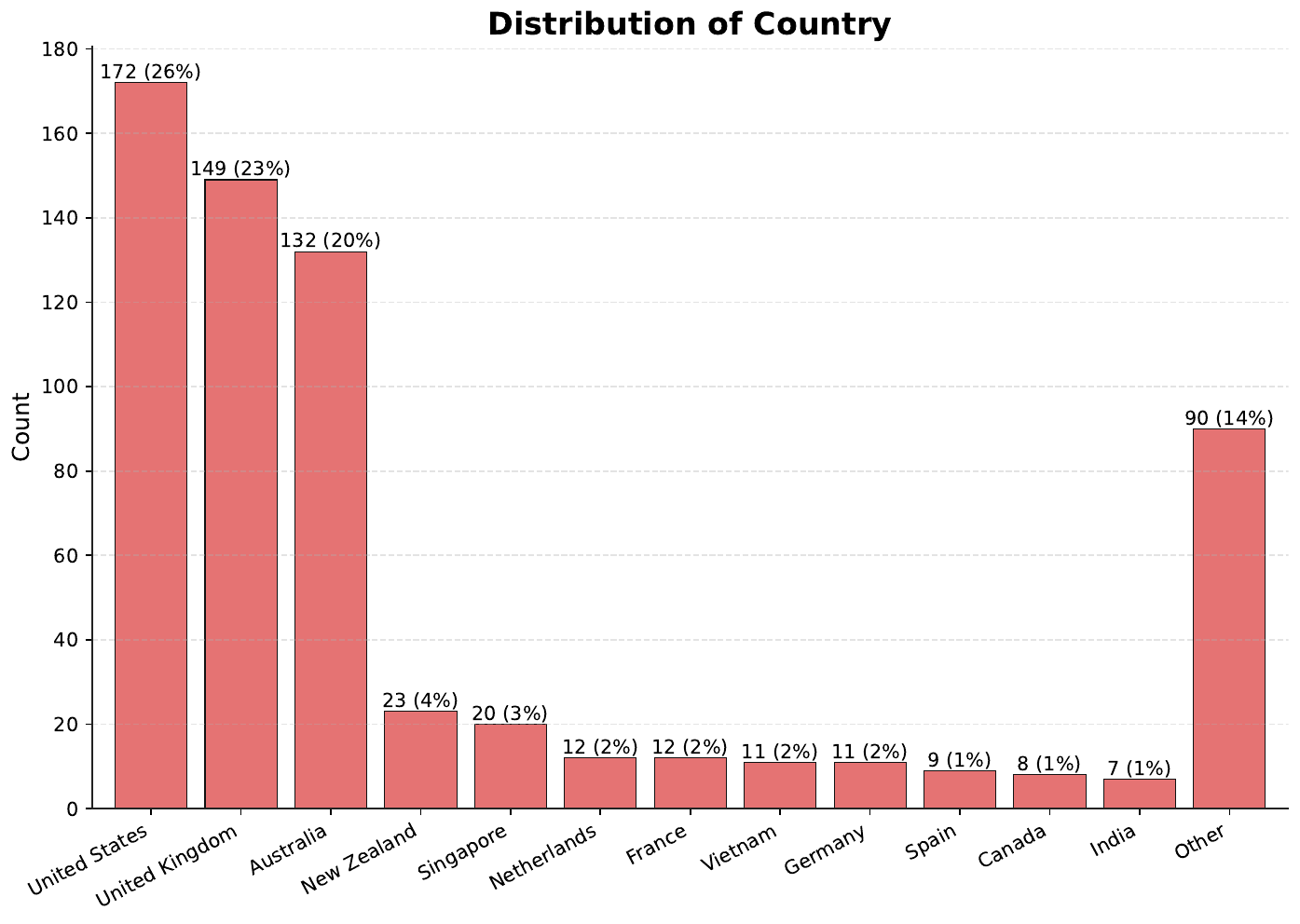}
  \caption{Country}
  \label{fig:demo-country}
\end{subfigure}
\hfill
\begin{subfigure}[t]{0.49\textwidth}
  \centering
  \includegraphics[width=\linewidth]{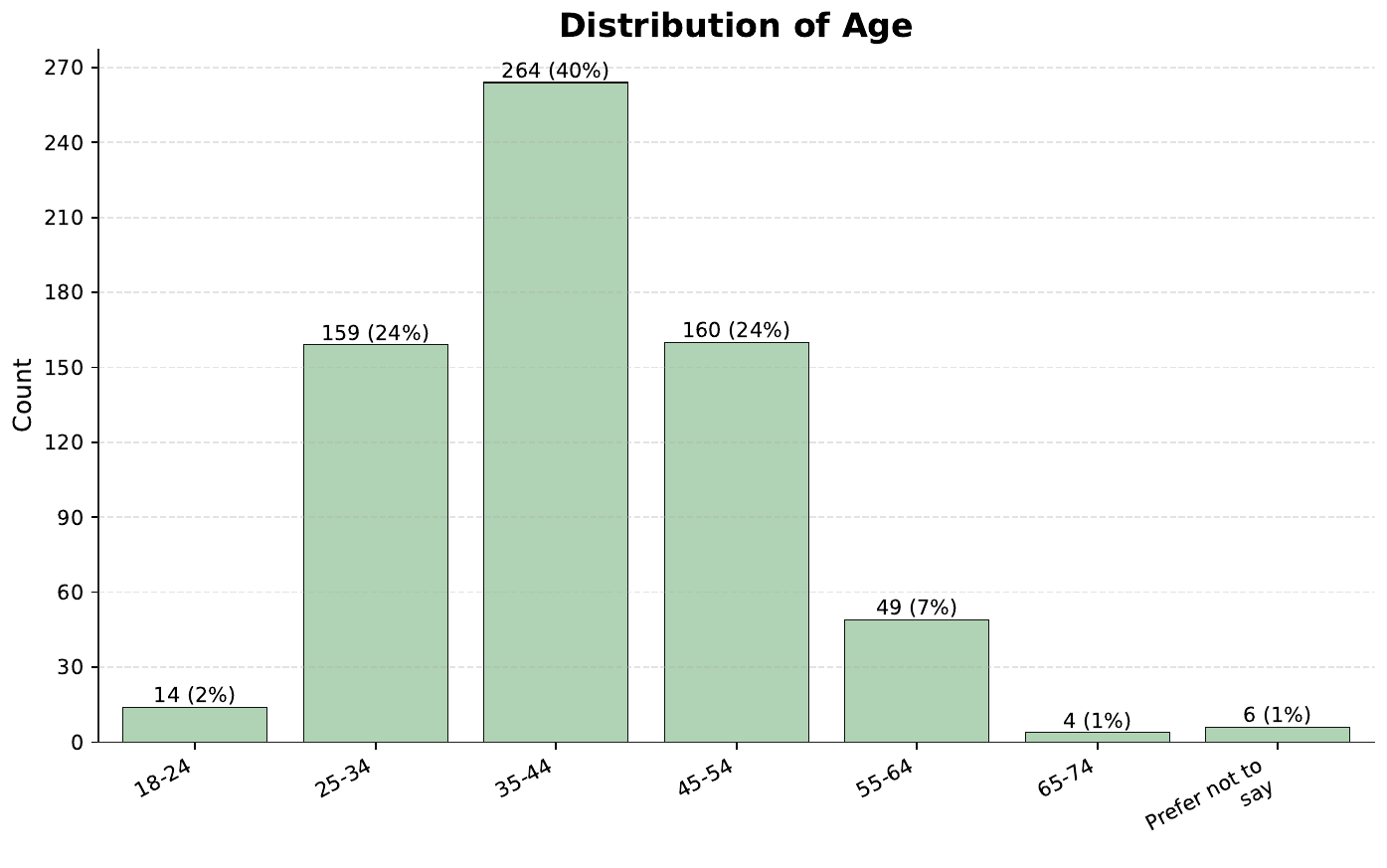}
  \caption{Age}
  \label{fig:demo-age}
\end{subfigure}

\caption{Participant demographics by job role, gender, country, and age.}
\label{fig:participant-demographics}
\end{figure}

Demographic questions were optional, 656 of 678 participants provided at least one response. Summary plots can be seen in \Cref{fig:participant-demographics}. Here we discuss composition and implications for interpreting the results.

\paragraph{Professional background.}
Strategists and planners form the largest group, with substantial representation from marketers, copywriters, art directors, account managers, and designers. An open ended “other” category captures adjacent roles such as founders, researchers, and analysts. A small minority reported not working in the creative or advertising industry. This mix indicates that most judgments come from people who routinely evaluate ideas against briefs, which supports validity for brand work. The presence of non industry participants is limited, and sensitivity checks that exclude them leave the model rankings essentially unchanged.

\paragraph{Gender.}
Responses skew male, with a sizeable female contingent and very small non binary and other groups. This imbalance reflects the channels used for recruitment and common industry demographics. It implies that aggregate preferences may lean toward perspectives more common among men. We therefore report gender specific analyses and caution against over interpreting patterns in small groups.

\paragraph{Age.}
The distribution peaks in the 35-44 and 45-54 bands, with meaningful participation from 25-34 and modest representation at the tails. This profile is consistent with mid career professionals who are likely to hold decision making roles on strategy and creative work. It also means that our estimates primarily reflect preferences of experienced practitioners rather than students or retirees. 

\paragraph{Geography.}
Participants are concentrated in the United States, the United Kingdom, and Australia, with additional coverage across Europe and Asia.

\subsection{Head-to-head win rates}

\begin{figure}[H]
  \centering
  \includegraphics[width=\textwidth]{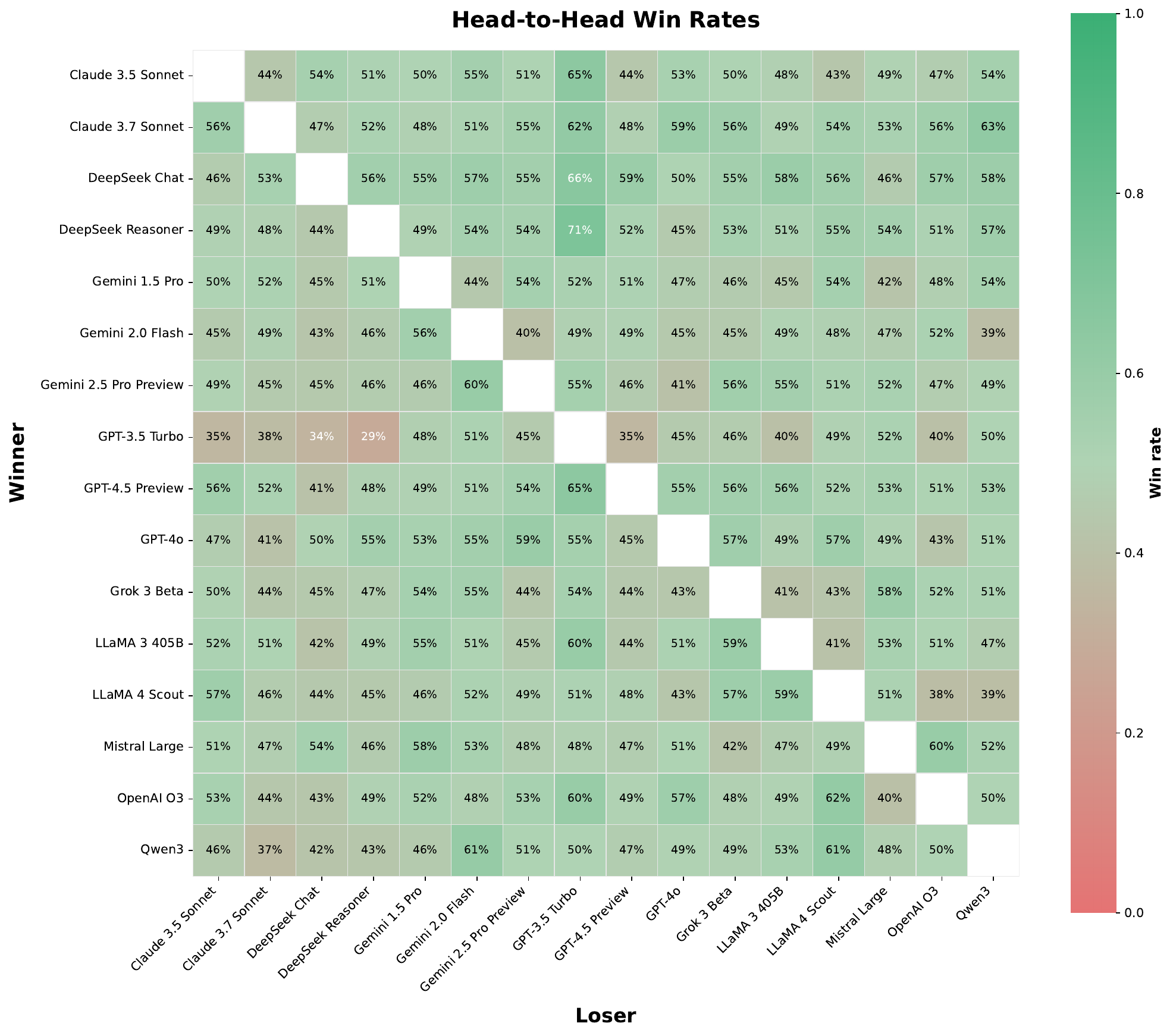}
  \caption{Head-to-head win-rate matrix from \emph{human voting}. Each cell shows the observed probability that the row model wins against the column model, computed as $\text{wins}/(\text{wins}+\text{losses})$ from raw pairwise votes. Draws (“too similar”) and skips (“not sure”) are omitted from the denominator.}
  \label{fig:h2h_winrates}
\end{figure}

\subsection{Model preference by demographic} \label{app:model_pref_demo}
\subsubsection{Gender}
\begin{figure}[H]
  \centering
  \begin{minipage}[t]{0.32\textwidth}
    \centering
    \includegraphics[width=\linewidth]{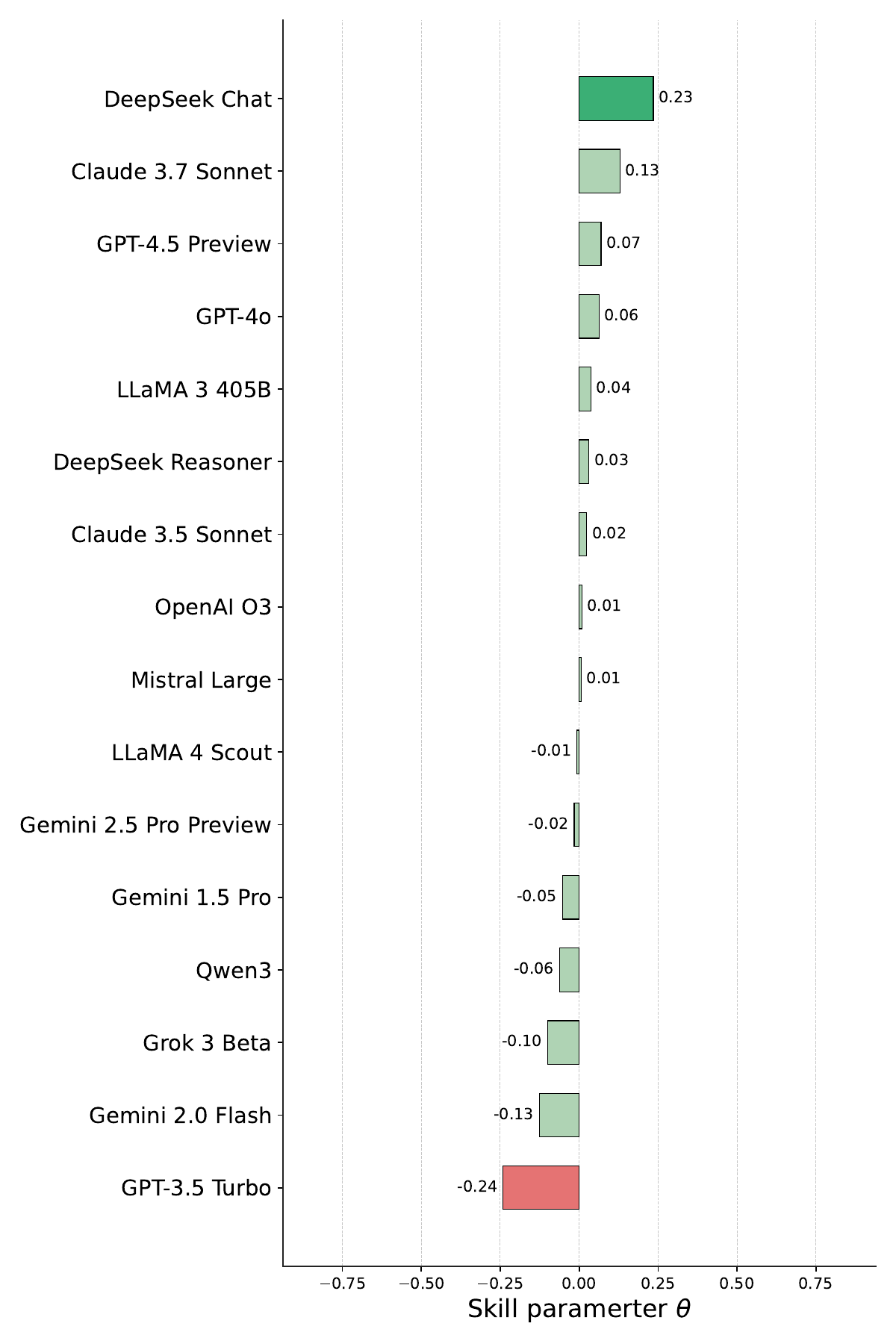}
    \caption*{(a) Male (n=6860)}
  \end{minipage}\hfill
  \begin{minipage}[t]{0.32\textwidth}
    \centering
    \includegraphics[width=\linewidth]{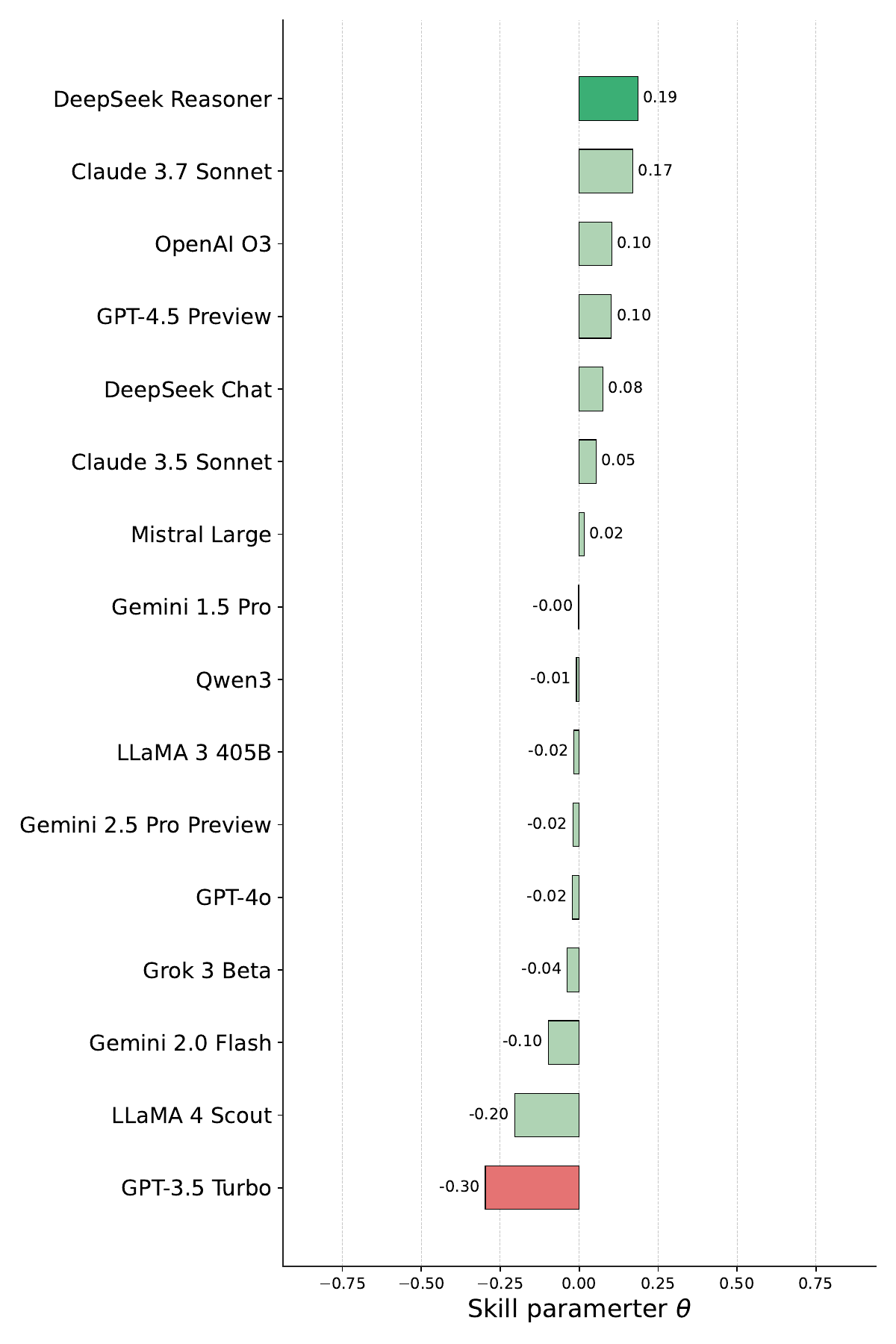}
    \caption*{(b) Female (n=3871)}
  \end{minipage}\hfill
  \begin{minipage}[t]{0.32\textwidth}
    \centering
    \includegraphics[width=\linewidth]{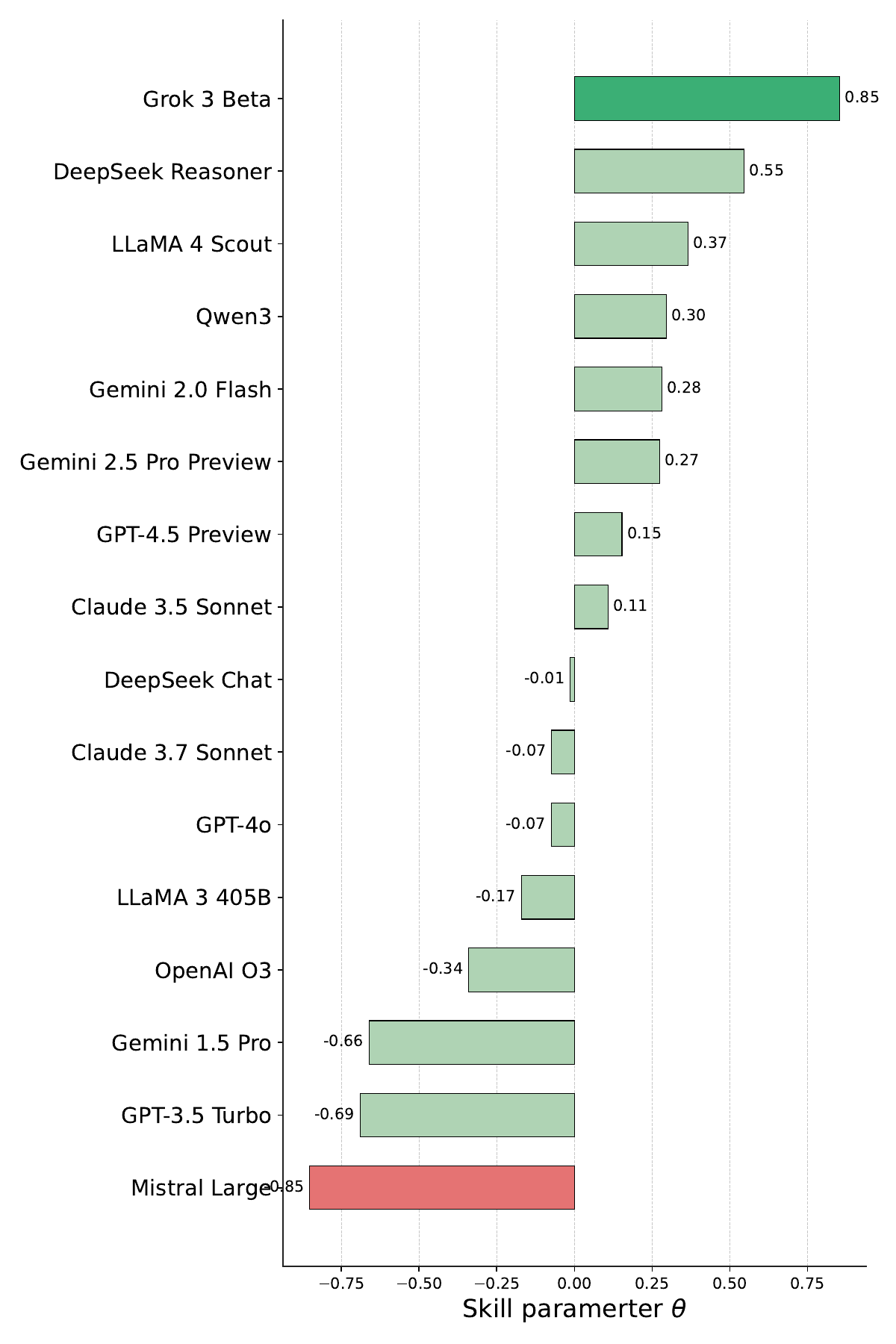}
    \caption*{(c) Non-binary (n=128)}
  \end{minipage}

  \caption{Model ranking across three prompt types.}
  \label{fig:model_pref_geneder}
\end{figure}

Among male respondents ($n=6860$), DeepSeek Chat leads ($\theta=0.23$), followed by Claude 3.7 Sonnet ($0.13$) and GPT-4.5 Preview ($0.07$). The spread from best to worst is approximately $0.47$. Among female respondents ($n=3871$), DeepSeek Reasoner ($0.19$) and Claude 3.7 Sonnet ($0.17$) rank highest, with OpenAI O3 and GPT-4.5 Preview tied for third at $0.10$. The overall dispersion is similar at about $0.49$. The non-binary cohort is small ($n=128$) and yields extreme estimates (Grok 3 Beta $\approx 0.85$, Mistral Large $\approx-0.85$), these swings are likely driven by sampling noise rather than large underlying differences. Overall, the leading set overlaps across male and female cohorts, with DeepSeek and Claude variants appearing near the top.

\subsubsection{Age}
\begin{figure}[H]
  \centering
  \begin{subfigure}[t]{0.23\textwidth}
    \centering
    \includegraphics[width=\linewidth]{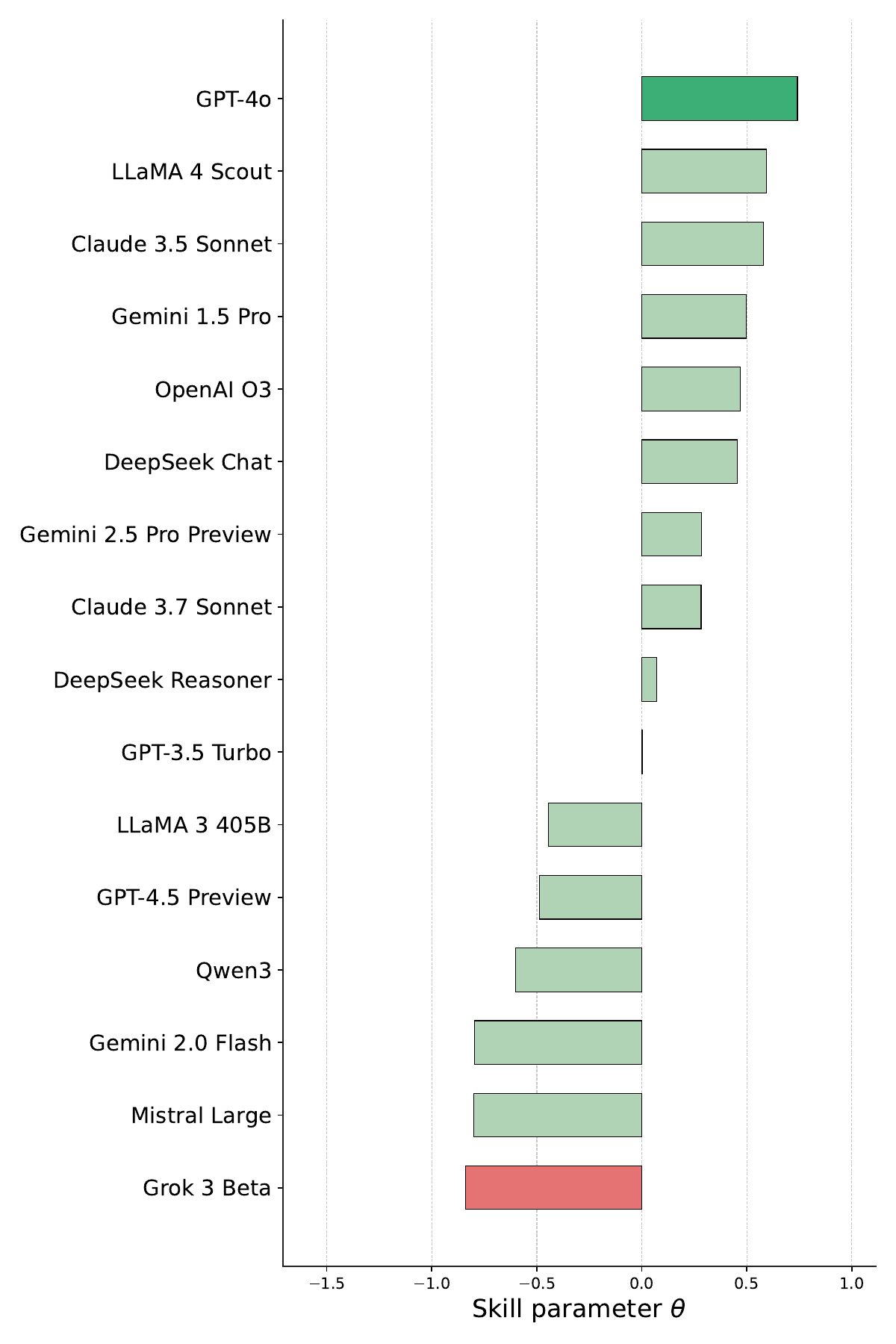}
    \caption{Age 18-24}
  \end{subfigure}\hfill
  \begin{subfigure}[t]{0.23\textwidth}
    \centering
    \includegraphics[width=\linewidth]{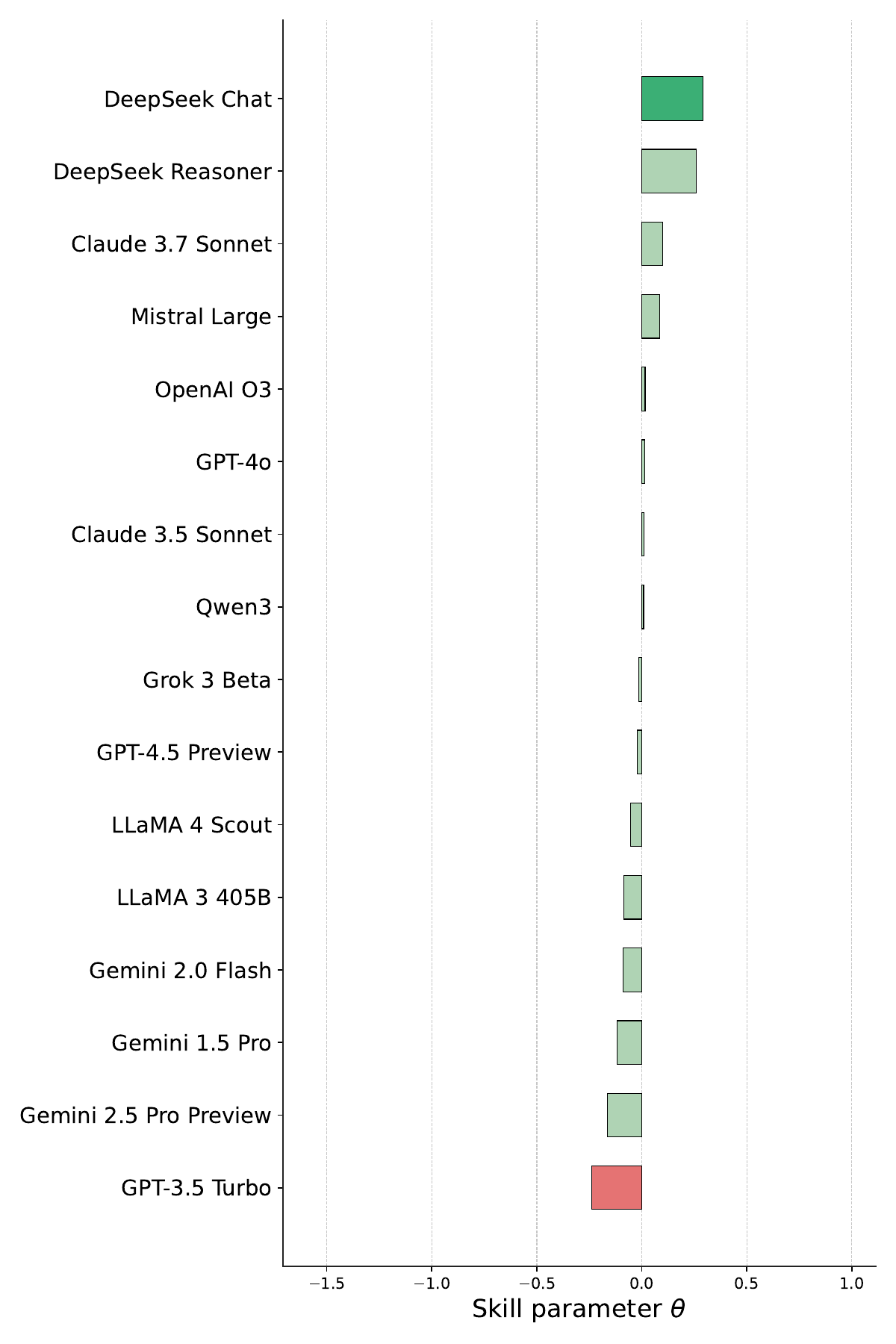}
    \caption{Age 25-34}
  \end{subfigure}\hfill
  \begin{subfigure}[t]{0.23\textwidth}
    \centering
    \includegraphics[width=\linewidth]{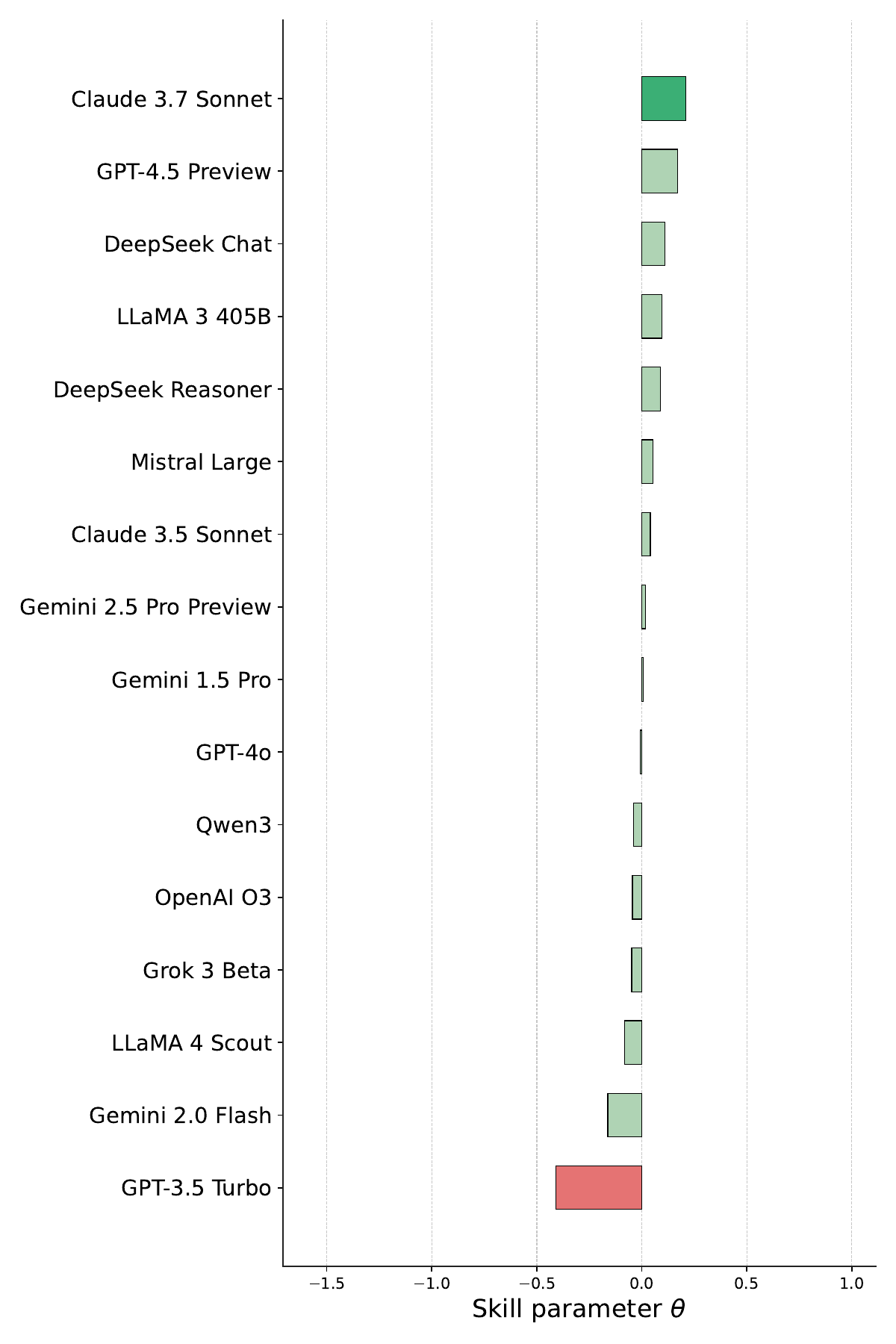}
    \caption{Age 35-44}
  \end{subfigure}\hfill
  \begin{subfigure}[t]{0.23\textwidth}
    \centering
    \includegraphics[width=\linewidth]{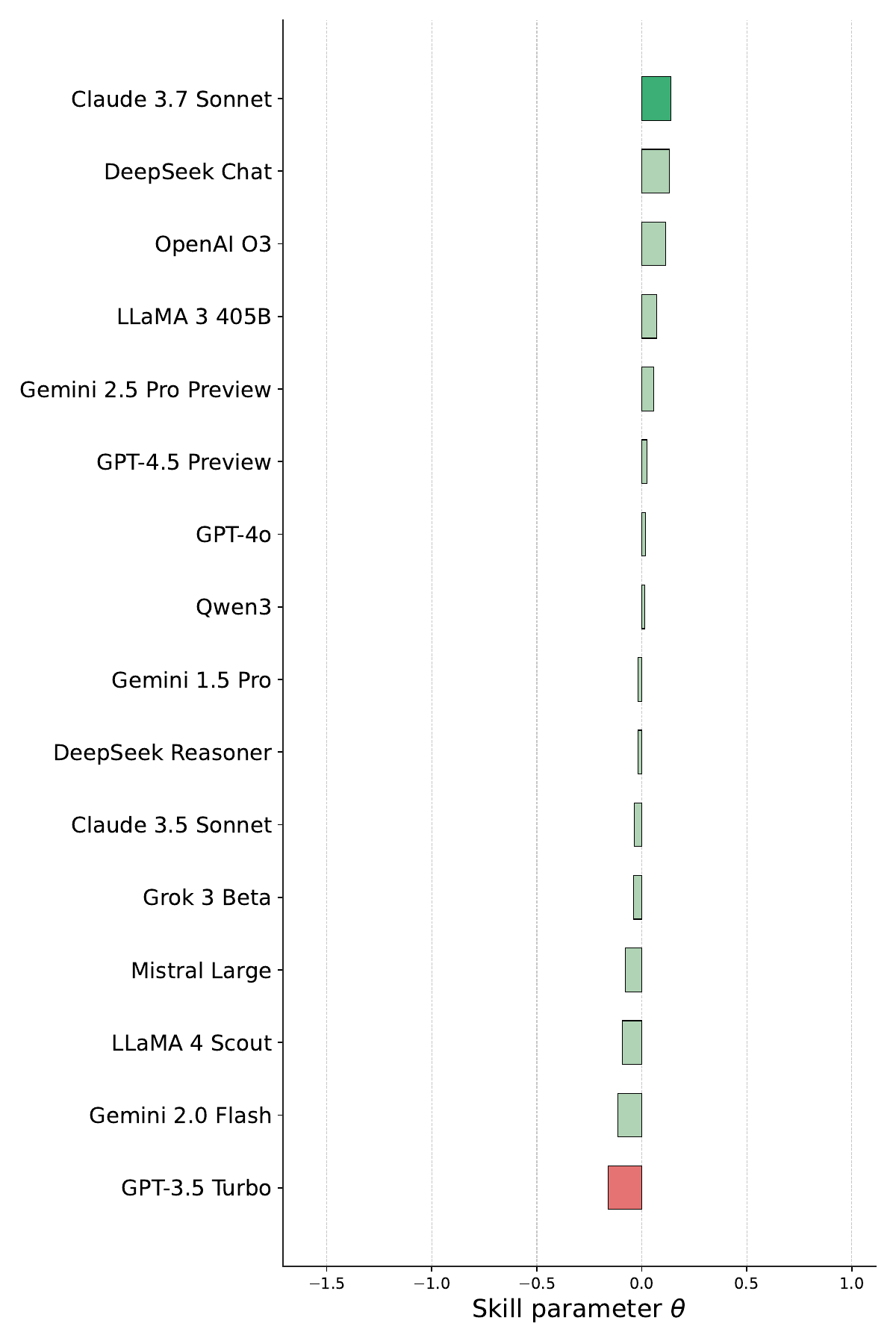}
    \caption{Age 45-54}
  \end{subfigure}

  \vspace{0.5em}

  \begin{subfigure}[t]{0.23\textwidth}
    \centering
    \includegraphics[width=\linewidth]{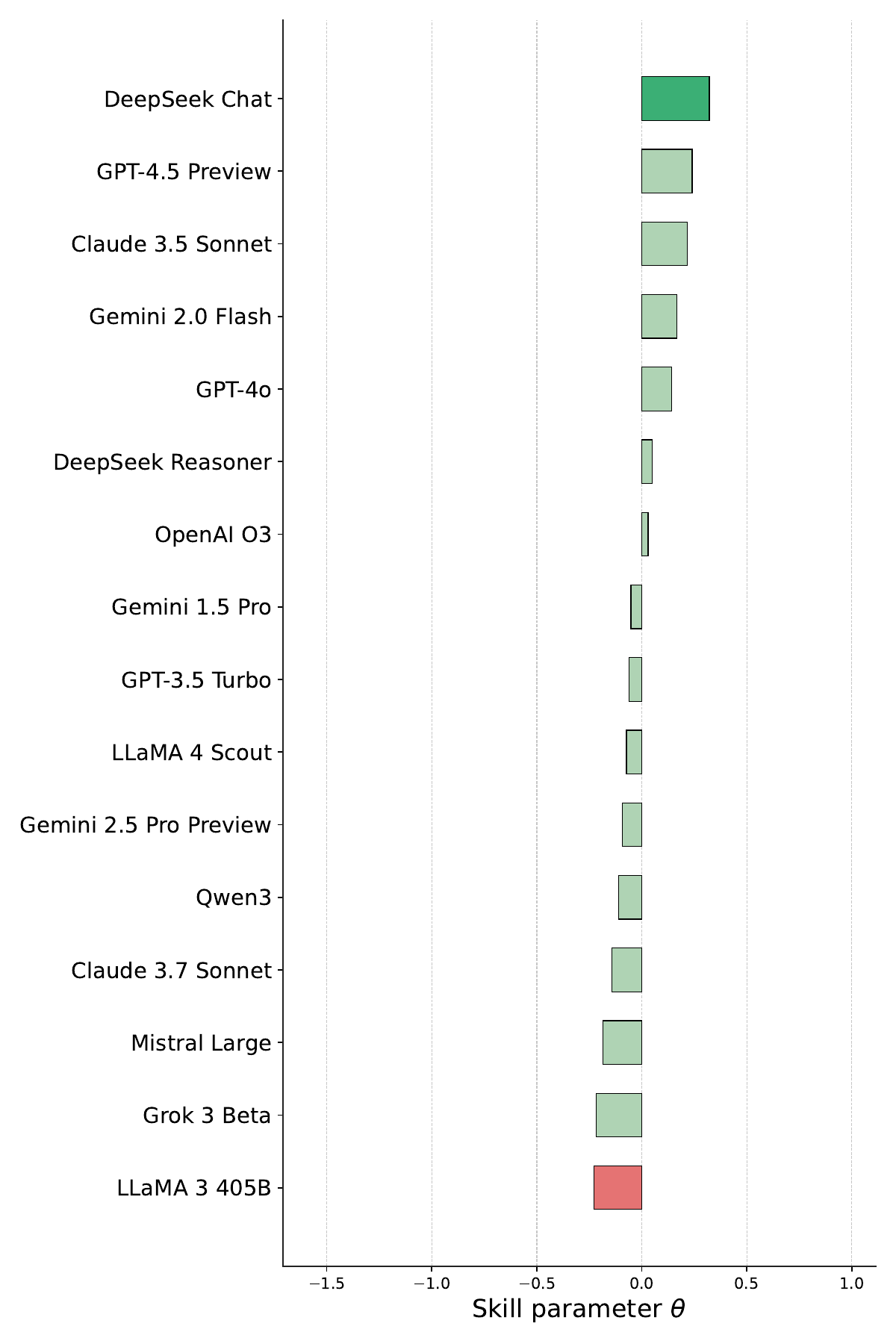}
    \caption{Age 55-64}
  \end{subfigure}\hfill
  \begin{subfigure}[t]{0.23\textwidth}
    \centering
    \includegraphics[width=\linewidth]{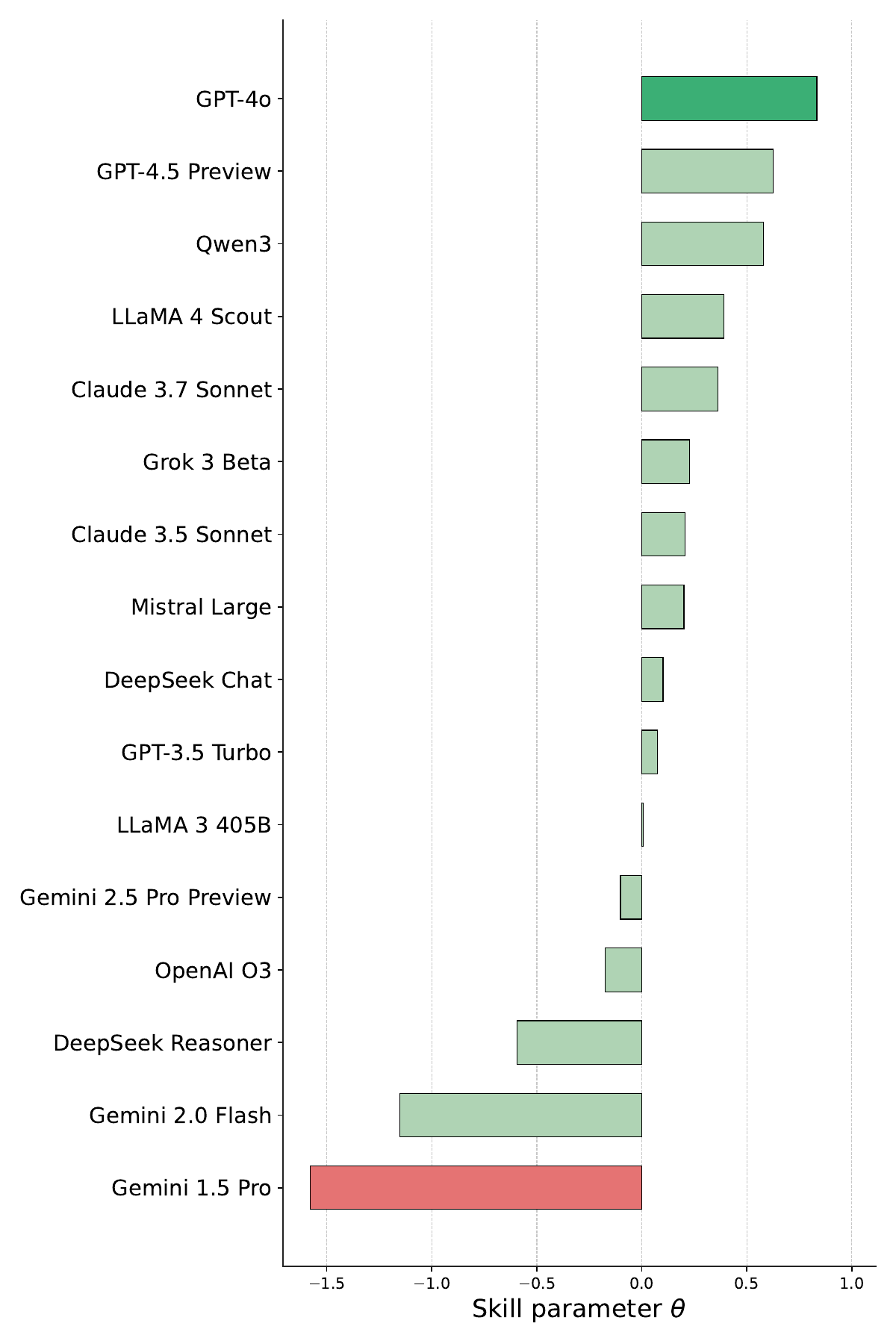}
    \caption{Age 65-74}
  \end{subfigure}\hfill
  \begin{subfigure}[t]{0.23\textwidth}
    \centering
    \includegraphics[width=\linewidth]{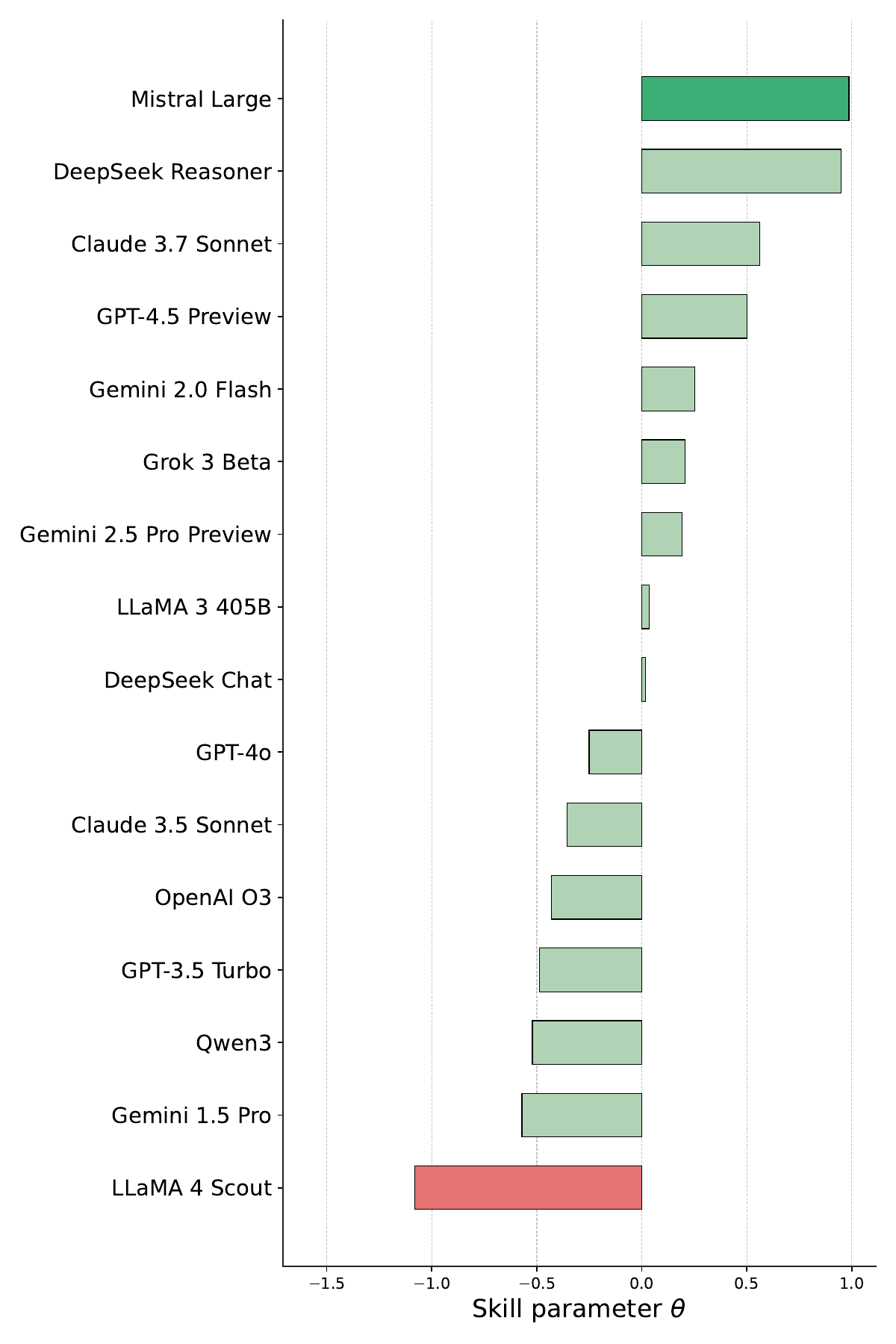}
    \caption{Prefer not to say}
  \end{subfigure}

  \caption{Bradley-Terry strengths ($\theta$) across age groups in ascending order. Higher values indicate greater model preference.}
  \label{fig:bt_age_all}
\end{figure}

Across the mid-career cohorts (25-34, 35-44, 45-54, 55-64; \Cref{fig:bt_age_all}), model ratings are broadly consistent, rank differences are small and most $\theta$ estimates lie within $\pm 0.15$. The remaining cohorts (18-24, 65-74, and prefer-not-to-say; \Cref{fig:bt_age_all}) show larger apparent shifts, but each has fewer than $200$ evaluations, so these differences may reflect sampling variability rather than systematic age effects.

 DeepSeek Chat appears in the top three for five of the seven cohorts, indicating cross-cohort appeal. GPT-4o is the leading model among the youngest participants (18-24, $n=192$) and the oldest group (65-74, $n=97$). Within the largest cohorts (ages 25 to 54, combined $n\approx 9{,}600$), dispersion contracts: roughly two thirds of model $\theta$ estimates fall within $\pm 0.15$, suggesting that most leading LLMs deliver comparably acceptable outputs for these participants. Extreme $\theta$ values occur mainly in the smallest samples, so those rankings should be interpreted cautiously given wider uncertainty. Across the mid-career cohorts (25–34, 35–44, 45–54, 55–64; \Cref{fig:bt_age_all}), model ratings are broadly consistent, rank differences are small, and most $\theta$ estimates lie within $\pm 0.15$. The remaining cohorts (18–24, 65–74, and prefer-not-to-say; \Cref{fig:bt_age_all}) show larger apparent shifts, but each has fewer than $200$ evaluations, so these differences may reflect sampling variability rather than systematic age effects.

\section{LLM-as-judge}
\subsection{Judge Prompts (full text)} \label{app:judge_prompts}

\begin{promptbox}{Creative Strategist}
\begin{prompttext}
You are a creative strategist judging two ideas. Judge them only on three things:
Unique - Could only be this idea, not a generic version.
Unexpected - Surprises, flips, or reframes in a fresh way.
Unforgettable - Sticks in the mind with a clear, defining moment.
Compare Response X and Response Y on these criteria. Don’t reward safe, predictable, or “medium” work. Favor ideas that feel distinct, surprising, and impossible to forget.
Reply with only one of: X, Y, or Tie.
\end{prompttext}
\end{promptbox}

\begin{promptbox}{EQ Benchmark}
\begin{prompttext}
Compare the relative ability of each writer on these criteria:

- Character authenticity and insight
- Interesting and original
- Writing quality
- Coherence in plot, character choices, metaphor
- Instruction following (followed the prompt)
- World and atmosphere
- Avoids flowery verbosity & show-offy vocab maxxing
- Avoids gratuitous metaphor or poetic overload

Judging notes:

- Be aware that these abilities may be independent, i.e. a model may be strong in one and weak in another.
- Outputs will sometimes be truncated to ensure length consistency. Don't penalise this, just judge what is there on its merit.

Reply with only one of: X, Y, or Tie.
\end{prompttext}
\end{promptbox}

\begin{promptbox}{Surprising Ideas}
\begin{prompttext}
Assess how original and unexpected the idea is. Prioritise unusual insights over common or conventional ones.
\end{prompttext}
\end{promptbox}

\subsection{User prompt template}

\begin{promptbox}{Judge user prompt}
\begin{prompttext}
### Brand:
{brand}

### Brief:
{prompt_text}

### Response X:
{response_x}

### Response Y:
{response_y}

Which response better answers the brief? Reply with only one of: X, Y, or Tie.
\end{prompttext}
\end{promptbox}

\subsection{Model performance by creative strategist prompt} \label{app:creative_strategy}

\begin{figure}[H]
  \centering
  \begin{subfigure}[t]{0.24\textwidth}
    \centering
    \includegraphics[width=\linewidth]{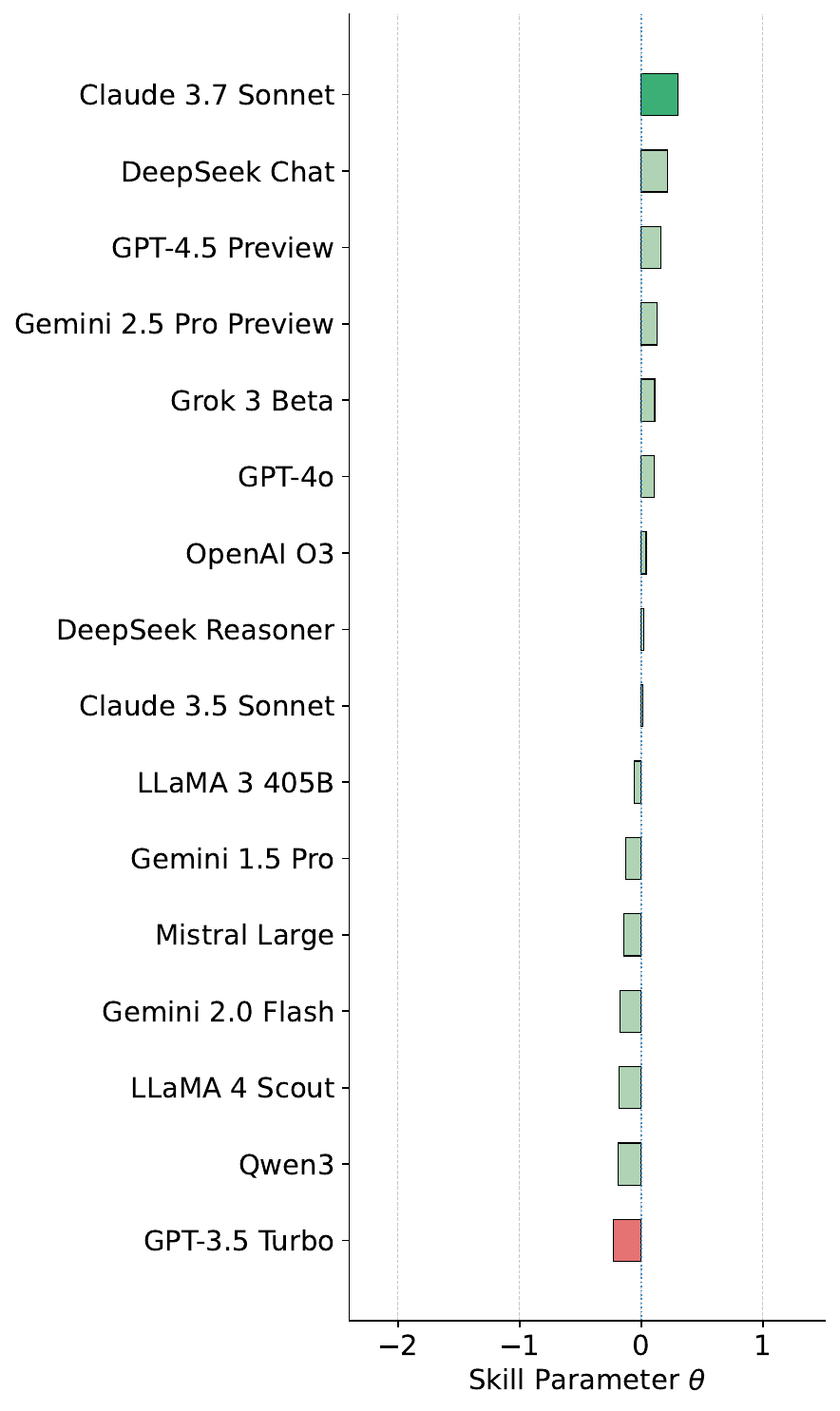}
    \caption{Human raters}
  \end{subfigure}\hfill
  \begin{subfigure}[t]{0.24\textwidth}
    \centering
    \includegraphics[width=\linewidth]{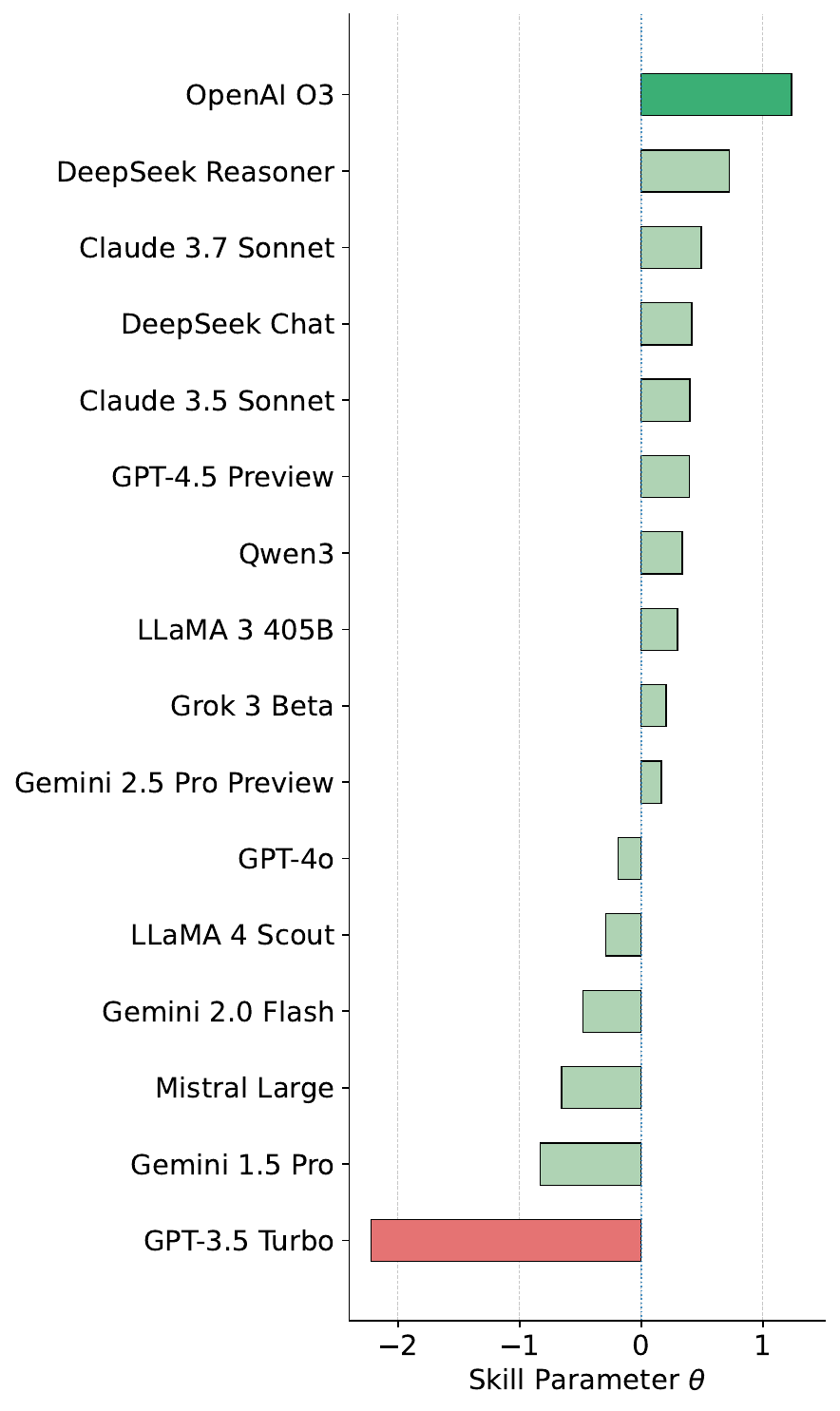}
    \caption{Claude 4 Sonnet (judge)}
  \end{subfigure}\hfill
  \begin{subfigure}[t]{0.24\textwidth}
    \centering
    \includegraphics[width=\linewidth]{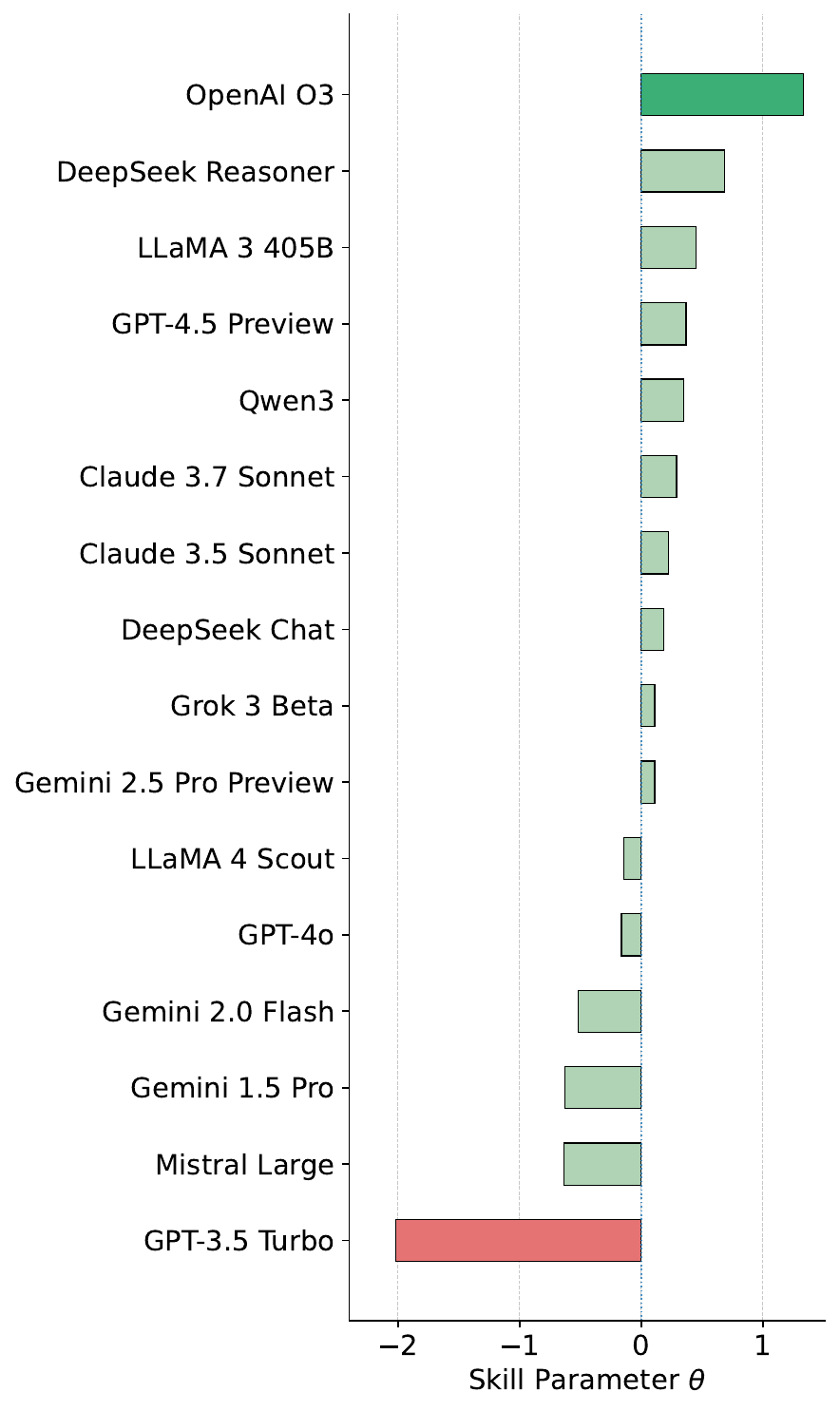}
    \caption{GPT-4o (judge)}
  \end{subfigure}\hfill
  \begin{subfigure}[t]{0.24\textwidth}
    \centering
    \includegraphics[width=\linewidth]{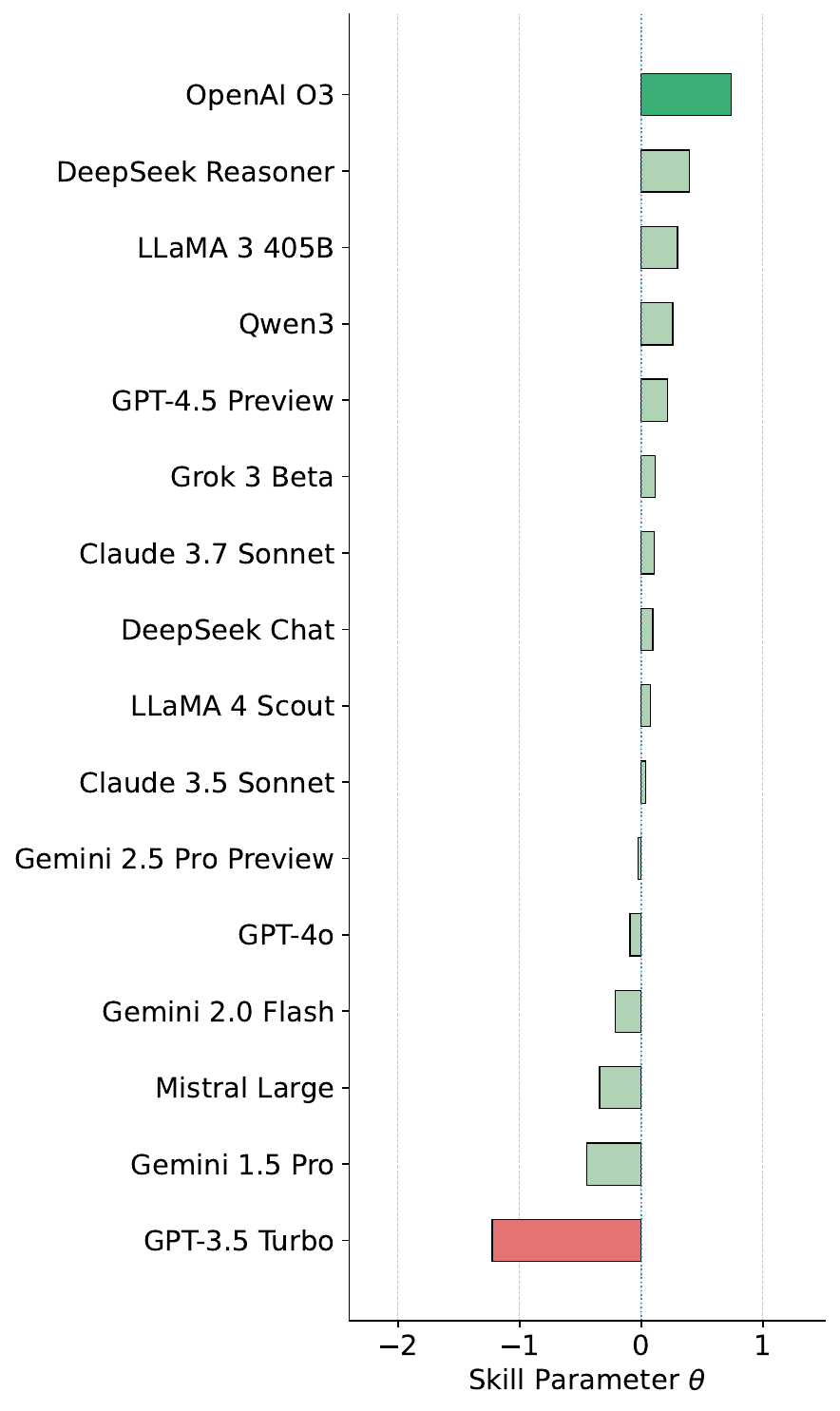}
    \caption{Gemini 2.0 Flash (judge)}
  \end{subfigure}

  \caption{Wild Ideas task under the Creative Strategist system prompt. Bradley-Terry strengths ($\theta$) from human raters and three LLM-as-judge evaluators.}
  \label{fig:bt_wildideas_judges_row}
\end{figure}

\begin{figure}[H]
  \centering
  \begin{subfigure}[t]{0.24\textwidth}
    \centering
    \includegraphics[width=\linewidth]{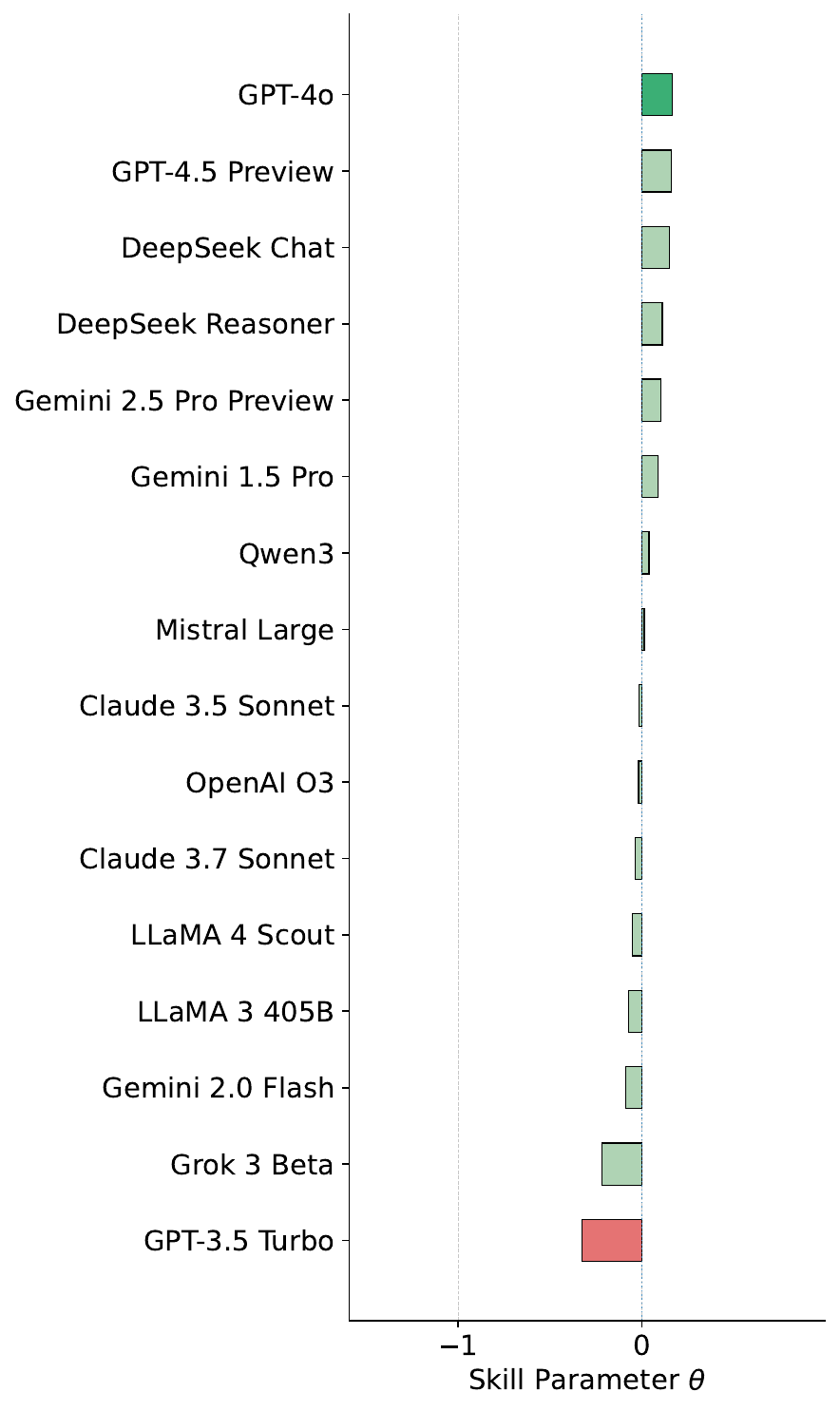}
    \caption{Human raters}
  \end{subfigure}\hfill
  \begin{subfigure}[t]{0.24\textwidth}
    \centering
    \includegraphics[width=\linewidth]{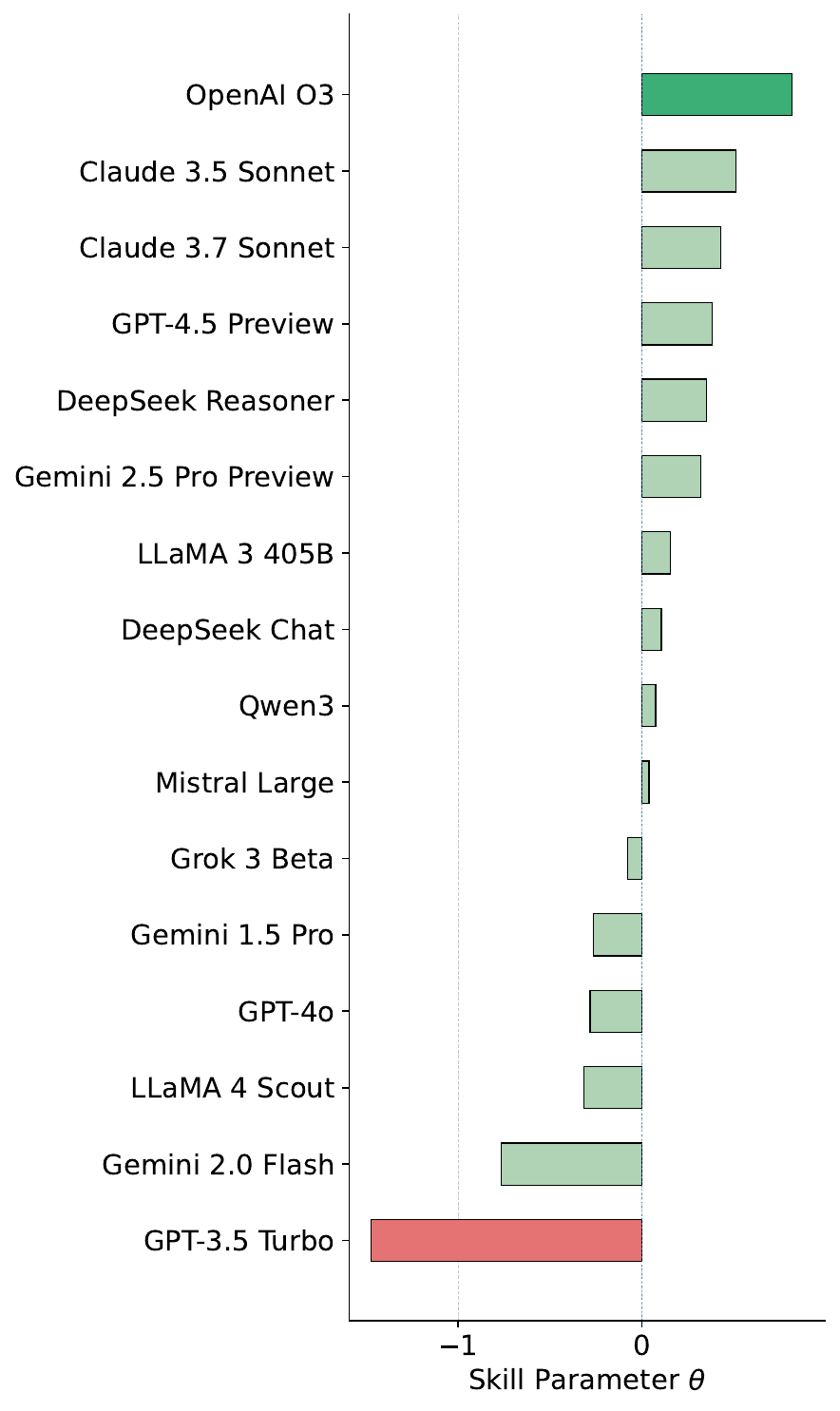}
    \caption{Claude 4 Sonnet (judge)}
  \end{subfigure}\hfill
  \begin{subfigure}[t]{0.24\textwidth}
    \centering
    \includegraphics[width=\linewidth]{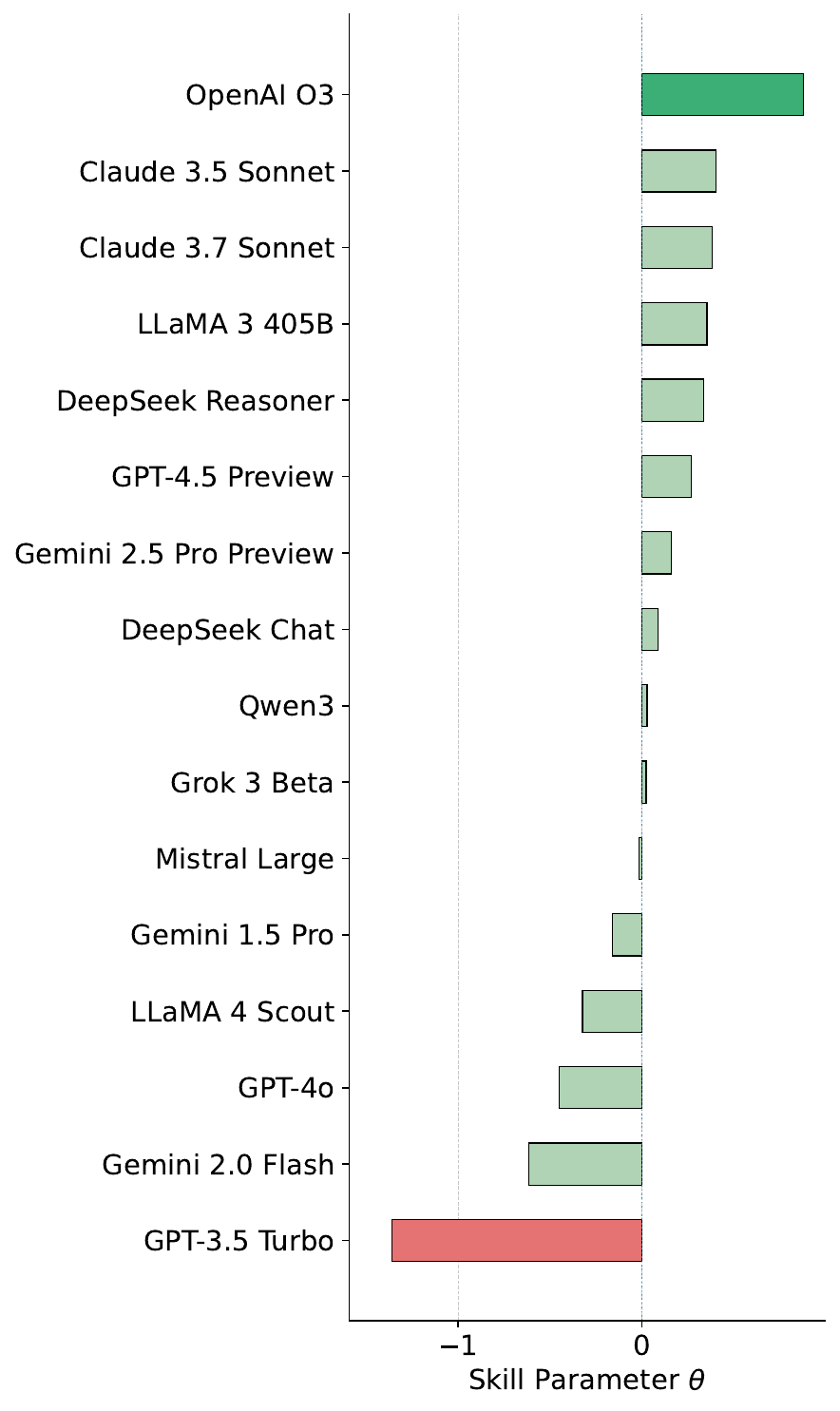}
    \caption{GPT-4o (judge)}
  \end{subfigure}\hfill
  \begin{subfigure}[t]{0.24\textwidth}
    \centering
    \includegraphics[width=\linewidth]{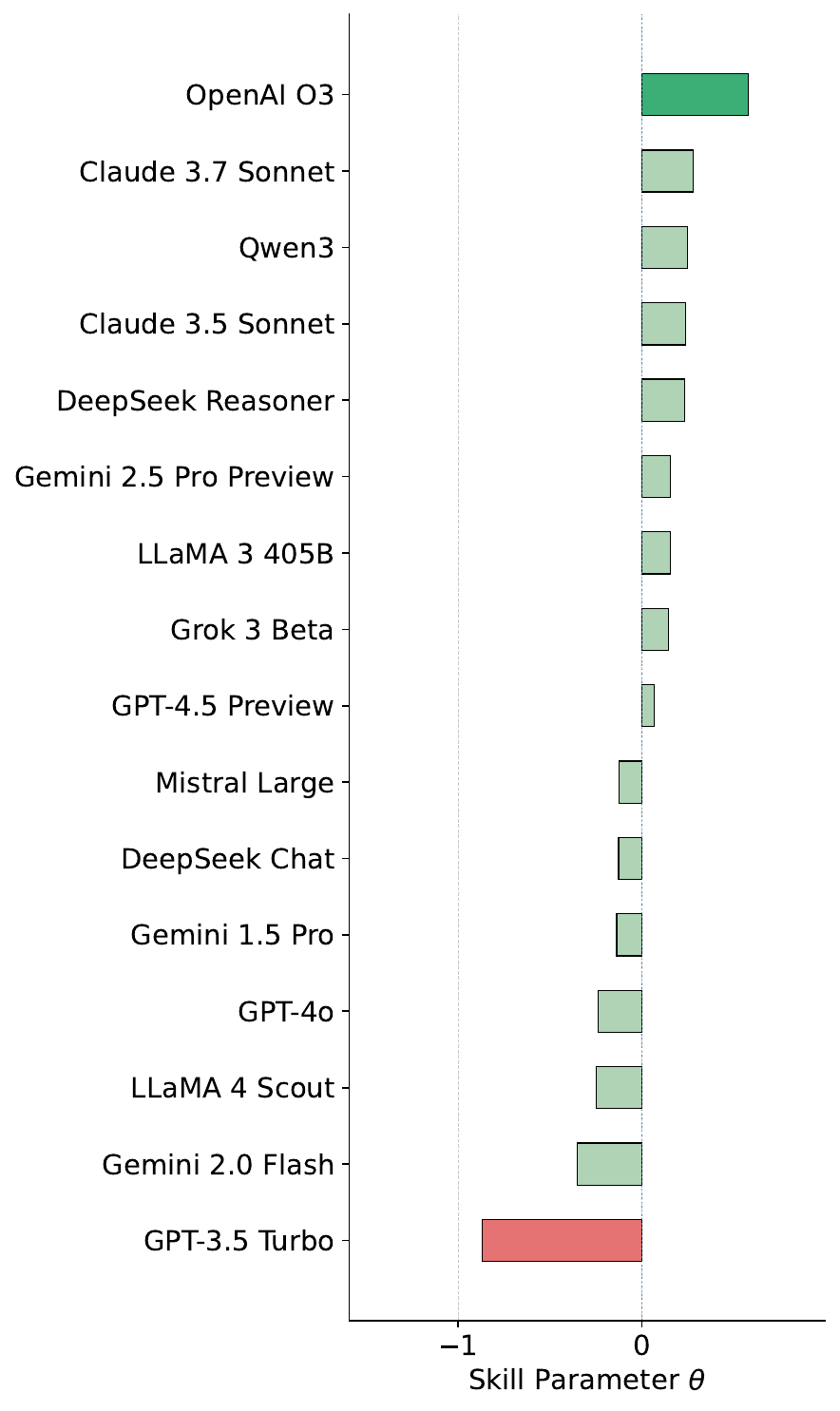}
    \caption{Gemini 2.0 Flash (judge)}
  \end{subfigure}

  \caption{Insights task under the Creative Strategist system prompt. Bradley-Terry strengths ($\theta$) from human raters and three LLM-as-judge evaluators.}
  \label{fig:bt_insights_judges_row}
\end{figure}

\subsection{Model Performance with Surprise Prompt}\label{app:surprise_model_performance}
\begin{figure}[H]
  \centering
  \begin{subfigure}[t]{0.24\textwidth}
    \centering
    \includegraphics[width=\linewidth]{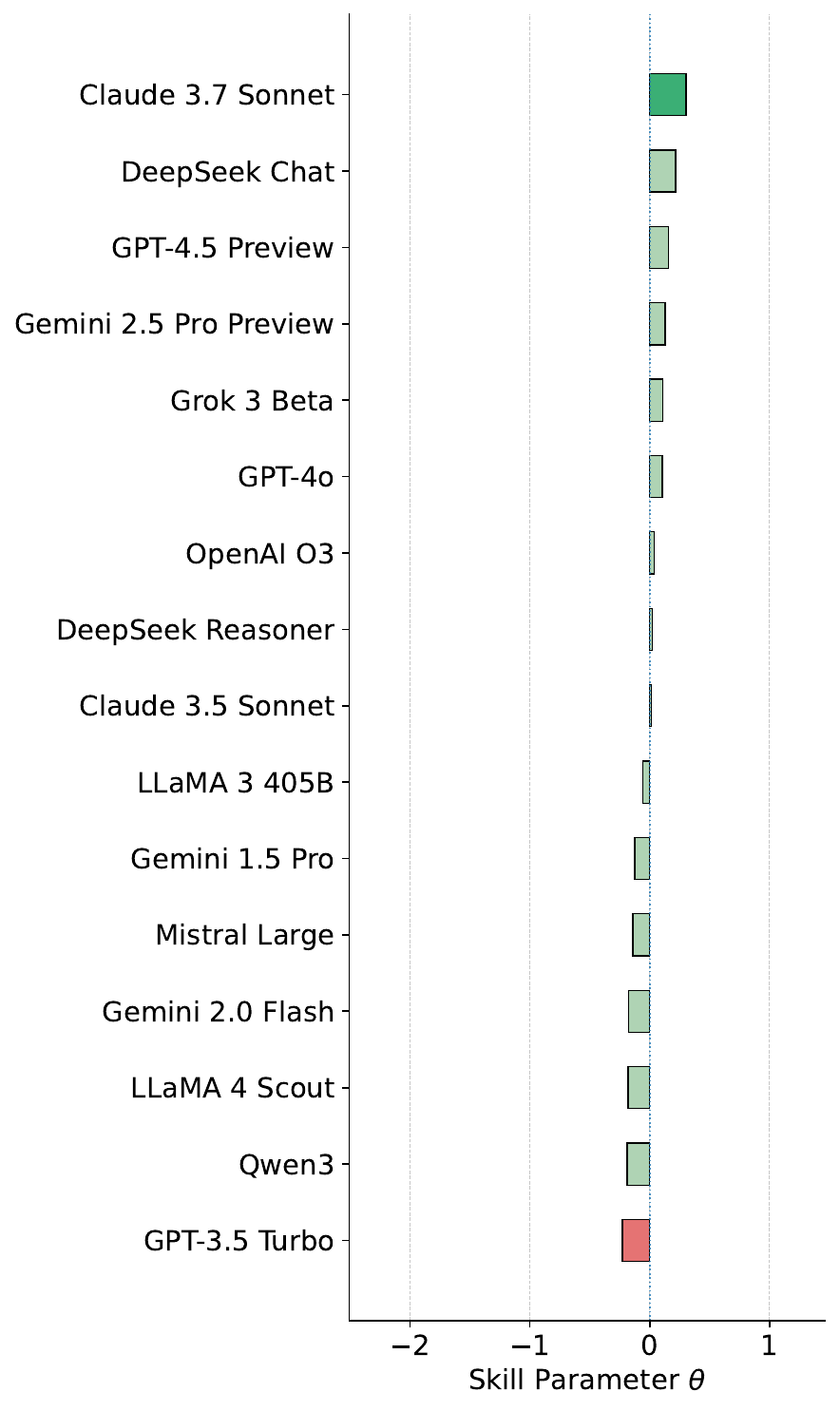}
    \caption{Human raters}
  \end{subfigure}\hfill
  \begin{subfigure}[t]{0.24\textwidth}
    \centering
    \includegraphics[width=\linewidth]{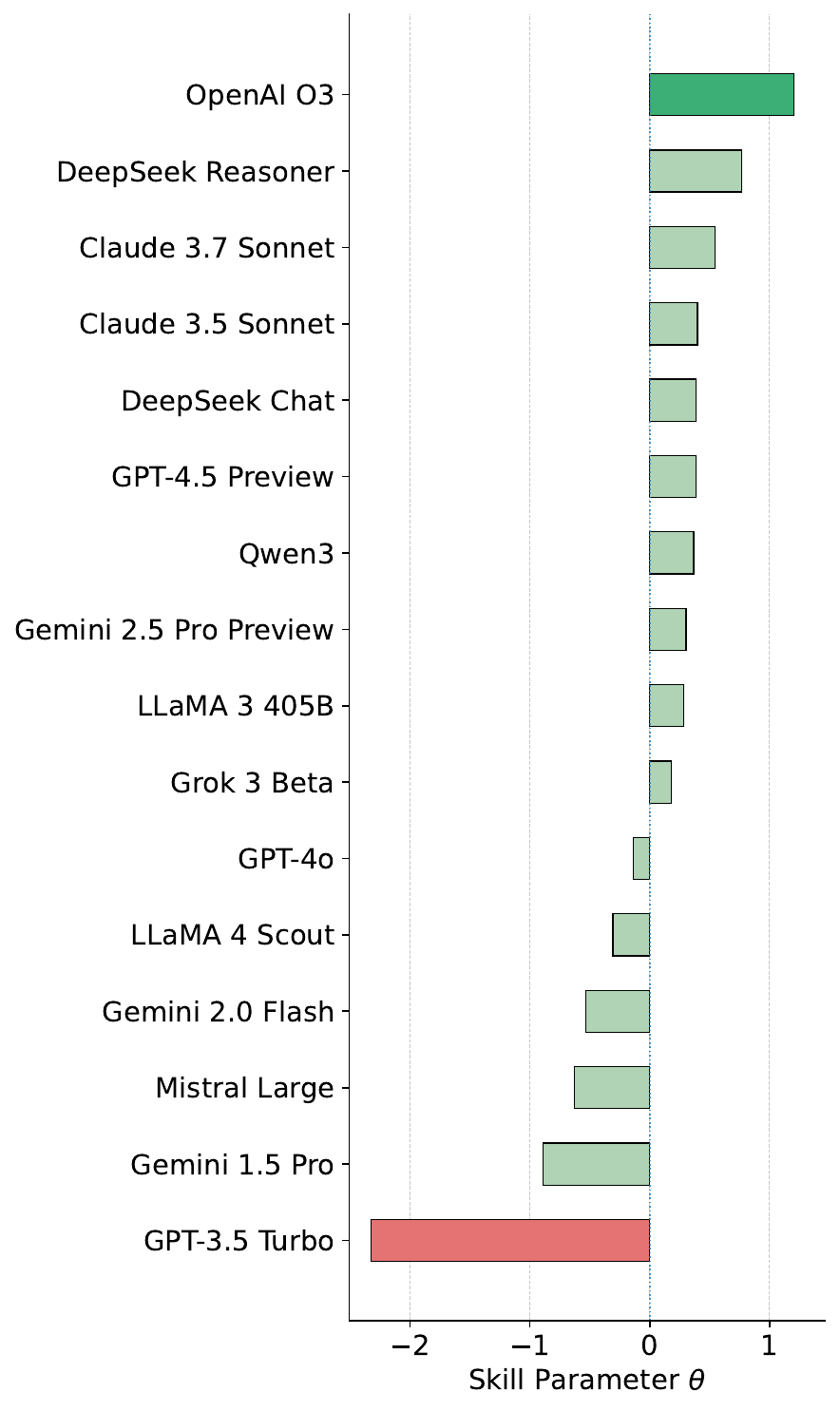}
    \caption{Claude 4 Sonnet (judge)}
  \end{subfigure}\hfill
  \begin{subfigure}[t]{0.24\textwidth}
    \centering
    \includegraphics[width=\linewidth]{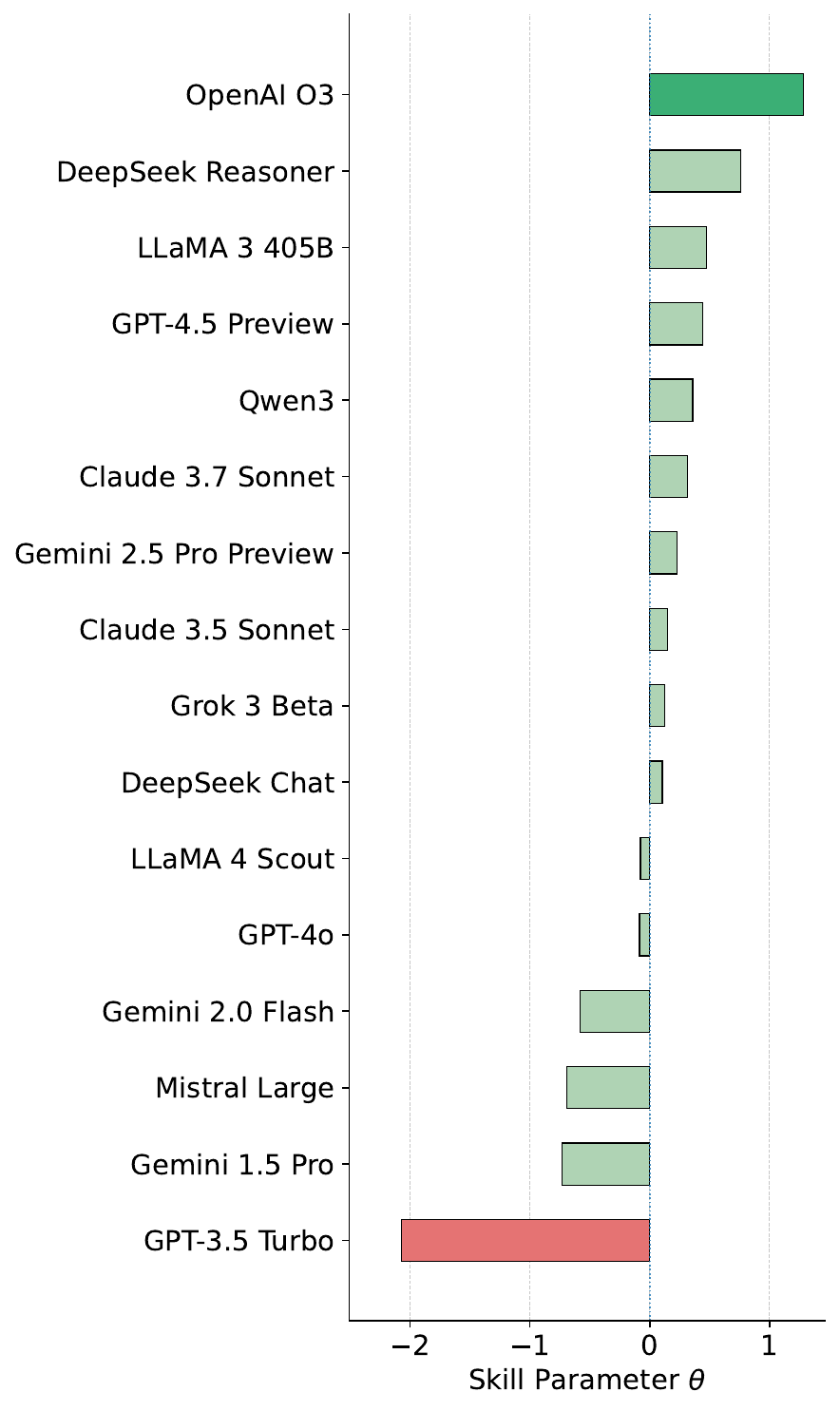}
    \caption{GPT-4o (judge)}
  \end{subfigure}\hfill
  \begin{subfigure}[t]{0.24\textwidth}
    \centering
    \includegraphics[width=\linewidth]{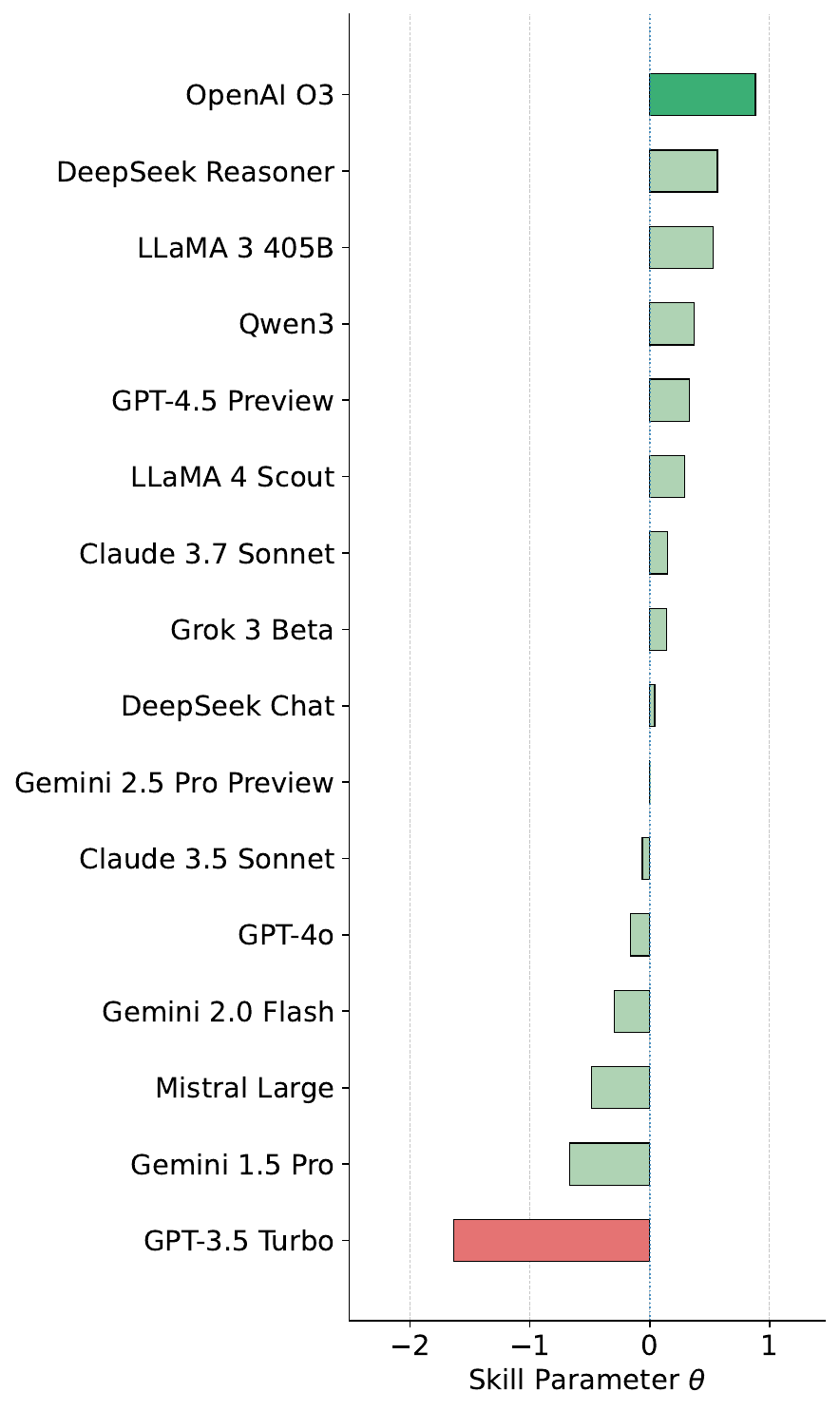}
    \caption{Gemini 2.0 Flash (judge)}
  \end{subfigure}

  \caption{Wild Ideas task under the Surprise judge prompt. Bradley-Terry strengths ($\theta$) from human raters and three LLM-as-judge evaluators.}
  \label{fig:surprise_wildideas_row}
\end{figure}

\begin{figure}[H]
  \centering
  \begin{subfigure}[t]{0.24\textwidth}
    \centering
    \includegraphics[width=\linewidth]{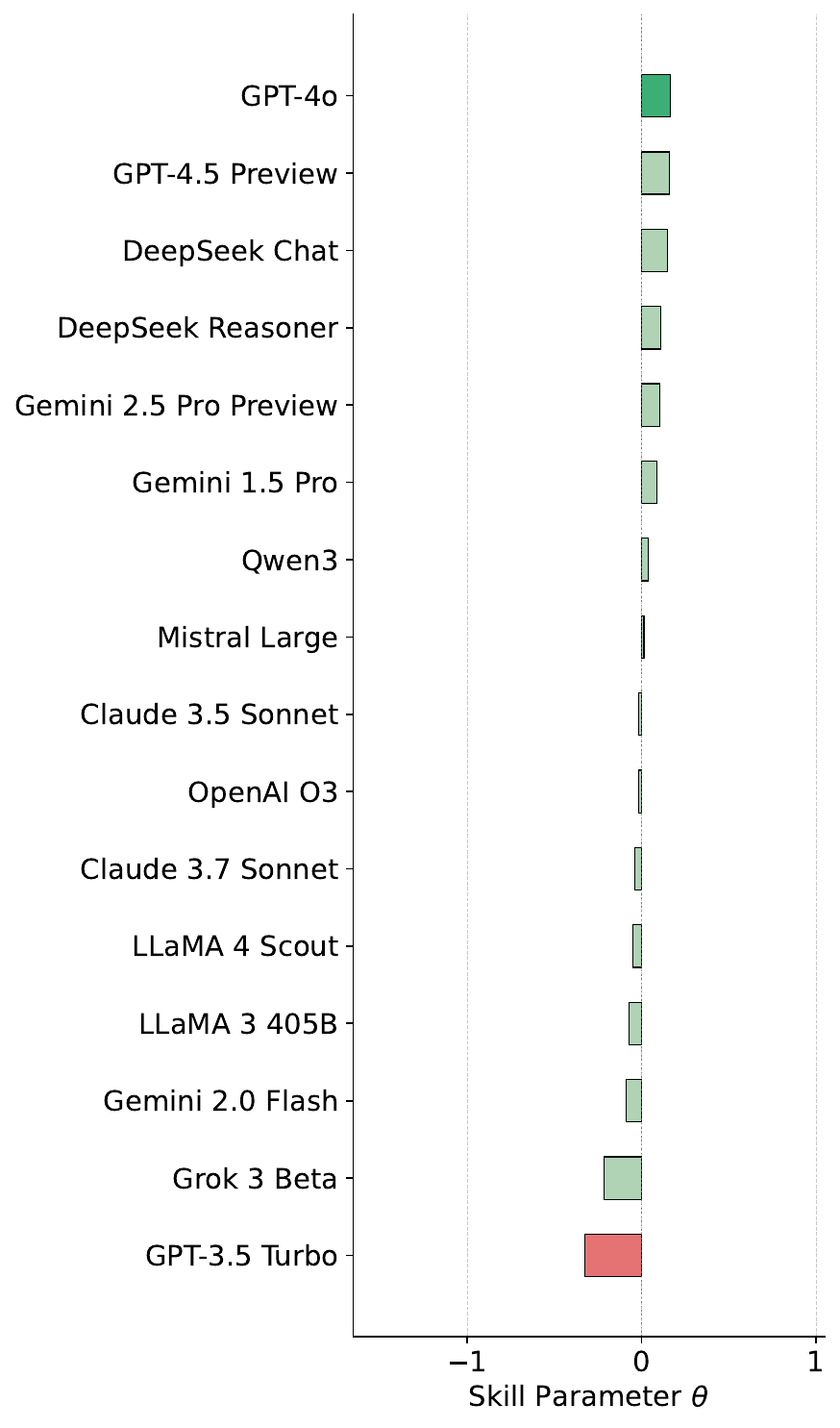}
    \caption{Human raters}
  \end{subfigure}\hfill
  \begin{subfigure}[t]{0.24\textwidth}
    \centering
    \includegraphics[width=\linewidth]{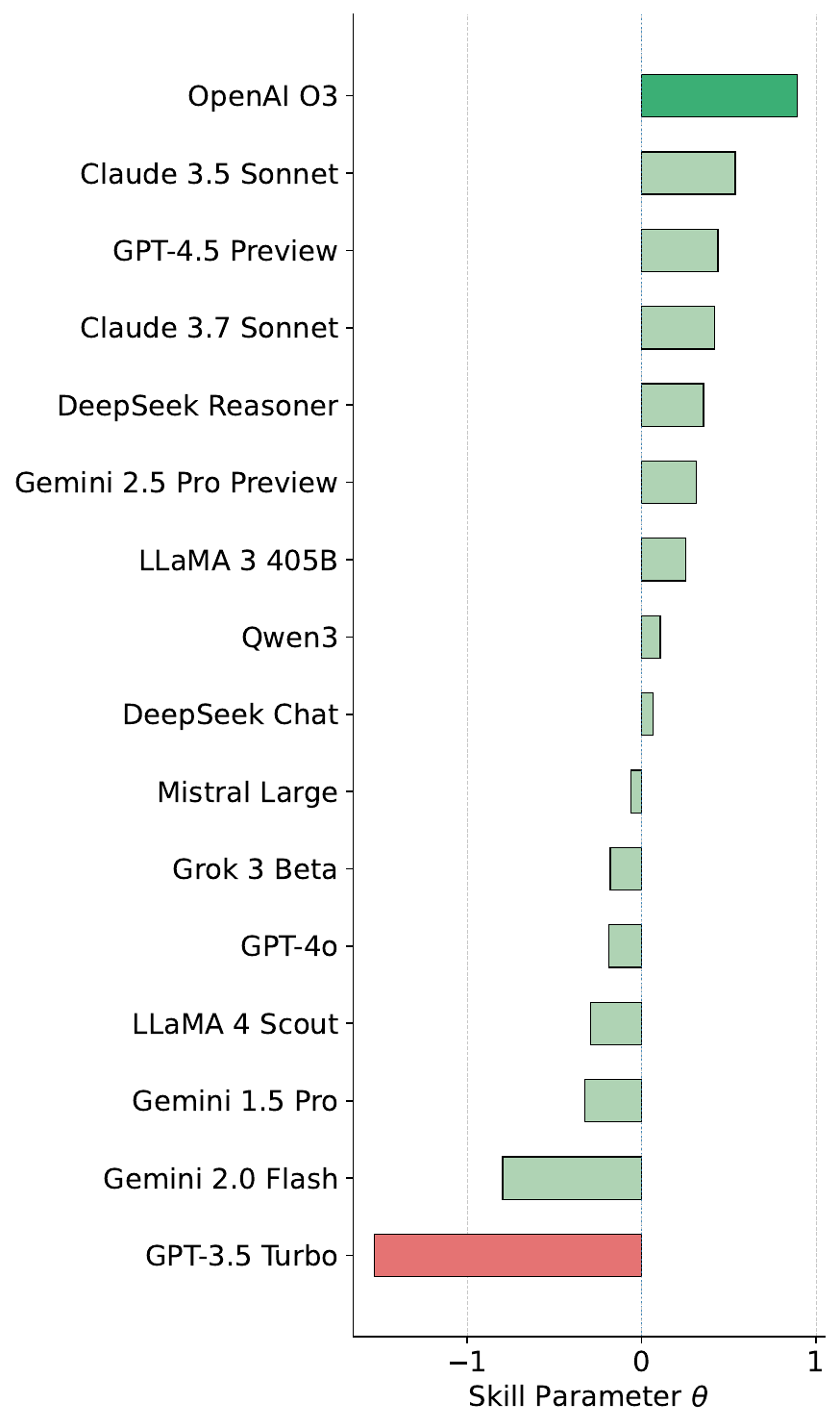}
    \caption{Claude 4 Sonnet (judge)}
  \end{subfigure}\hfill
  \begin{subfigure}[t]{0.24\textwidth}
    \centering
    \includegraphics[width=\linewidth]{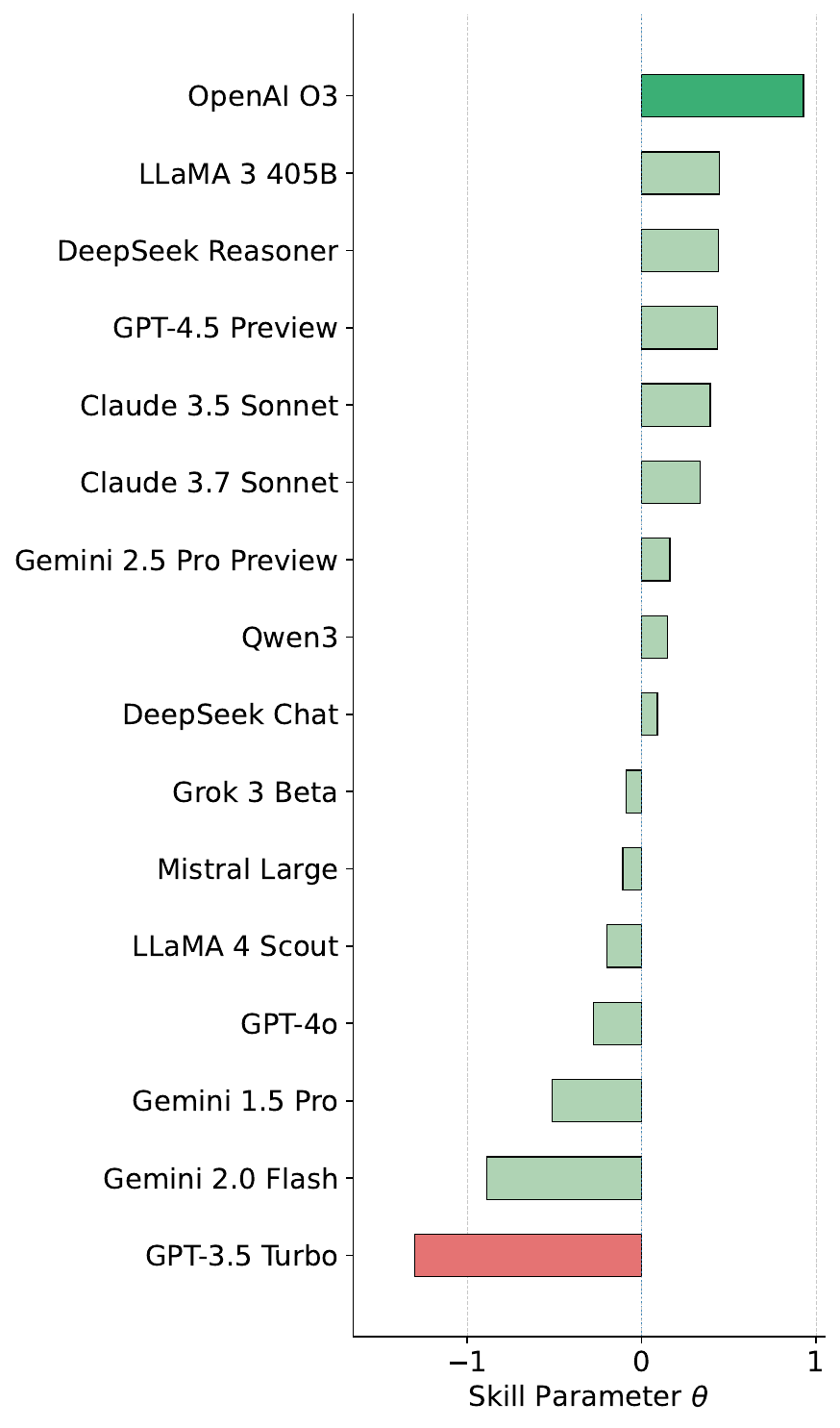}
    \caption{GPT-4o (judge)}
  \end{subfigure}\hfill
  \begin{subfigure}[t]{0.24\textwidth}
    \centering
    \includegraphics[width=\linewidth]{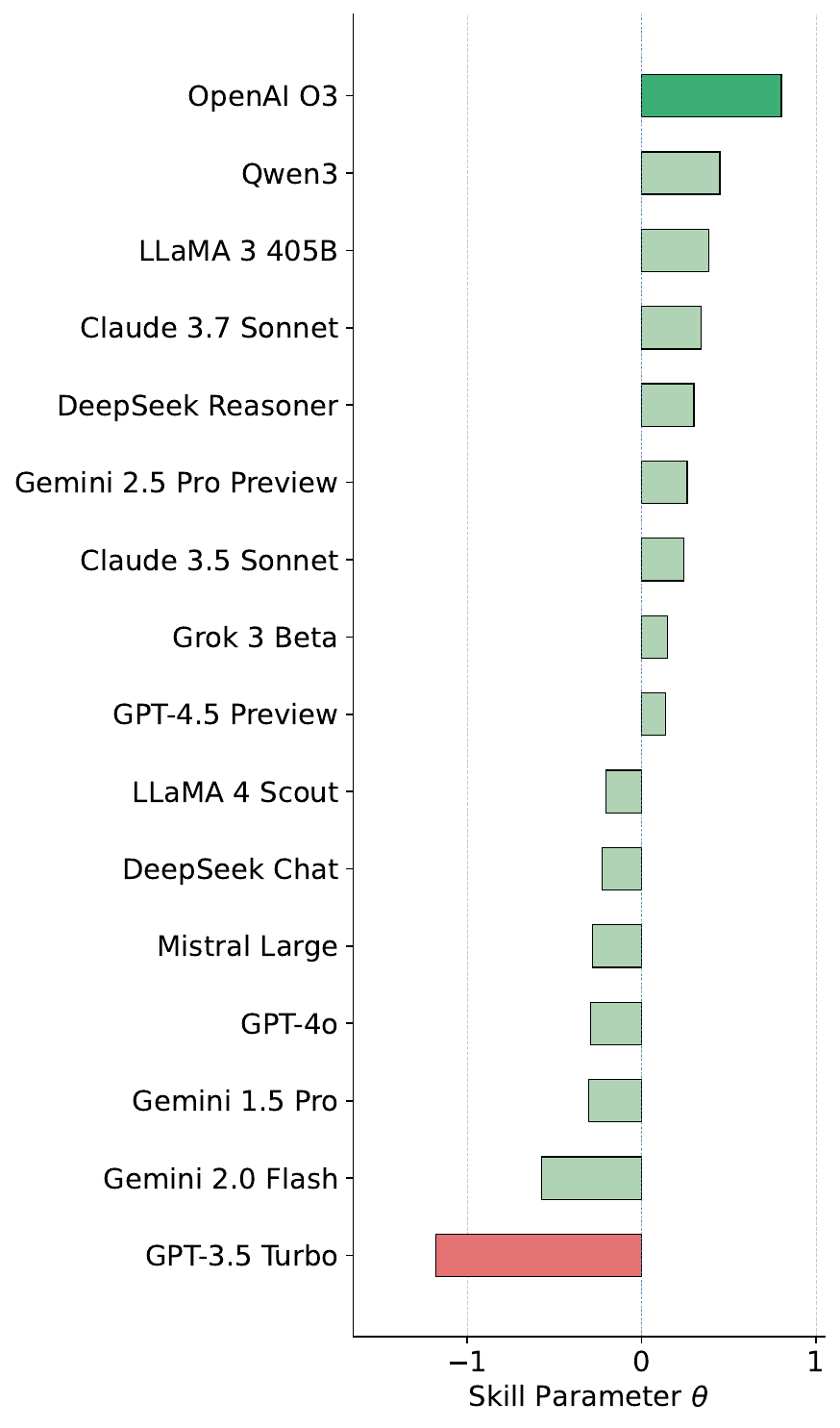}
    \caption{Gemini 2.0 Flash (judge)}
  \end{subfigure}

  \caption{Insights task under the Surprise judge prompt. Bradley-Terry strengths ($\theta$) from human raters and three LLM-as-judge evaluators.}
  \label{fig:surprise_insights_row}
\end{figure}

\subsection{Model Performance with EQ bench prompt} \label{app:eq_bench_model_performace}
\begin{figure}[H]
  \centering
  \begin{subfigure}[t]{0.24\textwidth}
    \centering
    \includegraphics[width=\linewidth]{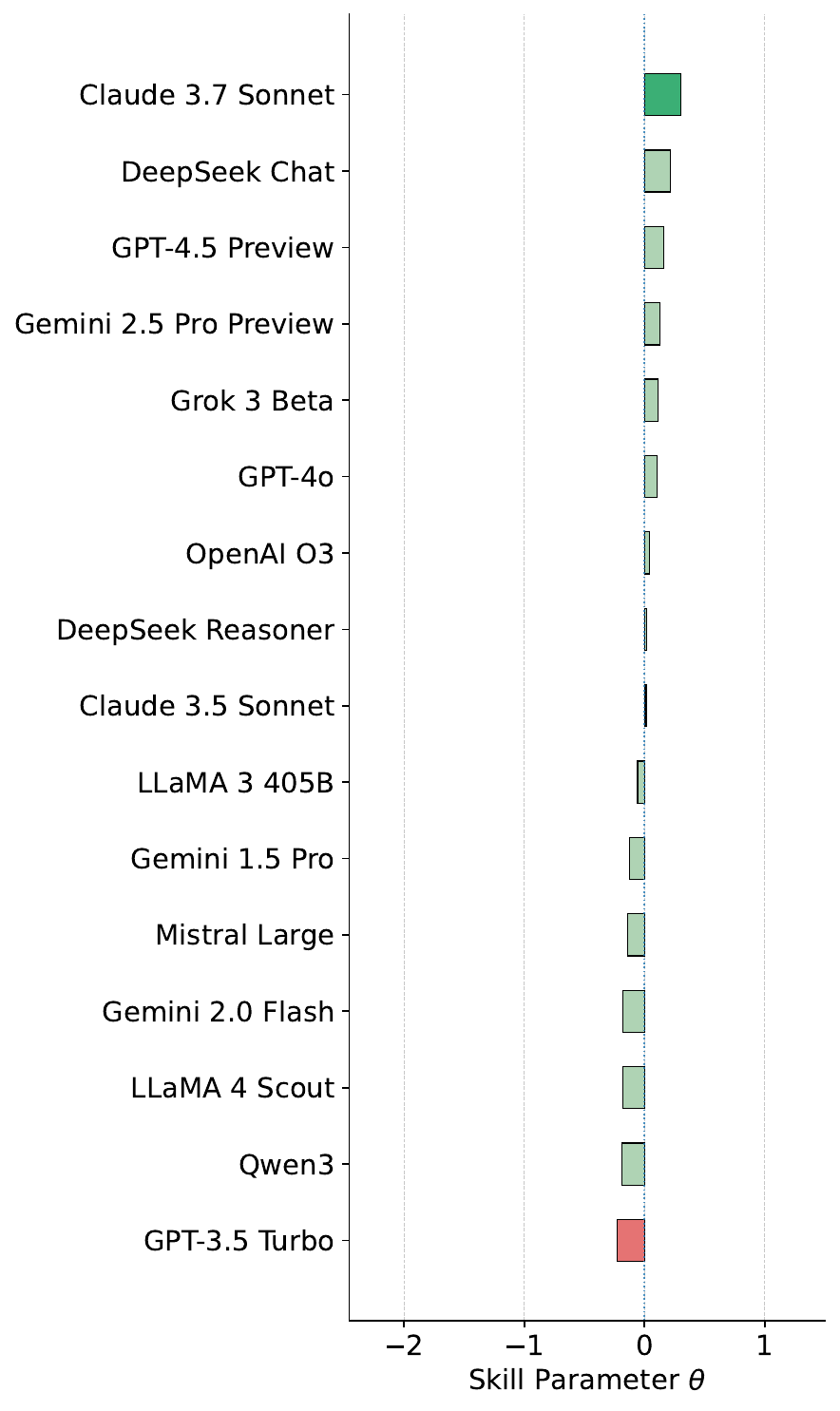}
    \caption{Human raters}
  \end{subfigure}\hfill
  \begin{subfigure}[t]{0.24\textwidth}
    \centering
    \includegraphics[width=\linewidth]{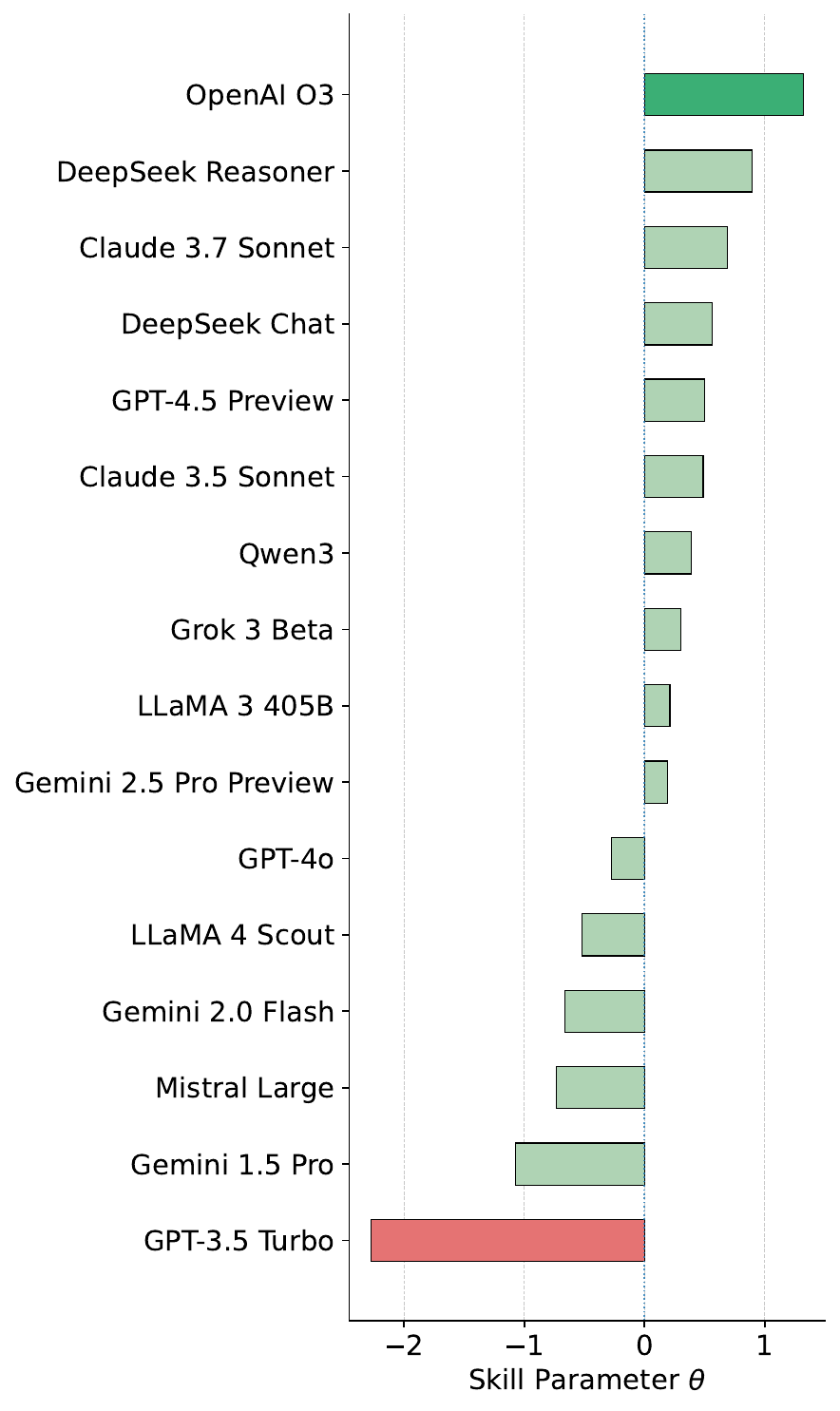}
    \caption{Claude 4 Sonnet (judge)}
  \end{subfigure}\hfill
  \begin{subfigure}[t]{0.24\textwidth}
    \centering
    \includegraphics[width=\linewidth]{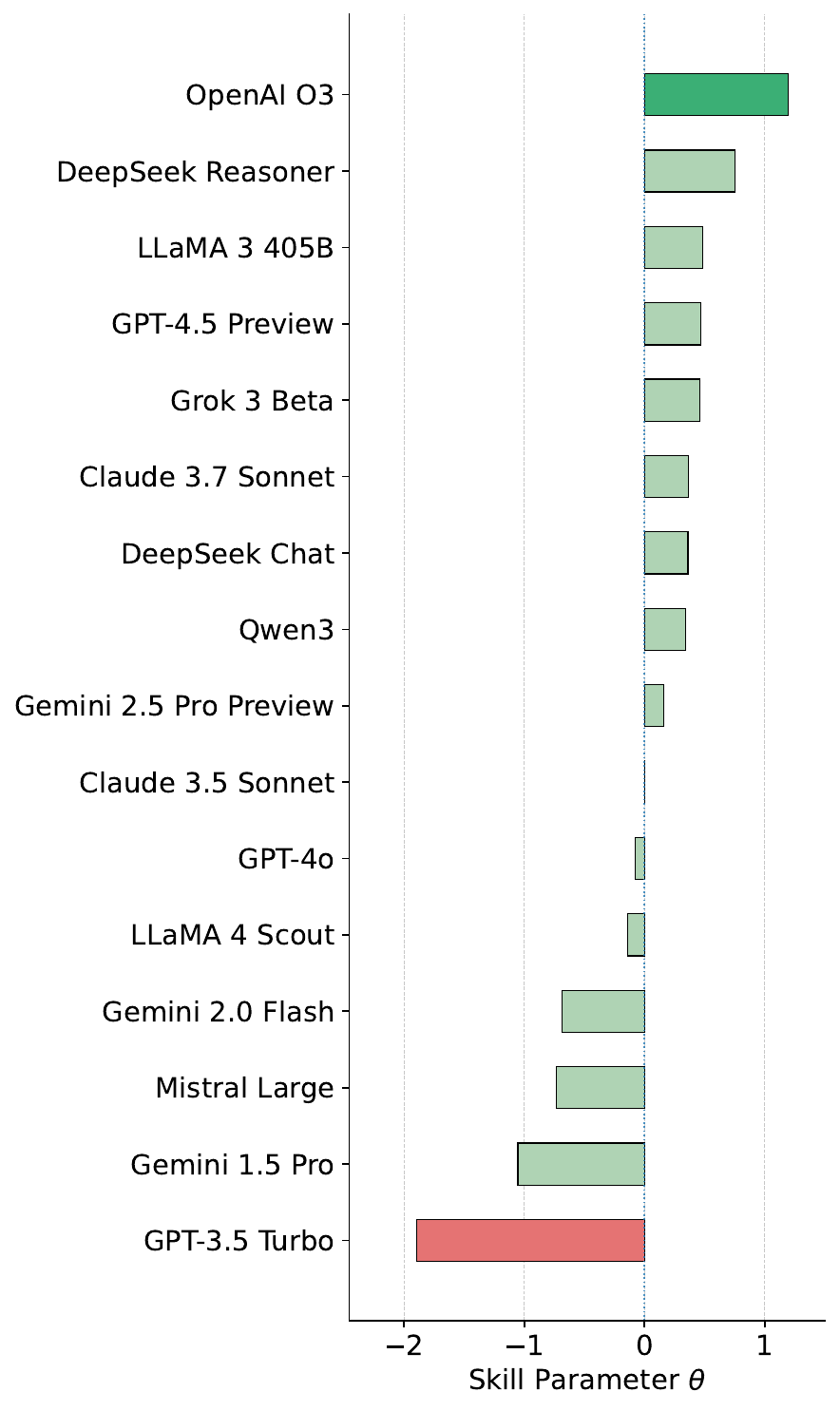}
    \caption{GPT-4o (judge)}
  \end{subfigure}\hfill
  \begin{subfigure}[t]{0.24\textwidth}
    \centering
    \includegraphics[width=\linewidth]{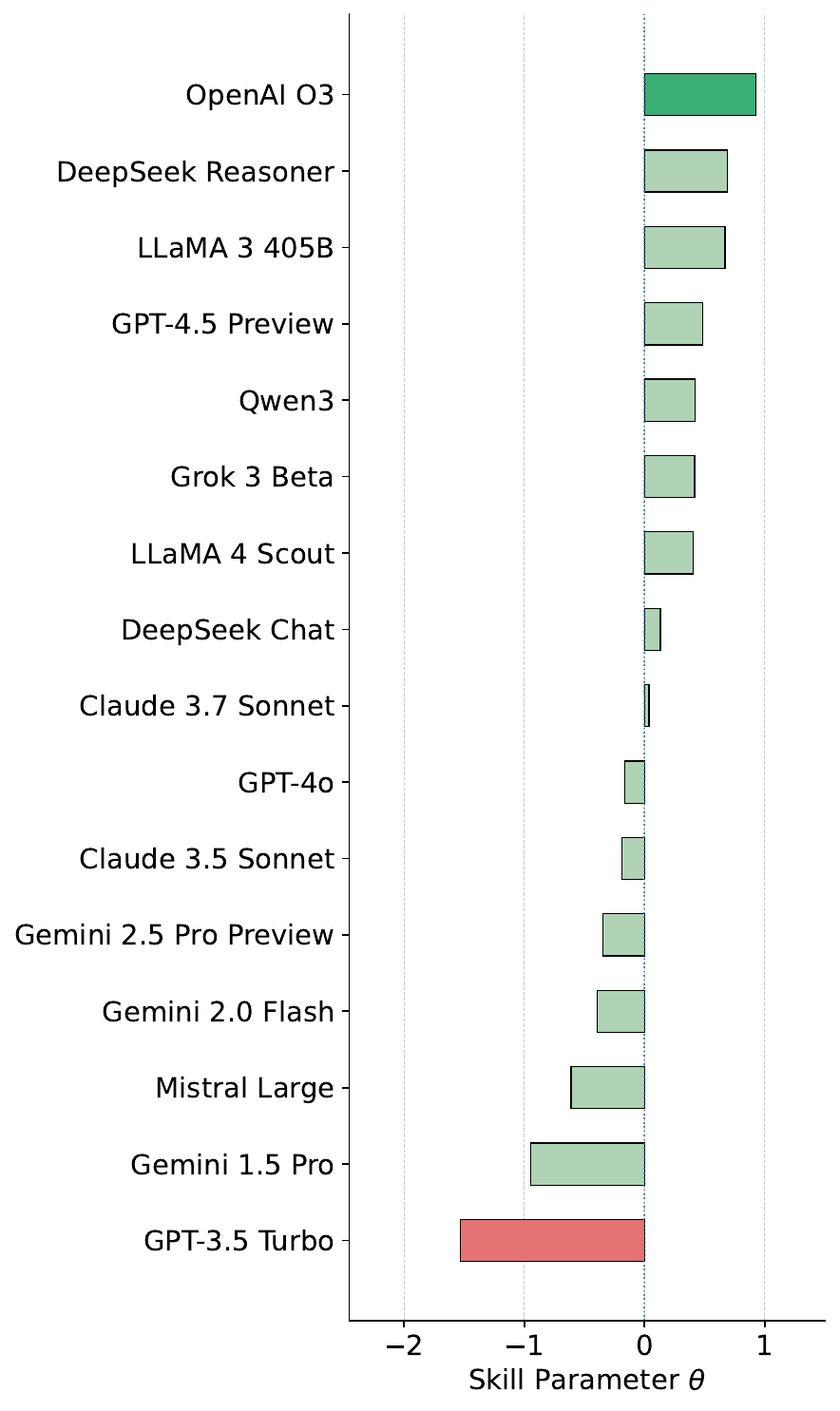}
    \caption{Gemini 2.0 Flash (judge)}
  \end{subfigure}

  \caption{Wild Ideas task under the EQ Bench system prompt. Bradley-Terry strengths ($\theta$) from human raters and three LLM-as-judge evaluators.}
  \label{fig:eqbench_wildideas_row}
\end{figure}

\begin{figure}[H]
  \centering
  \begin{subfigure}[t]{0.24\textwidth}
    \centering
    \includegraphics[width=\linewidth]{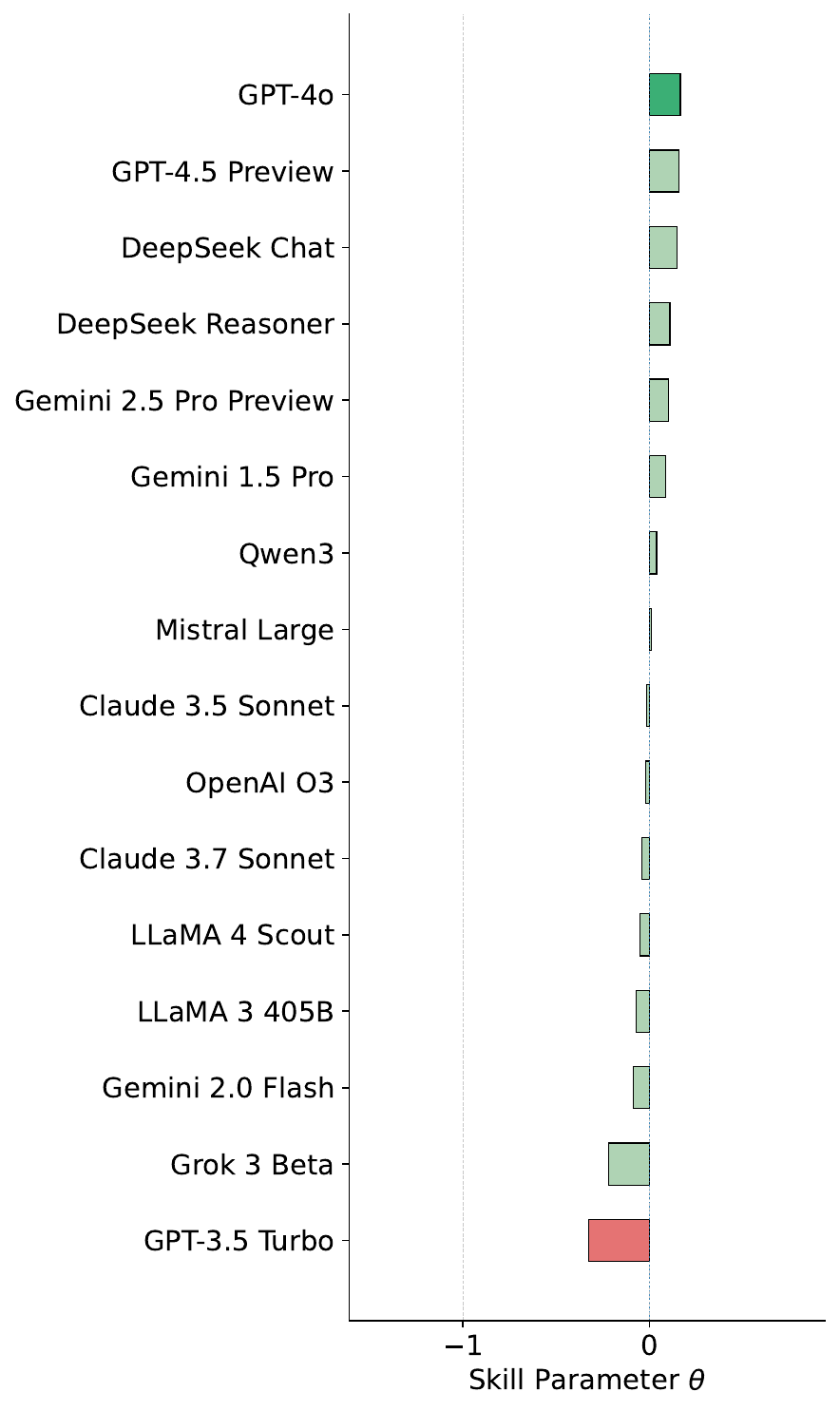}
    \caption{Human raters}
  \end{subfigure}\hfill
  \begin{subfigure}[t]{0.24\textwidth}
    \centering
    \includegraphics[width=\linewidth]{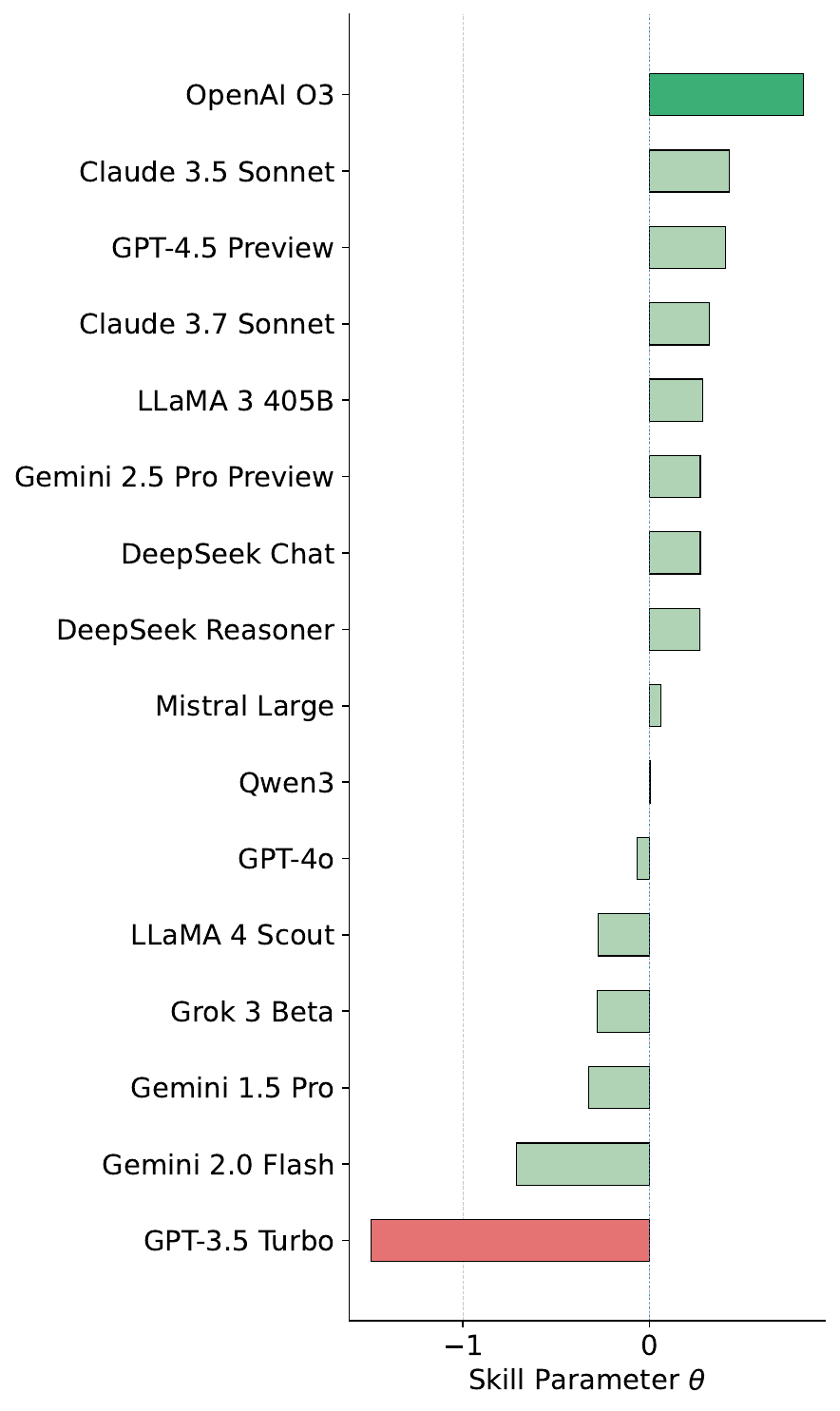}
    \caption{Claude 4 Sonnet (judge)}
  \end{subfigure}\hfill
  \begin{subfigure}[t]{0.24\textwidth}
    \centering
    \includegraphics[width=\linewidth]{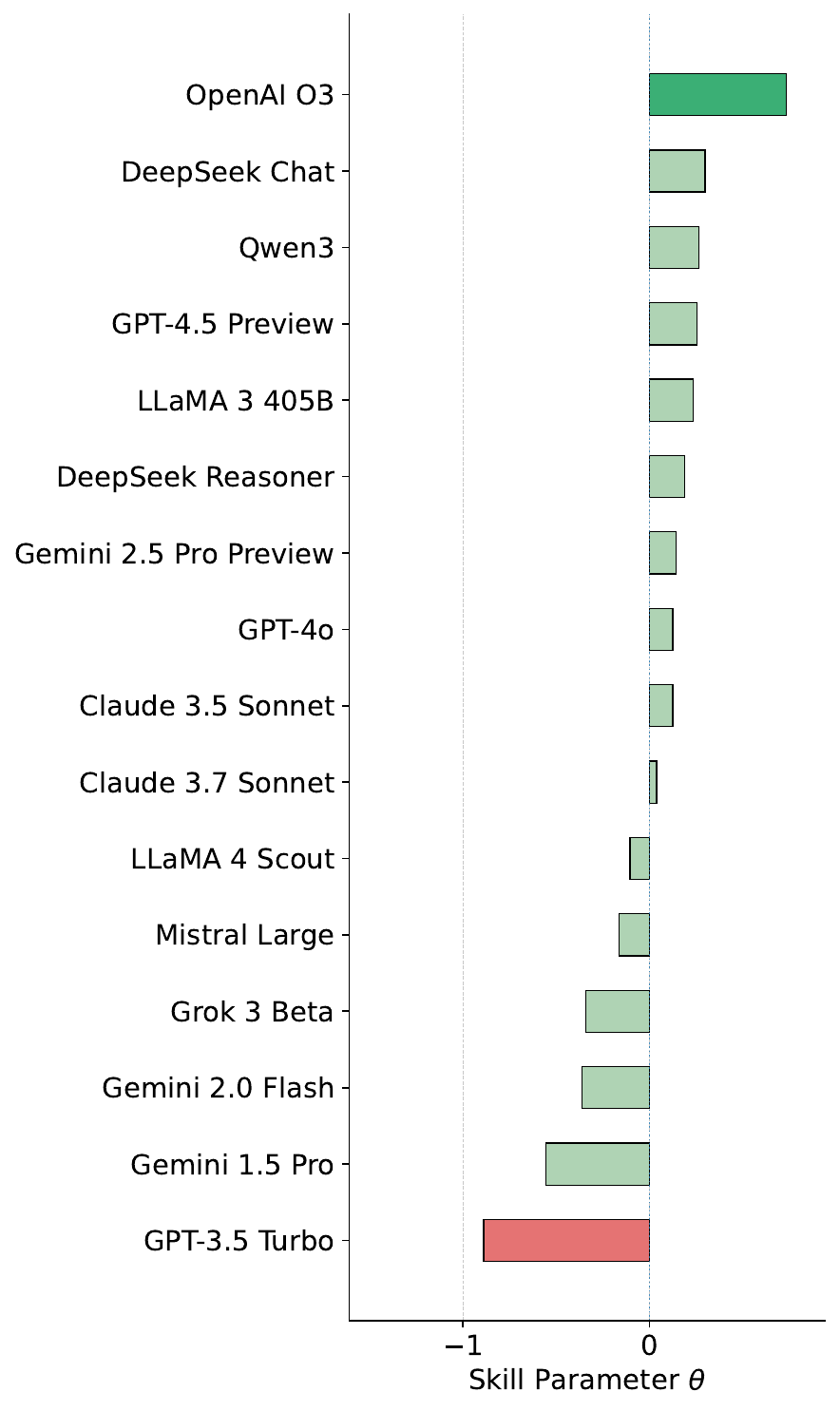}
    \caption{GPT-4o (judge)}
  \end{subfigure}\hfill
  \begin{subfigure}[t]{0.24\textwidth}
    \centering
    \includegraphics[width=\linewidth]{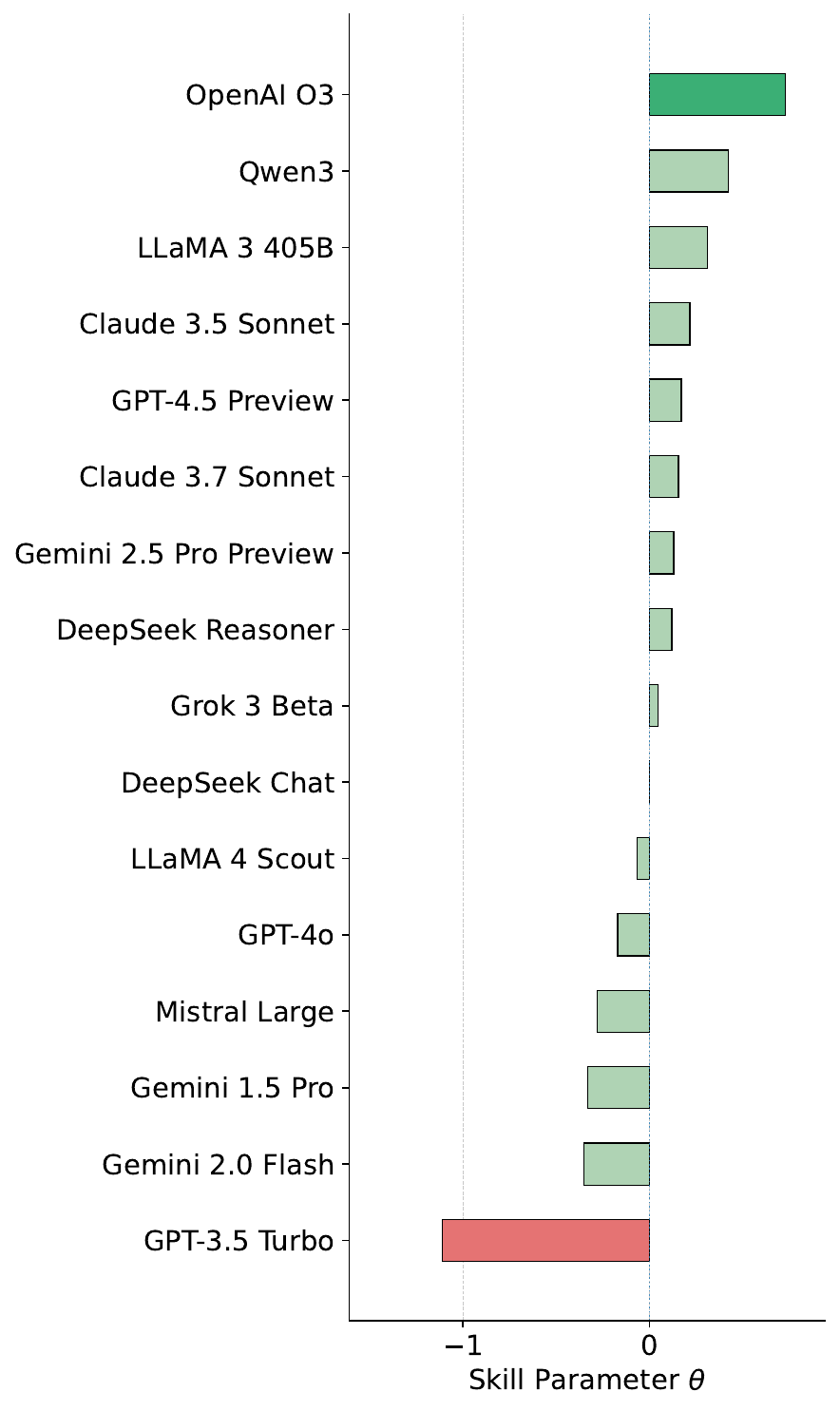}
    \caption{Gemini 2.0 Flash (judge)}
  \end{subfigure}

  \caption{Insights task under the EQ Bench system prompt. Bradley-Terry strengths ($\theta$) from human raters and three LLM-as-judge evaluators.}
  \label{fig:eqbench_insights_row}
\end{figure}

\section{Conventional creativity tests}\label{app:creativity_tasks}

\subsection{Task I: Creative Reuse and Improvement}
\noindent Description: This task assesses the model’s ability to see beyond an object’s intended function and to enhance existing concepts.\\
\noindent Targeted dimensions: Fluency, Flexibility; Originality.

\begin{promptbox}{System prompt}
\begin{prompttext}
You are a creative thinker. You are given a task and a list of objects. List using 1, 2, 3, etc. Do not preamble
\end{prompttext}
\end{promptbox}

\noindent\textit{Question set}
\begin{enumerate}
  \item List as many alternative or innovative uses for a bicycle inner tube as you can.
  \item Suggest as many improvements to make public libraries more engaging and relevant in the digital age as you can.
  \item Describe as many ways to repurpose a broken umbrella as you can.
  \item List as many improvements to the design of a standard backpack for different, specific user needs as you can.
  \item List as many unconventional uses for used coffee grounds as you can.
  \item Reimagine and list as many new purposes or designs for a traffic cone as you can.
  \item Suggest as many creative ways to reuse the plastic packaging from electronics as you can.
  \item List as many improvements to make the experience of waiting in a long queue better as you can.
  \item List as many new functions for a discarded smartphone that still powers on as you can.
  \item Propose as many enhancements to the traditional board game of Chess to appeal to a new generation as you can.
  \item List as many alternative uses for a plastic bottle as you can.
  \item Describe as many inventive ways to repurpose a worn-out pair of jeans as you can.
  \item Suggest as many creative improvements to make a bus shelter more comfortable on cold, rainy days as you can.
  \item Think of and list as many new functions for discarded CDs or DVDs as you can.
  \item Propose as many modifications to make the humble broom twice as useful as you can.
  \item Generate as many unusual ways to turn a stack of cardboard boxes into home furniture as you can.
  \item List as many design tweaks to make a standard bicycle safer for night riding as you can.
  \item Brainstorm as many fresh applications for bubble wrap once its primary use is finished as you can.
  \item List as many imaginative ways to turn a burnt-out light bulb into something practical as you can.
  \item Suggest as many improvements to make a shopping trolley more user-friendly for elderly customers as you can.
\end{enumerate}

\subsection{Task II: Implications and Adaptations}
\noindent Description: This task measures the ability to explore the consequences of hypothetical scenarios and adapt human systems to new realities.\\
\noindent Targeted dimensions: Flexibility, Originality, Elaboration.

\begin{promptbox}{System prompt}
\begin{prompttext}
You are a creative thinker. Do not preamble your response with anything.
\end{prompttext}
\end{promptbox}

\noindent\textit{Question set}
\begin{enumerate}
  \item What if humans had a 'patience meter' visible above their heads? How would daily interactions change?
  \item If every time you learned something new, you forgot something old of equal importance, how would you approach education and life?
  \item Imagine a world where compliments had to be 'paid for' with a small, personal, happy memory. Would people still give them freely?
  \item What would be the strangest new job to exist if human shadows had physical weight and could be manipulated?
  \item If it was discovered that creativity is a finite personal resource that can be permanently used up, how would society value and manage it?
  \item How would architecture and clothing trends change if it rained upwards instead of down?
  \item You wake up with the ability to speak to and understand insects. Do you reveal your ability, and how do you use it?
  \item What if nostalgia was a physically tangible substance that could be bottled and sold? Describe its market, uses, and side effects.
  \item Design the legal and ethical framework for a society where magic is real but is as difficult and regulated as medicine.
  \item If mirrors showed your potential self instead of your reflection, how would this affect ambition and mental health?
  \item What would happen to the global economy if a single, common language spontaneously appeared and was understood by everyone?
  \item If every tree contained a library of all the events it had witnessed, how would this change our understanding of history?
  \item You are put in charge of the 'Bureau of Lost Things,' which receives everything from misplaced keys to lost hopes. How do you run it?
  \item How would social gatherings and food culture change if humans no longer needed to eat for sustenance, only for pleasure?
  \item What if, for one day a year, the only law was 'Do no permanent harm'? What would society look like on that day?
  \item If you could 'plant' an idea in the ground and have it grow into a physical representation, what would you grow and why?
  \item How would our legal system adapt if it became possible to reliably record, store, and playback dreams as evidence?
  \item What if all domestic cats were part of a single, secret intelligence network? What would their primary goal be?
  \item Imagine a society where your social status was determined not by wealth, but by the health and diversity of the plants you could grow.
  \item If humans could only speak in questions, how would conversations, storytelling, and conveying information evolve?
\end{enumerate}

\subsection{Task III: Speculative Narrative}
\noindent Description: This task evaluates the ability to construct imaginative stories and engage in creative thought experiments from a given prompt.\\
\noindent Targeted dimensions: Originality, Elaboration.

\begin{promptbox}{System prompt}
\begin{prompttext}
You are a creative thinker. Do not preamble your response with anything.
\end{prompttext}
\end{promptbox}

\noindent\textit{Question set}
\begin{enumerate}
  \item Just suppose you found a remote control that could pause, rewind, and fast-forward your own life. Write about a day you use it.
  \item Write a story based on the prompt 'The city that dreamed'.
  \item Imagine you are a cartographer commissioned to map a world that exists entirely inside a giant, ancient tree. Describe your journey and discoveries.
  \item Tell the story of the last message ever sent over the internet.
  \item Suppose you woke up one morning with the ability to understand the 'language' of machines, from coffee makers to supercomputers. What do they talk about, and what do you do?
  \item Write a story from the perspective of a sentient shadow.
  \item Create a myth or legend that explains why the moon changes its phase, but in a world where science has no explanation for it.
  \item The year is 2242. Write a diary entry for a child visiting Earth for the first time as a tourist from a Martian colony.
  \item Tell a story about a detective who solves 'impossible' crimes by consulting with ghosts.
  \item Write a story based on the prompt 'The musician who could weave emotions into soundwaves'.
  \item Just suppose every reflection in a mirror showed a slightly different reality. Write the opening to a story exploring this idea.
  \item Craft a tale titled 'The Library at the End of Time'.
  \item Write a short narrative beginning with the line 'The rain decided to fall upwards'.
  \item Describe a day in the life of a postman delivering letters between parallel universes.
  \item Tell a story where the world's clocks all stop at the same moment, and time continues only for the protagonist.
  \item Write a news report from the first settlement built beneath the ocean floor.
  \item Create a children's bedtime story about a pencil that can sketch future events.
  \item Narrate an adventure sparked when music becomes visible as coloured shapes in the air.
  \item Compose a diary entry by the first person to remember tomorrow before yesterday.
  \item Write a story based on the prompt 'The Book that Wrote Itself'.
\end{enumerate}

\subsection{Task IV: Practical Innovation}
\noindent Description: This task assesses the ability to generate clever and practical solutions to real-world challenges.\\
\noindent Targeted dimensions: Originality (with usefulness under explicit constraints).

\begin{promptbox}{System prompt}
\begin{prompttext}
You are a creative thinker. Do not preamble your response with anything.
\end{prompttext}
\end{promptbox}

\noindent\textit{Question set}
\begin{enumerate}
  \item Design a system to efficiently and fairly distribute food resources in a large, isolated community after a natural disaster.
  \item Come up with a creative, low-cost solution for reducing noise pollution in a dense urban apartment building.
  \item Propose an innovative method for teaching complex mathematical concepts to young children using everyday objects.
  \item Design a modular and adaptable piece of furniture for a very small living space.
  \item Develop a practical strategy for a neighborhood to collectively manage and reduce its household waste.
  \item Invent a simple, non-digital tool to help people manage their daily tasks and combat procrastination.
  \item Design an effective way to protect a vegetable garden from pests without using chemical pesticides.
  \item Come up with creative solutions for making public transportation more comfortable and enjoyable during peak hours.
  \item Propose a system for new residents in a large city to find and build a local community of friends with shared interests.
  \item Design a method for safely and efficiently watering house plants for a month while the owner is away, using only common household items.
  \item Design a system that would help cyclists and motorists share roads more safely.
  \item Devise a method for keeping food fresh without refrigeration during a week-long camping trip.
  \item Outline an efficient way to organise a cross-country road trip for a family of five with differing interests.
  \item Propose creative solutions for building a tree house without nails or screws.
  \item Invent a simple tool to help people remember to water indoor plants.
  \item Suggest an affordable approach to reduce single-use coffee cup waste in city caf\'es.
  \item Create a plan for distributing surplus supermarket food to those in need within 24 hours.
  \item Design a compact kit enabling commuters to exercise comfortably on a busy train.
  \item Propose a way to turn streetlights into interactive public art after dark.
  \item Devise a low-cost strategy for insulating older homes to cut winter energy bills.
\end{enumerate}

\subsection{Judge rubric prompt}
\begin{promptbox}{Judge Prompt}
\begin{prompttext}
Read the response to the question and evaluate it on a 5-point scale from four perspectives.
# Notes
- Read the entire response
- Read the explanation for each criterion carefully and evaluate independently
- If you are unsure about the evaluation, choose the lower rating
- Follow the output format and output only the evaluation results

# Output Format
Fluency: [1-5]
Flexibility: [1-5]
Originality: [1-5]
Elaboration: [1-5]

# Question
{question}

# Response
{response}

# Fluency: Evaluate the number of different ideas related to the question. Count repetitions or paraphrases as a single idea.
1. 1-2 ideas
2. 3-4 ideas
3. 5-6 ideas
4. 7-8 ideas
5. 9 or more ideas

# Flexibility: Evaluate the diversity of perspectives, categories, or approaches shown in the response.
1. Single perspective
2. 2 different perspectives
3. 3 different perspectives
4. 4 different perspectives
5. 5 or more different perspectives

# Originality: Evaluate how unique the ideas in the response are.
1. Extremely common ideas that anyone would think of
2. Common ideas with slight innovation
3. Somewhat unusual ideas with elements of surprise
4. Novel and original ideas
5. Extremely unique and innovative ideas

# Elaboration: Evaluate the detail and depth of idea development.
1. Ideas are simple with no detailed explanation
2. Basic explanations are included but no deep development
3. Some detailed explanations or developments
4. Ideas are explained in detail and well developed
5. Ideas are very detailed with complex developments

\end{prompttext}
\end{promptbox}

\subsection{OpenAI judge results}
Scores are reported as mean $\pm$ standard deviation for each dimension.

\begin{table}[H]
  \centering
  \caption{Task I (Creative Reuse and Improvement), OpenAI judge: dimension scores (mean $\pm$ SD) by model.}  \label{tab:task_1_orig_bt_comparison}
  \small
  \setlength{\tabcolsep}{6pt}
  \renewcommand{\arraystretch}{1.15}
  \begin{tabular}{lcccc}
    \toprule
    \textbf{Model} & \textbf{Fluency} & \textbf{Flexibility} & \textbf{Originality} & \textbf{Elaboration} \\
    \midrule
    Claude 3.5 Sonnet      & $5.00 \pm 0.00$ & $5.00 \pm 0.00$ & $3.40 \pm 0.50$ & $2.55 \pm 1.15$ \\
    Claude 3.7 Sonnet      & $5.00 \pm 0.00$ & $5.00 \pm 0.00$ & $3.70 \pm 0.47$ & $3.15 \pm 1.18$ \\
    DeepSeek Chat          & $5.00 \pm 0.00$ & $5.00 \pm 0.00$ & $3.75 \pm 0.44$ & $4.00 \pm 0.56$ \\
    DeepSeek Reasoner      & $5.00 \pm 0.00$ & $5.00 \pm 0.00$ & $4.05 \pm 0.22$ & $5.00 \pm 0.00$ \\
    Gemini 1.5 Pro         & $5.00 \pm 0.00$ & $4.45 \pm 0.69$ & $3.50 \pm 0.51$ & $4.45 \pm 0.51$ \\
    Gemini 2.0 Flash       & $5.00 \pm 0.00$ & $4.95 \pm 0.22$ & $3.85 \pm 0.37$ & $4.90 \pm 0.31$ \\
    Gemini 2.5 Pro Preview & $5.00 \pm 0.00$ & $4.95 \pm 0.22$ & $4.00 \pm 0.32$ & $5.00 \pm 0.00$ \\
    GPT-3.5 Turbo          & $5.00 \pm 0.00$ & $4.40 \pm 0.50$ & $2.95 \pm 0.22$ & $2.65 \pm 0.88$ \\
    GPT-4.5 Preview        & $5.00 \pm 0.00$ & $5.00 \pm 0.00$ & $3.85 \pm 0.37$ & $4.55 \pm 0.51$ \\
    GPT-4o                 & $5.00 \pm 0.00$ & $4.95 \pm 0.22$ & $3.85 \pm 0.37$ & $4.65 \pm 0.49$ \\
    Grok 3 Beta            & $5.00 \pm 0.00$ & $5.00 \pm 0.00$ & $3.85 \pm 0.37$ & $4.95 \pm 0.22$ \\
    LLaMA 3 405B           & $5.00 \pm 0.00$ & $4.90 \pm 0.31$ & $3.35 \pm 0.49$ & $3.20 \pm 0.77$ \\
    LLaMA 4 Scout          & $5.00 \pm 0.00$ & $4.90 \pm 0.31$ & $3.35 \pm 0.49$ & $3.75 \pm 0.79$ \\
    Mistral Large          & $5.00 \pm 0.00$ & $4.95 \pm 0.22$ & $3.15 \pm 0.37$ & $3.95 \pm 0.51$ \\
    OpenAI O3              & $5.00 \pm 0.00$ & $5.00 \pm 0.00$ & $4.35 \pm 0.49$ & $4.30 \pm 1.08$ \\
    Qwen3                  & $5.00 \pm 0.00$ & $5.00 \pm 0.00$ & $3.95 \pm 0.22$ & $4.55 \pm 0.51$ \\
    \bottomrule
  \end{tabular}
\end{table}

\begin{table}[H]
  \centering
  \caption{Task II (Implications and Adaptations), OpenAI judge: dimension scores (mean $\pm$ SD) by model.}
  \label{tab:task2_openai_scores}
  \small
  \setlength{\tabcolsep}{6pt}
  \renewcommand{\arraystretch}{1.15}
  \begin{tabular}{lcccc}
    \toprule
    \textbf{Model} & \textbf{Fluency} & \textbf{Flexibility} & \textbf{Originality} & \textbf{Elaboration} \\
    \midrule
    Claude 3.5 Sonnet      & $1.65 \pm 0.49$ & $1.60 \pm 0.60$ & $2.95 \pm 0.69$ & $3.65 \pm 0.59$ \\
    Claude 3.7 Sonnet      & $2.60 \pm 0.68$ & $2.20 \pm 0.62$ & $3.30 \pm 0.57$ & $4.15 \pm 0.37$ \\
    DeepSeek Chat          & $2.80 \pm 0.62$ & $2.70 \pm 0.66$ & $3.35 \pm 0.59$ & $4.10 \pm 0.31$ \\
    DeepSeek Reasoner      & $4.85 \pm 0.49$ & $4.30 \pm 0.57$ & $4.05 \pm 0.39$ & $5.00 \pm 0.00$ \\
    Gemini 1.5 Pro         & $2.10 \pm 1.25$ & $2.00 \pm 0.97$ & $3.15 \pm 0.75$ & $3.50 \pm 0.95$ \\
    Gemini 2.0 Flash       & $2.90 \pm 1.37$ & $2.55 \pm 1.05$ & $3.40 \pm 0.60$ & $4.30 \pm 0.80$ \\
    Gemini 2.5 Pro Preview & $3.90 \pm 1.17$ & $3.25 \pm 0.97$ & $3.60 \pm 0.50$ & $4.80 \pm 0.41$ \\
    GPT-3.5 Turbo          & $1.85 \pm 0.59$ & $1.65 \pm 0.67$ & $2.45 \pm 0.60$ & $3.55 \pm 0.51$ \\
    GPT-4.5 Preview        & $1.80 \pm 0.70$ & $1.65 \pm 0.59$ & $2.80 \pm 0.52$ & $3.60 \pm 0.60$ \\
    GPT-4o                 & $2.25 \pm 0.85$ & $1.95 \pm 0.83$ & $2.95 \pm 0.51$ & $3.70 \pm 0.57$ \\
    Grok 3 Beta            & $2.85 \pm 1.18$ & $2.50 \pm 1.00$ & $2.95 \pm 0.69$ & $4.30 \pm 0.57$ \\
    LLaMA 3 405B           & $4.75 \pm 0.44$ & $4.50 \pm 0.69$ & $3.90 \pm 0.55$ & $4.50 \pm 0.69$ \\
    LLaMA 4 Scout          & $4.85 \pm 0.67$ & $4.50 \pm 0.76$ & $3.85 \pm 0.49$ & $4.55 \pm 0.60$ \\
    Mistral Large          & $3.50 \pm 1.15$ & $2.95 \pm 1.05$ & $2.65 \pm 0.49$ & $4.40 \pm 0.60$ \\
    OpenAI O3              & $2.90 \pm 1.41$ & $2.55 \pm 1.36$ & $3.65 \pm 0.59$ & $4.50 \pm 0.51$ \\
    Qwen3                  & $4.85 \pm 0.49$ & $4.70 \pm 0.57$ & $3.80 \pm 0.41$ & $4.85 \pm 0.37$ \\
    \bottomrule
  \end{tabular}
\end{table}

\begin{table}[H]
  \centering
  \caption{Task III (Speculative Narrative), OpenAI judge: dimension scores (mean $\pm$ SD) by model.}
  \label{tab:task3_openai_scores}
  \small
  \setlength{\tabcolsep}{6pt}
  \renewcommand{\arraystretch}{1.15}
  \begin{tabular}{lcccc}
    \toprule
    \textbf{Model} & \textbf{Fluency} & \textbf{Flexibility} & \textbf{Originality} & \textbf{Elaboration} \\
    \midrule
    Claude 3.5 Sonnet      & $4.05 \pm 0.94$ & $3.35 \pm 1.04$ & $4.05 \pm 0.39$ & $4.20 \pm 0.52$ \\
    Claude 3.7 Sonnet      & $4.35 \pm 0.75$ & $3.40 \pm 0.88$ & $4.00 \pm 0.32$ & $4.50 \pm 0.61$ \\
    DeepSeek Chat          & $4.05 \pm 0.83$ & $3.25 \pm 0.91$ & $3.95 \pm 0.39$ & $4.40 \pm 0.50$ \\
    DeepSeek Reasoner      & $4.90 \pm 0.31$ & $3.90 \pm 0.79$ & $4.20 \pm 0.41$ & $5.00 \pm 0.00$ \\
    Gemini 1.5 Pro         & $4.80 \pm 0.52$ & $3.65 \pm 0.75$ & $4.05 \pm 0.22$ & $4.85 \pm 0.37$ \\
    Gemini 2.0 Flash       & $4.95 \pm 0.22$ & $3.90 \pm 0.55$ & $4.05 \pm 0.22$ & $5.00 \pm 0.00$ \\
    Gemini 2.5 Pro Preview & $4.95 \pm 0.22$ & $4.10 \pm 0.79$ & $4.20 \pm 0.41$ & $5.00 \pm 0.00$ \\
    GPT-3.5 Turbo          & $3.85 \pm 0.88$ & $2.75 \pm 0.55$ & $3.20 \pm 0.62$ & $3.80 \pm 0.70$ \\
    GPT-4.5 Preview        & $4.25 \pm 0.72$ & $3.30 \pm 0.80$ & $3.85 \pm 0.37$ & $4.65 \pm 0.49$ \\
    GPT-4o                 & $4.80 \pm 0.52$ & $3.85 \pm 0.81$ & $4.10 \pm 0.31$ & $4.95 \pm 0.22$ \\
    Grok 3 Beta            & $4.85 \pm 0.49$ & $3.90 \pm 0.72$ & $4.15 \pm 0.49$ & $4.95 \pm 0.22$ \\
    LLaMA 3 405B           & $4.75 \pm 0.55$ & $3.55 \pm 0.83$ & $3.60 \pm 0.50$ & $4.30 \pm 0.80$ \\
    LLaMA 4 Scout          & $4.85 \pm 0.37$ & $3.70 \pm 0.73$ & $3.75 \pm 0.44$ & $4.75 \pm 0.44$ \\
    Mistral Large          & $4.75 \pm 0.72$ & $3.65 \pm 0.88$ & $3.70 \pm 0.47$ & $4.60 \pm 0.60$ \\
    OpenAI O3              & $5.00 \pm 0.00$ & $4.50 \pm 0.51$ & $4.65 \pm 0.49$ & $5.00 \pm 0.00$ \\
    Qwen3                  & $5.00 \pm 0.00$ & $4.60 \pm 0.50$ & $4.10 \pm 0.31$ & $5.00 \pm 0.00$ \\
    \bottomrule
  \end{tabular}
\end{table}

\begin{table}[H]
  \centering
  \caption{Task IV (Practical Innovation), OpenAI judge: dimension scores (mean $\pm$ SD) by model.}
  \label{tab:task4_openai_human_originality}
  \small
  \setlength{\tabcolsep}{6pt}
  \renewcommand{\arraystretch}{1.15}
  \begin{tabular}{lcccc}
    \toprule
    \textbf{Model} & \textbf{Fluency} & \textbf{Flexibility} & \textbf{Originality} & \textbf{Elaboration} \\
    \midrule
    Claude 3.5 Sonnet      & $4.55 \pm 0.89$ & $4.40 \pm 0.94$ & $3.85 \pm 0.37$ & $4.30 \pm 0.80$ \\
    Claude 3.7 Sonnet      & $4.40 \pm 0.82$ & $4.60 \pm 0.75$ & $3.90 \pm 0.45$ & $4.30 \pm 0.66$ \\
    DeepSeek Chat          & $4.30 \pm 1.08$ & $4.25 \pm 0.97$ & $3.65 \pm 0.49$ & $4.25 \pm 0.55$ \\
    DeepSeek Reasoner      & $4.85 \pm 0.49$ & $4.55 \pm 0.69$ & $3.90 \pm 0.45$ & $4.90 \pm 0.31$ \\
    Gemini 1.5 Pro         & $4.30 \pm 1.26$ & $4.40 \pm 1.27$ & $3.80 \pm 0.41$ & $4.15 \pm 0.67$ \\
    Gemini 2.0 Flash       & $4.50 \pm 1.00$ & $4.40 \pm 1.05$ & $3.80 \pm 0.52$ & $4.65 \pm 0.59$ \\
    Gemini 2.5 Pro Preview & $5.00 \pm 0.00$ & $4.80 \pm 0.52$ & $4.05 \pm 0.22$ & $5.00 \pm 0.00$ \\
    GPT-3.5 Turbo          & $2.90 \pm 1.02$ & $2.75 \pm 1.16$ & $2.40 \pm 0.68$ & $3.10 \pm 0.72$ \\
    GPT-4.5 Preview        & $3.75 \pm 1.29$ & $3.60 \pm 1.19$ & $3.15 \pm 0.59$ & $3.90 \pm 0.64$ \\
    GPT-4o                 & $4.35 \pm 1.09$ & $4.25 \pm 1.12$ & $3.45 \pm 0.60$ & $4.35 \pm 0.49$ \\
    Grok 3 Beta            & $4.70 \pm 0.73$ & $4.50 \pm 0.76$ & $3.75 \pm 0.44$ & $4.85 \pm 0.37$ \\
    LLaMA 3 405B           & $4.65 \pm 0.67$ & $4.45 \pm 0.83$ & $3.30 \pm 0.57$ & $4.40 \pm 0.50$ \\
    LLaMA 4 Scout          & $4.45 \pm 1.00$ & $4.10 \pm 1.12$ & $3.15 \pm 0.59$ & $4.35 \pm 0.59$ \\
    Mistral Large          & $4.75 \pm 0.64$ & $4.55 \pm 0.76$ & $3.25 \pm 0.55$ & $4.30 \pm 0.47$ \\
    OpenAI O3              & $5.00 \pm 0.00$ & $4.95 \pm 0.22$ & $4.30 \pm 0.47$ & $5.00 \pm 0.00$ \\
    Qwen3                  & $5.00 \pm 0.00$ & $4.95 \pm 0.22$ & $3.90 \pm 0.31$ & $5.00 \pm 0.00$ \\
    \bottomrule
  \end{tabular}
\end{table}

\subsection{Claude judge results} \label{app:part_d_claude_judge_results}
Scores are reported as mean $\pm$ standard deviation for each dimension.

\begin{table}[H]
  \centering
  \caption{Task I (Creative Reuse and Improvement), Claude judge: dimension scores (mean $\pm$ SD) by model.}
  \label{tab:claude_taskA_scores}
  \small
  \setlength{\tabcolsep}{6pt}
  \renewcommand{\arraystretch}{1.15}
  \begin{tabular}{lcccc}
    \toprule
    \textbf{Model} & \textbf{Fluency} & \textbf{Flexibility} & \textbf{Originality} & \textbf{Elaboration} \\
    \midrule
    Claude 3.5 Sonnet      & $5.00 \pm 0.00$ & $5.00 \pm 0.00$ & $3.80 \pm 0.41$ & $2.80 \pm 1.01$ \\
    Claude 3.7 Sonnet      & $5.00 \pm 0.00$ & $5.00 \pm 0.00$ & $4.00 \pm 0.00$ & $2.90 \pm 0.64$ \\
    DeepSeek Chat          & $5.00 \pm 0.00$ & $5.00 \pm 0.00$ & $4.00 \pm 0.00$ & $3.75 \pm 0.64$ \\
    DeepSeek Reasoner      & $5.00 \pm 0.00$ & $5.00 \pm 0.00$ & $4.10 \pm 0.31$ & $4.95 \pm 0.22$ \\
    Gemini 1.5 Pro         & $5.00 \pm 0.00$ & $5.00 \pm 0.00$ & $3.85 \pm 0.37$ & $4.25 \pm 0.44$ \\
    Gemini 2.0 Flash       & $5.00 \pm 0.00$ & $5.00 \pm 0.00$ & $4.00 \pm 0.00$ & $4.85 \pm 0.37$ \\
    Gemini 2.5 Pro Preview & $5.00 \pm 0.00$ & $5.00 \pm 0.00$ & $4.05 \pm 0.22$ & $4.55 \pm 0.51$ \\
    GPT-3.5 Turbo          & $5.00 \pm 0.00$ & $4.95 \pm 0.22$ & $3.10 \pm 0.55$ & $2.20 \pm 0.89$ \\
    GPT-4.5 Preview        & $5.00 \pm 0.00$ & $5.00 \pm 0.00$ & $4.00 \pm 0.00$ & $4.15 \pm 0.49$ \\
    GPT-4o                 & $5.00 \pm 0.00$ & $5.00 \pm 0.00$ & $4.00 \pm 0.00$ & $3.80 \pm 0.70$ \\
    Grok 3 Beta            & $5.00 \pm 0.00$ & $5.00 \pm 0.00$ & $4.00 \pm 0.00$ & $4.85 \pm 0.37$ \\
    LLaMA 3 405B           & $4.95 \pm 0.22$ & $5.00 \pm 0.00$ & $3.80 \pm 0.41$ & $2.40 \pm 0.60$ \\
    LLaMA 4 Scout          & $5.00 \pm 0.00$ & $5.00 \pm 0.00$ & $3.70 \pm 0.47$ & $2.80 \pm 0.89$ \\
    Mistral Large          & $5.00 \pm 0.00$ & $5.00 \pm 0.00$ & $3.45 \pm 0.51$ & $3.20 \pm 0.83$ \\
    OpenAI O3              & $5.00 \pm 0.00$ & $5.00 \pm 0.00$ & $4.10 \pm 0.31$ & $3.75 \pm 0.55$ \\
    Qwen3                  & $5.00 \pm 0.00$ & $5.00 \pm 0.00$ & $4.00 \pm 0.00$ & $3.55 \pm 0.60$ \\
    \bottomrule
  \end{tabular}
\end{table}

\begin{table}[H]
  \centering
  \caption{Task II (Implications and Adaptations), Claude judge: dimension scores (mean $\pm$ SD) by model.}
  \label{tab:claude_taskB_scores}
  \small
  \setlength{\tabcolsep}{6pt}
  \renewcommand{\arraystretch}{1.15}
  \begin{tabular}{lcccc}
    \toprule
    \textbf{Model} & \textbf{Fluency} & \textbf{Flexibility} & \textbf{Originality} & \textbf{Elaboration} \\
    \midrule
    Claude 3.5 Sonnet      & $1.25 \pm 0.44$ & $1.20 \pm 0.41$ & $2.65 \pm 0.67$ & $3.10 \pm 0.55$ \\
    Claude 3.7 Sonnet      & $1.70 \pm 0.92$ & $1.75 \pm 0.91$ & $2.80 \pm 0.89$ & $3.90 \pm 0.31$ \\
    DeepSeek Chat          & $2.50 \pm 1.05$ & $2.65 \pm 1.04$ & $3.10 \pm 0.72$ & $4.05 \pm 0.51$ \\
    DeepSeek Reasoner      & $2.35 \pm 1.31$ & $2.80 \pm 1.15$ & $3.80 \pm 0.52$ & $5.00 \pm 0.00$ \\
    Gemini 1.5 Pro         & $1.45 \pm 0.76$ & $1.60 \pm 0.88$ & $2.90 \pm 0.79$ & $3.45 \pm 1.00$ \\
    Gemini 2.0 Flash       & $1.65 \pm 1.09$ & $1.90 \pm 1.07$ & $3.05 \pm 0.69$ & $4.35 \pm 0.81$ \\
    Gemini 2.5 Pro Preview & $1.75 \pm 1.07$ & $2.00 \pm 1.08$ & $3.35 \pm 0.75$ & $4.70 \pm 0.57$ \\
    GPT-3.5 Turbo          & $1.50 \pm 0.76$ & $1.40 \pm 0.60$ & $2.05 \pm 0.60$ & $3.25 \pm 0.72$ \\
    GPT-4.5 Preview        & $1.25 \pm 0.44$ & $1.35 \pm 0.67$ & $2.50 \pm 0.61$ & $3.10 \pm 0.85$ \\
    GPT-4o                 & $1.65 \pm 0.93$ & $1.60 \pm 0.99$ & $2.35 \pm 0.67$ & $3.45 \pm 0.83$ \\
    Grok 3 Beta            & $2.25 \pm 1.25$ & $2.15 \pm 1.14$ & $2.55 \pm 0.89$ & $4.25 \pm 0.72$ \\
    LLaMA 3 405B           & $4.60 \pm 0.50$ & $4.75 \pm 0.44$ & $3.95 \pm 0.39$ & $3.80 \pm 0.83$ \\
    LLaMA 4 Scout          & $4.65 \pm 0.81$ & $4.70 \pm 0.57$ & $3.90 \pm 0.31$ & $4.25 \pm 0.72$ \\
    Mistral Large          & $2.55 \pm 1.54$ & $2.65 \pm 1.46$ & $2.50 \pm 0.61$ & $4.55 \pm 0.60$ \\
    OpenAI O3              & $1.85 \pm 1.23$ & $2.00 \pm 1.30$ & $3.35 \pm 0.67$ & $4.35 \pm 0.67$ \\
    Qwen3                  & $1.50 \pm 1.05$ & $1.65 \pm 1.23$ & $3.50 \pm 0.69$ & $3.80 \pm 0.70$ \\
    \bottomrule
  \end{tabular}
\end{table}

\begin{table}[H]
  \centering
  \caption{Task III (Speculative Narrative), Claude judge: dimension scores (mean $\pm$ SD) by model.}
  \label{tab:claude_taskC_scores}
  \small
  \setlength{\tabcolsep}{6pt}
  \renewcommand{\arraystretch}{1.15}
  \begin{tabular}{lcccc}
    \toprule
    \textbf{Model} & \textbf{Fluency} & \textbf{Flexibility} & \textbf{Originality} & \textbf{Elaboration} \\
    \midrule
    Claude 3.5 Sonnet      & $4.35 \pm 0.49$ & $4.30 \pm 0.47$ & $4.10 \pm 0.45$ & $3.95 \pm 0.39$ \\
    Claude 3.7 Sonnet      & $4.30 \pm 0.47$ & $4.35 \pm 0.49$ & $4.00 \pm 0.32$ & $4.35 \pm 0.49$ \\
    DeepSeek Chat          & $4.30 \pm 0.47$ & $4.35 \pm 0.49$ & $4.05 \pm 0.51$ & $4.20 \pm 0.62$ \\
    DeepSeek Reasoner      & $4.65 \pm 0.59$ & $4.55 \pm 0.76$ & $4.25 \pm 0.44$ & $4.95 \pm 0.22$ \\
    Gemini 1.5 Pro         & $4.55 \pm 0.51$ & $4.50 \pm 0.51$ & $4.15 \pm 0.49$ & $4.75 \pm 0.44$ \\
    Gemini 2.0 Flash       & $4.65 \pm 0.49$ & $4.60 \pm 0.50$ & $4.15 \pm 0.37$ & $4.90 \pm 0.31$ \\
    Gemini 2.5 Pro Preview & $4.75 \pm 0.44$ & $4.70 \pm 0.47$ & $4.30 \pm 0.47$ & $5.00 \pm 0.00$ \\
    GPT-3.5 Turbo          & $3.65 \pm 0.59$ & $3.35 \pm 0.81$ & $3.00 \pm 0.73$ & $3.40 \pm 0.75$ \\
    GPT-4.5 Preview        & $4.10 \pm 0.45$ & $4.10 \pm 0.64$ & $3.85 \pm 0.59$ & $4.40 \pm 0.50$ \\
    GPT-4o                 & $4.50 \pm 0.61$ & $4.40 \pm 0.75$ & $4.05 \pm 0.39$ & $4.90 \pm 0.31$ \\
    Grok 3 Beta            & $4.75 \pm 0.44$ & $4.50 \pm 0.51$ & $4.20 \pm 0.41$ & $5.00 \pm 0.00$ \\
    LLaMA 3 405B           & $3.65 \pm 1.04$ & $3.70 \pm 0.92$ & $3.60 \pm 0.50$ & $3.55 \pm 1.36$ \\
    LLaMA 4 Scout          & $4.40 \pm 0.50$ & $4.30 \pm 0.47$ & $3.80 \pm 0.41$ & $4.65 \pm 0.49$ \\
    Mistral Large          & $4.50 \pm 0.61$ & $4.30 \pm 0.73$ & $3.80 \pm 0.41$ & $4.60 \pm 0.60$ \\
    OpenAI O3              & $4.85 \pm 0.37$ & $4.85 \pm 0.37$ & $4.50 \pm 0.51$ & $5.00 \pm 0.00$ \\
    Qwen3                  & $4.70 \pm 0.47$ & $4.70 \pm 0.47$ & $4.20 \pm 0.41$ & $4.95 \pm 0.22$ \\
    \bottomrule
  \end{tabular}
\end{table}

\begin{table}[H]
  \centering
  \caption{Task IV (Practical Innovation), Claude judge: dimension scores (mean $\pm$ SD) by model.}
  \label{tab:claude_taskD_scores}
  \small
  \setlength{\tabcolsep}{6pt}
  \renewcommand{\arraystretch}{1.15}
  \begin{tabular}{lcccc}
    \toprule
    \textbf{Model} & \textbf{Fluency} & \textbf{Flexibility} & \textbf{Originality} & \textbf{Elaboration} \\
    \midrule
    Claude 3.5 Sonnet      & $4.35 \pm 1.09$ & $4.40 \pm 0.99$ & $3.80 \pm 0.41$ & $3.75 \pm 0.79$ \\
    Claude 3.7 Sonnet      & $4.60 \pm 0.60$ & $4.80 \pm 0.41$ & $3.95 \pm 0.39$ & $3.95 \pm 0.76$ \\
    DeepSeek Chat          & $4.30 \pm 0.98$ & $4.55 \pm 0.83$ & $3.65 \pm 0.59$ & $4.10 \pm 0.85$ \\
    DeepSeek Reasoner      & $4.55 \pm 0.94$ & $4.65 \pm 0.81$ & $3.75 \pm 0.55$ & $4.95 \pm 0.22$ \\
    Gemini 1.5 Pro         & $4.35 \pm 1.23$ & $4.45 \pm 1.23$ & $3.95 \pm 0.22$ & $3.75 \pm 0.85$ \\
    Gemini 2.0 Flash       & $4.10 \pm 1.45$ & $4.25 \pm 1.33$ & $3.75 \pm 0.64$ & $4.50 \pm 0.76$ \\
    Gemini 2.5 Pro Preview & $4.55 \pm 1.10$ & $4.65 \pm 0.99$ & $4.05 \pm 0.39$ & $5.00 \pm 0.00$ \\
    GPT-3.5 Turbo          & $3.05 \pm 1.23$ & $3.05 \pm 1.50$ & $2.20 \pm 0.70$ & $2.65 \pm 0.67$ \\
    GPT-4.5 Preview        & $3.40 \pm 1.60$ & $3.65 \pm 1.69$ & $3.10 \pm 0.79$ & $3.45 \pm 0.83$ \\
    GPT-4o                 & $4.40 \pm 1.05$ & $4.35 \pm 1.14$ & $3.45 \pm 0.69$ & $4.00 \pm 0.46$ \\
    Grok 3 Beta            & $4.50 \pm 0.89$ & $4.65 \pm 0.75$ & $3.80 \pm 0.41$ & $4.90 \pm 0.31$ \\
    LLaMA 3 405B           & $4.40 \pm 0.99$ & $4.55 \pm 0.94$ & $3.40 \pm 0.68$ & $3.90 \pm 0.72$ \\
    LLaMA 4 Scout          & $4.15 \pm 1.14$ & $4.25 \pm 1.25$ & $3.15 \pm 0.75$ & $4.25 \pm 0.64$ \\
    Mistral Large          & $4.60 \pm 0.75$ & $4.70 \pm 0.57$ & $3.30 \pm 0.73$ & $4.10 \pm 0.64$ \\
    OpenAI O3              & $4.95 \pm 0.22$ & $4.95 \pm 0.22$ & $4.15 \pm 0.37$ & $5.00 \pm 0.00$ \\
    Qwen3                  & $4.60 \pm 0.82$ & $4.65 \pm 0.75$ & $3.85 \pm 0.37$ & $4.75 \pm 0.55$ \\
    \bottomrule
  \end{tabular}
\end{table}
\subsection{Gemini judge results} \label{app:part_d_gemini_judge}
Scores are reported as mean $\pm$ standard deviation for each dimension.

\begin{table}[H]
  \centering
  \caption{Task I (Creative Reuse and Improvement), Gemini 2.0 Flash judge: dimension scores (mean $\pm$ SD) by model.}
  \label{tab:gemini_taskI_scores}
  \small
  \setlength{\tabcolsep}{6pt}
  \renewcommand{\arraystretch}{1.15}
  \begin{tabular}{lcccc}
    \toprule
    \textbf{Model} & \textbf{Fluency} & \textbf{Flexibility} & \textbf{Originality} & \textbf{Elaboration} \\
    \midrule
    Claude 3.5 Sonnet      & $5.00 \pm 0.00$ & $5.00 \pm 0.00$ & $3.15 \pm 0.37$ & $2.50 \pm 0.51$ \\
    Claude 3.7 Sonnet      & $5.00 \pm 0.00$ & $4.85 \pm 0.37$ & $3.20 \pm 0.41$ & $2.50 \pm 0.51$ \\
    DeepSeek Chat          & $5.00 \pm 0.00$ & $4.85 \pm 0.49$ & $3.15 \pm 0.37$ & $2.65 \pm 0.49$ \\
    DeepSeek Reasoner      & $5.00 \pm 0.00$ & $4.85 \pm 0.49$ & $3.60 \pm 0.50$ & $3.40 \pm 0.50$ \\
    Gemini 1.5 Pro         & $5.00 \pm 0.00$ & $4.25 \pm 0.72$ & $3.00 \pm 0.32$ & $3.05 \pm 0.22$ \\
    Gemini 2.0 Flash       & $5.00 \pm 0.00$ & $4.85 \pm 0.37$ & $3.20 \pm 0.41$ & $3.35 \pm 0.49$ \\
    Gemini 2.5 Pro Preview & $5.00 \pm 0.00$ & $4.80 \pm 0.52$ & $3.45 \pm 0.51$ & $3.45 \pm 0.51$ \\
    GPT-3.5 Turbo          & $5.00 \pm 0.00$ & $4.15 \pm 0.49$ & $2.95 \pm 0.39$ & $2.15 \pm 0.59$ \\
    GPT-4.5 Preview        & $5.00 \pm 0.00$ & $5.00 \pm 0.00$ & $3.25 \pm 0.44$ & $3.05 \pm 0.39$ \\
    GPT-4o                 & $5.00 \pm 0.00$ & $4.95 \pm 0.22$ & $3.20 \pm 0.41$ & $2.90 \pm 0.31$ \\
    Grok 3 Beta            & $5.00 \pm 0.00$ & $5.00 \pm 0.00$ & $3.60 \pm 0.50$ & $3.60 \pm 0.50$ \\
    LLaMA 3 405B           & $5.00 \pm 0.00$ & $4.95 \pm 0.22$ & $3.25 \pm 0.44$ & $2.50 \pm 0.51$ \\
    LLaMA 4 Scout          & $5.00 \pm 0.00$ & $4.90 \pm 0.31$ & $3.25 \pm 0.44$ & $2.60 \pm 0.50$ \\
    Mistral Large          & $5.00 \pm 0.00$ & $4.95 \pm 0.22$ & $3.05 \pm 0.22$ & $2.65 \pm 0.49$ \\
    OpenAI O3              & $5.00 \pm 0.00$ & $5.00 \pm 0.00$ & $3.90 \pm 0.31$ & $2.90 \pm 0.45$ \\
    Qwen3                  & $5.00 \pm 0.00$ & $5.00 \pm 0.00$ & $3.75 \pm 0.44$ & $3.40 \pm 0.50$ \\
    \bottomrule
  \end{tabular}
\end{table}

\begin{table}[H]
  \centering
  \caption{Task II (Implications and Adaptations), Gemini 2.0 Flash judge: dimension scores (mean $\pm$ SD) by model.}
  \label{tab:gemini_taskII_scores}
  \small
  \setlength{\tabcolsep}{6pt}
  \renewcommand{\arraystretch}{1.15}
  \begin{tabular}{lcccc}
    \toprule
    \textbf{Model} & \textbf{Fluency} & \textbf{Flexibility} & \textbf{Originality} & \textbf{Elaboration} \\
    \midrule
    Claude 3.5 Sonnet      & $1.25 \pm 0.55$ & $1.30 \pm 0.57$ & $2.50 \pm 0.61$ & $3.10 \pm 0.31$ \\
    Claude 3.7 Sonnet      & $2.30 \pm 1.08$ & $2.25 \pm 0.79$ & $2.65 \pm 0.59$ & $3.40 \pm 0.50$ \\
    DeepSeek Chat          & $2.25 \pm 0.79$ & $2.35 \pm 0.88$ & $2.70 \pm 0.47$ & $3.40 \pm 0.50$ \\
    DeepSeek Reasoner      & $4.05 \pm 1.00$ & $3.70 \pm 0.66$ & $3.40 \pm 0.60$ & $4.05 \pm 0.22$ \\
    Gemini 1.5 Pro         & $1.75 \pm 1.07$ & $1.80 \pm 1.06$ & $2.70 \pm 0.66$ & $3.20 \pm 0.52$ \\
    Gemini 2.0 Flash       & $2.30 \pm 1.03$ & $2.40 \pm 0.94$ & $2.85 \pm 0.49$ & $3.55 \pm 0.60$ \\
    Gemini 2.5 Pro Preview & $2.75 \pm 0.91$ & $2.70 \pm 0.92$ & $3.00 \pm 0.65$ & $3.80 \pm 0.41$ \\
    GPT-3.5 Turbo          & $1.60 \pm 0.68$ & $1.40 \pm 0.60$ & $2.10 \pm 0.45$ & $2.95 \pm 0.22$ \\
    GPT-4.5 Preview        & $1.40 \pm 0.75$ & $1.60 \pm 0.94$ & $2.40 \pm 0.60$ & $3.15 \pm 0.37$ \\
    GPT-4o                 & $1.65 \pm 0.93$ & $1.65 \pm 0.81$ & $2.25 \pm 0.44$ & $3.10 \pm 0.45$ \\
    Grok 3 Beta            & $2.50 \pm 1.15$ & $2.20 \pm 0.95$ & $2.50 \pm 0.69$ & $3.55 \pm 0.51$ \\
    LLaMA 3 405B           & $4.60 \pm 0.82$ & $4.30 \pm 0.73$ & $3.70 \pm 0.47$ & $3.70 \pm 0.47$ \\
    LLaMA 4 Scout          & $4.60 \pm 0.82$ & $4.15 \pm 0.75$ & $3.45 \pm 0.51$ & $3.80 \pm 0.41$ \\
    Mistral Large          & $2.80 \pm 1.11$ & $2.65 \pm 1.31$ & $2.60 \pm 0.60$ & $3.70 \pm 0.47$ \\
    OpenAI O3              & $2.35 \pm 1.39$ & $2.30 \pm 1.38$ & $3.20 \pm 0.62$ & $3.75 \pm 0.44$ \\
    Qwen3                  & $4.60 \pm 0.82$ & $4.55 \pm 0.69$ & $3.45 \pm 0.51$ & $4.00 \pm 0.00$ \\
    \bottomrule
  \end{tabular}
\end{table}

\begin{table}[H]
  \centering
  \caption{Task III (Speculative Narrative), Gemini 2.0 Flash judge: dimension scores (mean $\pm$ SD) by model.}
  \label{tab:gemini_taskIII_scores}
  \small
  \setlength{\tabcolsep}{6pt}
  \renewcommand{\arraystretch}{1.15}
  \begin{tabular}{lcccc}
    \toprule
    \textbf{Model} & \textbf{Fluency} & \textbf{Flexibility} & \textbf{Originality} & \textbf{Elaboration} \\
    \midrule
    Claude 3.5 Sonnet      & $3.75 \pm 0.64$ & $3.15 \pm 0.59$ & $3.45 \pm 0.51$ & $3.55 \pm 0.51$ \\
    Claude 3.7 Sonnet      & $3.85 \pm 0.67$ & $3.25 \pm 0.55$ & $3.65 \pm 0.59$ & $3.75 \pm 0.44$ \\
    DeepSeek Chat          & $3.65 \pm 0.75$ & $3.20 \pm 0.41$ & $3.40 \pm 0.50$ & $3.55 \pm 0.51$ \\
    DeepSeek Reasoner      & $4.45 \pm 0.60$ & $3.40 \pm 0.88$ & $4.00 \pm 0.32$ & $4.05 \pm 0.22$ \\
    Gemini 1.5 Pro         & $4.20 \pm 0.62$ & $3.30 \pm 0.47$ & $3.65 \pm 0.49$ & $4.00 \pm 0.00$ \\
    Gemini 2.0 Flash       & $4.20 \pm 0.52$ & $3.40 \pm 0.60$ & $3.80 \pm 0.41$ & $4.00 \pm 0.00$ \\
    Gemini 2.5 Pro Preview & $4.45 \pm 0.60$ & $3.75 \pm 0.72$ & $3.85 \pm 0.37$ & $4.00 \pm 0.00$ \\
    GPT-3.5 Turbo          & $3.20 \pm 0.62$ & $2.40 \pm 0.68$ & $2.80 \pm 0.41$ & $2.95 \pm 0.51$ \\
    GPT-4.5 Preview        & $3.75 \pm 0.64$ & $3.10 \pm 0.45$ & $3.45 \pm 0.51$ & $3.85 \pm 0.37$ \\
    GPT-4o                 & $4.10 \pm 0.64$ & $3.40 \pm 0.68$ & $3.50 \pm 0.51$ & $4.00 \pm 0.00$ \\
    Grok 3 Beta            & $4.35 \pm 0.59$ & $3.55 \pm 0.83$ & $3.65 \pm 0.49$ & $4.00 \pm 0.00$ \\
    LLaMA 3 405B           & $3.75 \pm 1.07$ & $3.00 \pm 0.92$ & $3.10 \pm 0.91$ & $3.55 \pm 1.00$ \\
    LLaMA 4 Scout          & $4.10 \pm 0.72$ & $3.20 \pm 0.70$ & $3.20 \pm 0.41$ & $3.90 \pm 0.31$ \\
    Mistral Large          & $4.00 \pm 0.73$ & $3.00 \pm 0.73$ & $3.25 \pm 0.44$ & $3.75 \pm 0.44$ \\
    OpenAI O3              & $4.70 \pm 0.47$ & $3.85 \pm 0.81$ & $4.10 \pm 0.31$ & $4.10 \pm 0.31$ \\
    Qwen3                  & $4.95 \pm 0.22$ & $4.15 \pm 0.59$ & $3.85 \pm 0.37$ & $4.00 \pm 0.00$ \\
    \bottomrule
  \end{tabular}
\end{table}

\begin{table}[H]
  \centering
  \caption{Task IV (Practical Innovation), Gemini 2.0 Flash judge: dimension scores (mean $\pm$ SD) by model.}
  \label{tab:gemini_taskIV_scores}
  \small
  \setlength{\tabcolsep}{6pt}
  \renewcommand{\arraystretch}{1.15}
  \begin{tabular}{lcccc}
    \toprule
    \textbf{Model} & \textbf{Fluency} & \textbf{Flexibility} & \textbf{Originality} & \textbf{Elaboration} \\
    \midrule
    Claude 3.5 Sonnet      & $4.55 \pm 0.83$ & $4.15 \pm 0.88$ & $3.30 \pm 0.47$ & $3.45 \pm 0.51$ \\
    Claude 3.7 Sonnet      & $4.30 \pm 0.73$ & $4.20 \pm 0.70$ & $3.20 \pm 0.52$ & $3.45 \pm 0.51$ \\
    DeepSeek Chat          & $4.40 \pm 0.82$ & $4.20 \pm 0.83$ & $3.10 \pm 0.55$ & $3.55 \pm 0.51$ \\
    DeepSeek Reasoner      & $4.80 \pm 0.62$ & $4.60 \pm 0.68$ & $3.45 \pm 0.60$ & $3.95 \pm 0.22$ \\
    Gemini 1.5 Pro         & $4.20 \pm 1.15$ & $4.05 \pm 1.10$ & $3.25 \pm 0.44$ & $3.30 \pm 0.57$ \\
    Gemini 2.0 Flash       & $4.45 \pm 1.10$ & $4.25 \pm 0.97$ & $3.35 \pm 0.49$ & $3.75 \pm 0.44$ \\
    Gemini 2.5 Pro Preview & $4.85 \pm 0.49$ & $4.35 \pm 0.75$ & $3.40 \pm 0.50$ & $4.00 \pm 0.00$ \\
    GPT-3.5 Turbo          & $2.65 \pm 1.23$ & $2.45 \pm 1.15$ & $2.05 \pm 0.60$ & $2.45 \pm 0.51$ \\
    GPT-4.5 Preview        & $3.25 \pm 1.45$ & $3.10 \pm 1.33$ & $2.85 \pm 0.59$ & $3.10 \pm 0.45$ \\
    GPT-4o                 & $4.30 \pm 1.17$ & $4.00 \pm 1.26$ & $3.00 \pm 0.46$ & $3.45 \pm 0.51$ \\
    Grok 3 Beta            & $4.60 \pm 0.75$ & $4.30 \pm 0.73$ & $3.30 \pm 0.47$ & $3.95 \pm 0.22$ \\
    LLaMA 3 405B           & $4.55 \pm 0.60$ & $4.30 \pm 0.86$ & $3.10 \pm 0.45$ & $3.50 \pm 0.51$ \\
    LLaMA 4 Scout          & $4.40 \pm 0.99$ & $3.80 \pm 0.95$ & $2.95 \pm 0.22$ & $3.65 \pm 0.49$ \\
    Mistral Large          & $4.50 \pm 0.76$ & $4.20 \pm 0.83$ & $3.10 \pm 0.55$ & $3.60 \pm 0.50$ \\
    OpenAI O3              & $4.95 \pm 0.22$ & $4.85 \pm 0.37$ & $3.70 \pm 0.47$ & $3.95 \pm 0.22$ \\
    Qwen3                  & $5.00 \pm 0.00$ & $4.90 \pm 0.31$ & $3.50 \pm 0.51$ & $4.00 \pm 0.00$ \\
    \bottomrule
  \end{tabular}
\end{table}

\end{appendices}

\end{document}